\documentclass[default,iicol]{sn-jnl}

\usepackage{graphicx}%
\usepackage{multirow}%
\usepackage{amsmath,amssymb,amsfonts}%
\usepackage{amsthm}%
\usepackage{mathrsfs}%
\usepackage[title]{appendix}%
\usepackage{xcolor}%
\usepackage{textcomp}%
\usepackage{manyfoot}%
\usepackage{booktabs}%
\usepackage{algorithm}%
\usepackage{algorithmicx}%
\usepackage{algpseudocode}%
\usepackage{listings}%
\usepackage{caption}
\usepackage{subcaption}
\usepackage[bottom]{footmisc}
\usepackage[utf8]{inputenc}
\usepackage{enumitem}
\usepackage{tabularx}
\usepackage{ragged2e}
\usepackage{stfloats}
\usepackage{xspace}

\usepackage{rotating}

\usepackage[sort&compress,capitalize,nameinlink]{cleveref}
\creflabelformat{equation}{#2#1#3}

\newcommand\Tstrut{\rule{0pt}{2.6ex}}         
\newcommand\Bstrut{\rule[-0.8ex]{0pt}{0pt}}

\newcommand{\pps}{panoptic-part segmentation}

\newcommand{\ours}{JPPF (Ours)}
\newcommand{\things}{\emph{things}}
\newcommand{\stuff}{\emph{stuff}}

\makeatletter
\newcommand*{\onedot}{\@ifnextchar{.}{}{.\@\xspace}}
\makeatother

\newcommand{\eg}{{e.g}\onedot} \newcommand{\Eg}{{E.g}\onedot}
\newcommand{\ie}{{i.e}\onedot} \newcommand{\Ie}{{I.e}\onedot}
\newcommand{\cf}{{c.f}\onedot} 
 \newcommand{\vs}{{vs}\onedot}

\begin{document}

\title[JPPF]{JPPF: Multi-task Fusion for Consistent Panoptic-Part Segmentation}


\author[1]{\fnm{Shishir} \sur{Muralidhara}}\email{shishir.muralidhara@dfki.de}

\author[1]{\fnm{Sravan Kumar} \sur{Jagadeesh}}\email{sravan.jagadeesh@dfki.de}

\author*[1]{\fnm{René} \sur{Schuster}}\email{rene.schuster@dfki.de}

\author[1]{\fnm{Didier} \sur{Stricker}}\email{didier.stricker@dfki.de}

\affil[1]{\orgdiv{Augmented Vision}, \orgname{German Research Center for Artificial Intelligence -- DFKI}, \orgaddress{\street{Trippstadter Straße 122}, \city{Kaiserslautern}, \postcode{67663}, \country{Germany}}}


\abstract{
Part-aware panoptic segmentation is a problem of computer vision that aims to provide a semantic understanding of the scene at multiple levels of granularity. More precisely, semantic areas, object instances, and semantic parts are predicted simultaneously. In this paper, we present our Joint Panoptic Part Fusion (JPPF) that combines the three individual segmentations effectively to obtain a panoptic-part segmentation. Two aspects are of utmost importance for this: First, a unified model for the three problems is desired that allows for mutually improved and consistent representation learning. Second, balancing the combination so that it gives equal importance to all individual results during fusion. Our proposed JPPF is parameter-free and dynamically balances its input. The method is evaluated and compared on the Cityscapes Panoptic Parts (CPP) and Pascal Panoptic Parts (PPP) datasets in terms of PartPQ and Part-Whole Quality (PWQ). In extensive experiments, we verify the importance of our fair fusion, highlight its most significant impact for areas that can be further segmented into parts, and demonstrate the generalization capabilities of our design without fine-tuning on 5 additional datasets.
}


\maketitle

\section{Introduction} \label{sec:intro}
Humans are able to perceive various levels of detail and abstraction of a scene.
We can not only understand different semantic categories such as bus, car, and sky, but we can also distinguish between individual entities (instances) and their components (parts), such as windows or wheels.
In computer vision, the estimation of these parallel layers of abstraction has recently been introduced as panoptic-part segmentation \cite{de2021part}.
Yet, there exists no completely unified and joint approach for this problem.

According to \citep{cordts2016cityscapes}, the two pieces that make up a scene are \stuff{} and \things{}.
Things are countable objects such as persons, cars, or buses, whereas \stuff{}, like the sky or road, is usually amorphous and innumerable.
Those two categories are identified in the well studied tasks of semantic segmentation and instance segmentation.
However, both tasks are incapable of describing the entirety of the scene.
To fill this gap, panoptic segmentation \citep{PanopticSegKiri} was presented, which recognizes and segments both, \stuff{} and \things.
After this, several approaches for panoptic segmentation have been proposed \citep{ChengPD2019,kirillov2019panoptic,li2020unifying,mohan2021efficientps,PorziSSS,yuwenUPSnet}.

Part segmentation, or part parsing, on the other hand, seeks to semantically analyze the image based on part-level.
There has been some effort in this area, where part segmentation is often treated as a semantic segmentation problem \citep{gong2019graphonomy,jiang2018cnn,jiang2019cnn,li2017holistic,liu2018cross,luo2013pedestrian}.
A few methods are instance-aware \citep{gong2018instance,li2017holistic,zhao2018understanding} and even fewer handle multi-class part objects \citep{zhao2019multi,michieli2020gmnet}.

With the release of datasets for panoptic-part segmentation \citep{de2021part, meletis2020cityscapes}, the first methods for this problem have been proposed \citep{li2022panoptic, jagadeesh2023jppf, li2023panopticpartformer}.
In \citep{de2021part}, a baseline approach is presented in which two networks for panoptic and part segmentation are used.
These two networks are trained independently and the results of both are combined using a uni-directional (top-down) merging strategy.
This technique of independent training has significant drawbacks.
Due to the use of two different networks, there is a computational overhead.
As the authors employ different networks, there will be no consistency in their predictions, making the merging process ineffective.
Also, the independent training strategy leads to learning redundancy since they could potentially share semantic information between segmentation heads.

Afterwards, Panoptic-PartFormer (PPF) \citep{li2022panoptic} has been proposed, in which the authors present a unified, combined transformer for \things, \stuff, and parts that iteratively refines the individual segmentations to achieve consistency.
In this design, redundancies are avoided and similarities between tasks are exploited, but we argue that an explicit modeling of multi-task fusion can produce more accurate results.

\begin{figure}[t]
    \centering
    \includegraphics[width=\linewidth]{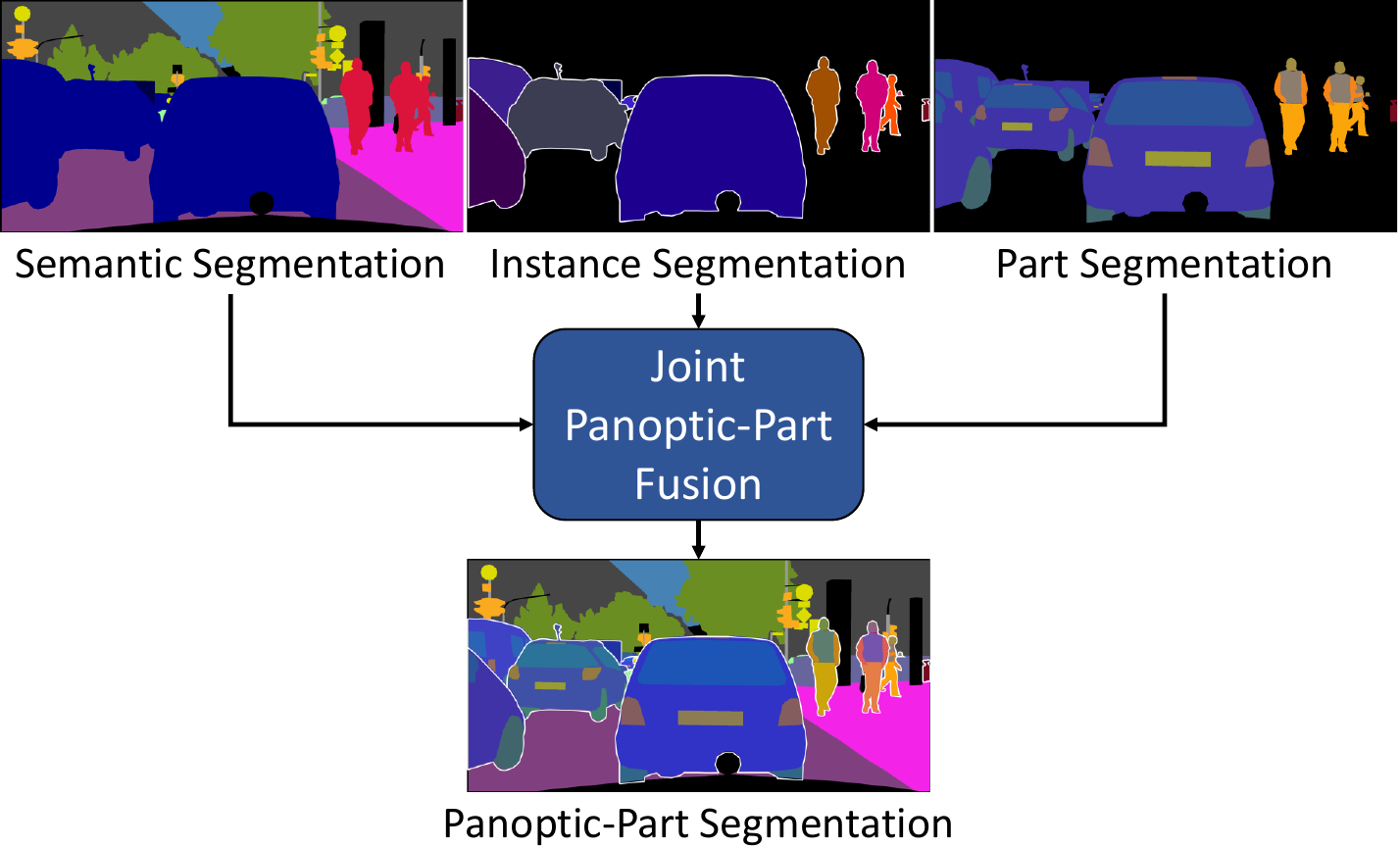}
    \caption{Our Joint Panoptic-Part Fusion (JPPF) combines individual predictions into a consistent panoptic-part segmentation}
    \label{fig:teaser}
\end{figure}

To this end, and to overcome the limitations of the top-down merging, we have presented a Joint Panoptic-Part Fusion (JPPF) for panoptic-part segmentation in \citep{jagadeesh2023jppf}, in which each sub-task is treated equally to allow for mutual benefits and maximal consistency (\cf \cref{fig:teaser}).
By sharing a backbone for all three tasks, the joint fusion is outperforming the top-down baseline, while being more efficient at the same time.

In this work, we re-present our JPPF \citep{jagadeesh2023jppf} and extend the experiments, validation, and discussion.
In short, 
 \begin{itemize}
    \item we present a single neural network that uses a shared encoder to perform semantic, instance, and part segmentation and fuses them efficiently to produce panoptic-part segmentation.
    \item we propose a parameter-free Joint Panoptic-Part Fusion (JPPF) module that dynamically considers the logits from the semantic, instance, and part head and consistently integrates the three predictions.
    \item we conduct a thorough analysis of our approach and demonstrate the efficacy, accuracy, and consistency of the joint fusion strategy.
    \item we obtain state-of-the-art results for panoptic-part segmentation on various datasets and metrics, surpassing our previous work \citep{jagadeesh2023jppf}, the top-down baseline \citep{de2021part}, and the transformer-based competitor PPF \citep{li2022panoptic}.
    \item we demonstrate that our approach generalizes to many other datasets without fine-tuning.
 \end{itemize}

\section{Related Work} \label{sec:related}

\subsection{Towards Panoptic-Part Segmentation} \label{sec:related:segmentation}

Part-aware panoptic segmentation \citep{de2021part} is a recently introduced problem that brings semantic, instance, and part segmentation together. There have been several methods proposed for these individual tasks, including panoptic segmentation, which is a blend of semantic and instance segmentation.

\paragraph{Semantic Segmentation}
PSPnet \citep{zhao2017pyramid} introduced the pyramid pooling module, which focuses on the importance of multi-scale features by learning them at many scales, then concatenating and up-sampling them.
\citet{ChenPSA17} proposed Atrous Spatial Pyramid Pooling (ASPP), which is based on spatial pyramid pooling and combines features from several parallel atrous convolutions with varying dilation rates, as well as global average pooling.
The incorporation of multi-scale characteristics and the capturing of global context increases computational complexity.
So, \citet{chen2018a} introduced the Dense Prediction Cell (DPC)  and \citet{Valada2018} suggested multi-scale residual units with changing dilation rates to compute high-resolution features at various spatial densities, as well as an efficient atrous spatial pyramid pooling module called eASPP to learn multi-scale representation with fewer parameters and a broader receptive field.
In the encoder-decoder architecture, a lot of effort has been advocated for improving the decoder's upsampling layer.
\citet{chen2018b} extend DeepLabV3 \citep{ChenPSA17} by adding an efficient decoder module to enhance segmentation results at object boundaries.
Later, \citet{Tian2019} suggest replacing it with data-dependent up-sampling (DUpsampling), which can recover pixel-wise prediction from low-resolution CNN outputs and take advantage of the redundant label space in semantic segmentation. 

\paragraph{Instance Segmentation}
Here, we mainly concentrate on proposal based approaches.
\citet{HariharanAGM14} proposed a simultaneous object recognition and segmentation technique that uses Multi-scale Combinatorial Grouping (MCG) \citep{Pont-TusetABMM15} to generate proposals and then run them through a CNN for feature extraction.
In addition, \citet{HariharanAGM14a} presented a hyper-column pixel descriptor that captures feature representations of all layers in a CNN with a strong correlation for simultaneous object detection and segmentation.
\citet{PinheiroCD15} proposed the DeepMask network, which employs a CNN to predict the segmentation mask of each object as well as the likelihood of the object being in the patch. FCIS \citep{LiFCIS2017} employs position sensitive inside/outside score maps to simultaneously predict object detection and segmentation.  Later, one of the most popular networks for instance segmentation, Mask-RCNN \citep{HeGDG17}, was introduced. It extends Faster-RCNN \citep{RenHG015} with an extra network that segments each of the detected objects. RoI-align, which preserves exact spatial position, replaces RoI-pool, which performs coarse spatial quantization for feature encoding.

\paragraph{Part Segmentation}
Dense part-level segmentation, on the other hand, is instance agnostic and is regarded as a semantic segmentation problem \citep{gong2019graphonomy,jiang2018cnn,jiang2019cnn,li2017holistic,liu2018cross,luo2018trusted,michieli2020gmnet,zhao2019multi}. Most of the research has been conducted to perform human part parsing \citep{zhao2018understanding,gong2018instance,dong2013deformable,ladicky2013human,li2020self,liang2018look,lin2020cross,ruan2019devil,yang2019parsing}, and only little work has addressed multi-part segmentation tasks \citep{zhao2019multi, michieli2020gmnet}.

\paragraph{Panoptic Segmentation}
The authors of \citep{PanopticSegKiri} combined the output of two independent networks for semantic and instance segmentation and coined the term panoptic segmentation. Panoptic segmentation approaches can be divided into top-down methods \citep{LiAnurag2018, LiuPS2019, Jie2018, yuwenUPSnet, SofibarkonAdaptIS, PorziSSS} that prioritize semantic segmentation prediction and bottom-up methods \citep{Tien2019DL, ChengPD2019, NaiyuSSAP2019} that prioritize instance prediction.
Our previous in \citep{jagadeesh2023jppf} builds on EfficientPS \citep{mohan2021efficientps} and extends this model to obtain panoptic-part segmentation.
This work, builds on our previous design of a joint architecture and exploits its modularity to replace individual components.

\subsection{Panoptic-Part Segmentation} \label{sec:related:partpanoptic}

\subsubsection{Datasets and Baselines} \label{sec:related:partpanoptic:degeus}
\begin{figure}[t]
    \centering
    \includegraphics[width=0.405\linewidth]{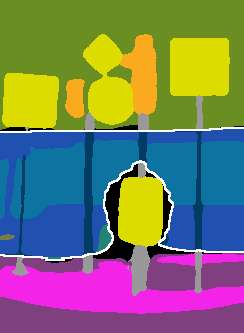}
    \includegraphics[width=0.58\linewidth]{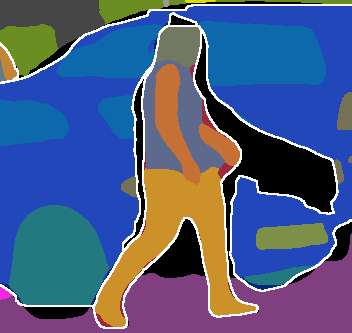}
    \caption{A typical issue with the top-down merging approach of \cite{de2021part} are the gaps around the contours of objects due to inconsistencies and difficulties in distinguishing between \textit{stuff} and \textit{things}}
    \label{fig:topdown}
\end{figure}

In recent years, Part-Aware Panoptic Segmentation \citep{de2021part} was introduced, which aims at a unified scene and part-parsing.
Also, \citet{de2021part} introduced a baseline model using a state-of-the-art panoptic segmentation network and a part segmentation network, merging them using heuristics.
The panoptic and part segmentation is merged in top-down or bottom-up manner.
In the top-down merge, the prediction from panoptic segmentation is re-used for scene-level semantic classes that do not consist of parts. Then for partitionable semantic classes, the corresponding segment of the part prediction is extracted.
In case of conflicting predictions, a void label will be assigned.
According to \citet{de2021part}, top-down merge produces better results than the bottom-up approach.
In addition, their paper has released two datasets with panoptic-part annotations: Cityscapes Panoptic Part (CPP) dataset and Pascal Panoptic Part (PPP) dataset \citep{meletis2020cityscapes}.
Along with the drawbacks of employing independent networks as mentioned in \cref{sec:intro}, there are concerns with the usage of top-down merge.
Due to inconsistencies, top-down merging may result in undefined regions around the contours of objects. Due to some imbalance between \stuff\ and \things{}, it also has trouble separating them.
These issues are highlighted in \cref{fig:topdown}.
Furthermore, the uni-directional merge accounts higher importance to one of the predictions, neglecting the potential of mutual refinement during fusion.
With our unified fusion for semantics, instances, and parts, we resolve these issues, giving equal priority to all individual predictions.

\subsubsection{Unified Models} \label{sec:related:partpanoptic:ppf}
Panoptic-PartFormer (PPF) \citep{li2022panoptic} was developed in parallel to \citep{jagadeesh2023jppf} and follows a similar goal as our line of work: To unify panoptic-part segmentation.
However, the authors of \cite{li2022panoptic} approach the unification from the other side.
While we suggest a unified fusion module to combine individual results in a well balanced manner, they propose a shared encoder and transformer-based decoder to predict \stuff{}, \things{}, and parts together via a single model.
This way, they achieve remarkable consistency and results, however though the prediction of the individual tasks is fully unified in a single architecture that uses task-specific queries, it is followed by the same uni-directional top-down merging as in \citep{de2021part}, leading to \textit{void} labels where inconsistencies remain.

\Citet{li2023panopticpartformer} propose a second version of their Panoptic-PartFormer (PPF++), in which they also introduce a new metric, called Part-Whole Quality (PWQ).
Compared to the PartPQ of \citep{de2021part}, PWQ is supposed to resolve the bias towards the PQ metric of panoptic segmentation.
In our experiments, we will consider both these metrics for thorough comparisons.

\section{Unified Panoptic-Part Segmentation} \label{sec:method}
The main contribution of our work is the Joint Panoptic-Part Fusion (JPPF) that produces highly dense and consistent panoptic-part segmentations in an efficient manner.
To obtain individual predictions for our fusion, in theory any method could be applied.
However, we argue that a combined network for all three segmentation tasks produces better results through mutual learning and reasoning.
Therefore in this section, we first formalize the problem of panoptic-part segmentation, then explain our unified network architecture presented in \citep{jagadeesh2023jppf}, and lastly describe the inner workings of JPPF.

\begin{figure*}[t]
    \centering
    \includegraphics[width=0.9\linewidth]{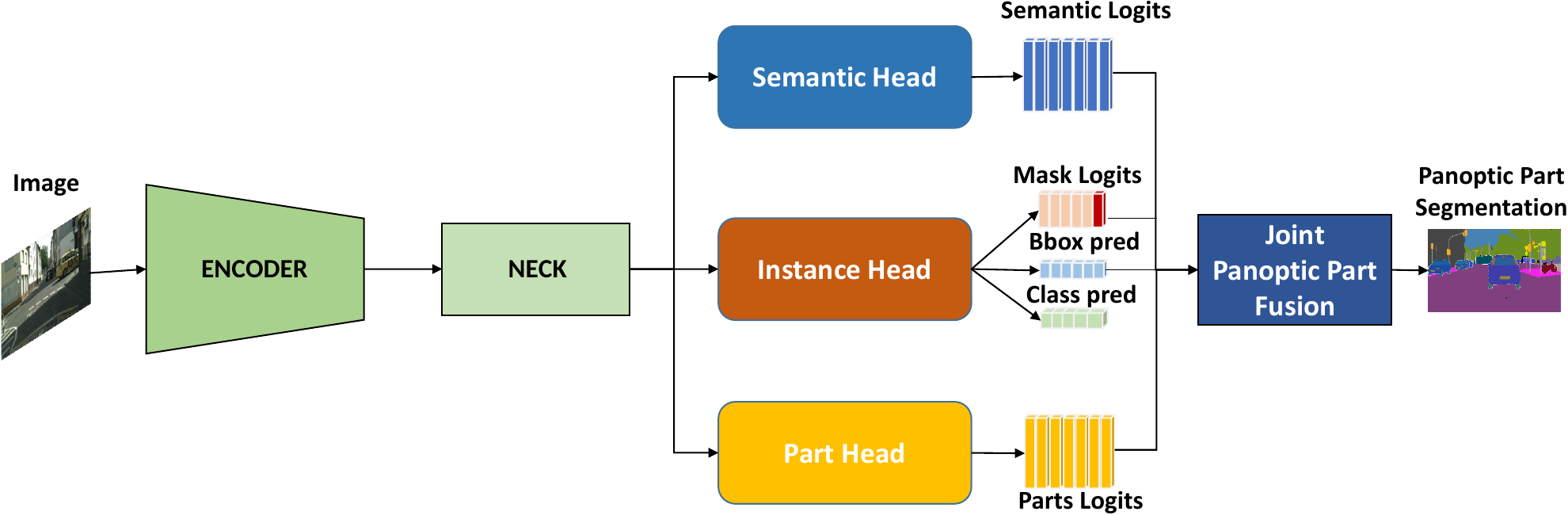}
    \caption{Our overall architecture for panoptic-part segmentation features a shared encoder, three specialized prediction heads, and the unified joint fusion module. Its modular structure allows to easily replace the feature backbone or use intermediate results from other approaches to perform a consistent fusion}
    \label{fig:network}
\end{figure*}

\subsection{Panoptic-Part Segmentation} \label{sec:method:pps}
The goal of panoptic-part segmentation is to predict a panoptic-part label $(s, id, p)$ for each pixel of an image $I$.
Here, $s$ represents semantic scene level class, $p$ represents the part-level class and $id$ indicates the instance identifier for each object.
It is important to note, that not all pixels in an image may represent all components of panoptic-part segmentation, \eg \stuff{} is not instantiable, and there are many semantic classes for which it does not make sense to further subdivide them into parts, \eg the sky.
Anyhow, the three labels can be obtained independently, however a valid panoptic-part label must be consistent, \ie free from contradiction.
\Eg a car can not share the object identifier of a bicycle or consist of human body parts.
Achieving this consistency is the fundamental challenge in panoptic-part segmentation.
To obtain this goal, different strategies can be followed, including naive merging \citep{de2021part}, joint prediction \citep{li2023panopticpartformer, li2022panoptic}, or -- as in our case -- fusion \citep{jagadeesh2023jppf}.

\subsection{Overall Architecture} \label{sec:method:network}
To obtain individual predictions for semantics, instances, and parts, our previous work extends EfficientPS \citep{mohan2021efficientps} by incorporating a part segmentation head.
We reuse the backbone, semantic head, and instance head of EfficientPS.
As part segmentation can be regarded as a semantic segmentation problem, we are replicating the architecture of the semantic branch of EfficientPS and train it for part-level segmentation.
All three resulting heads share a common backbone -- in our case EfficientNet \citep{tan2019efficientnet} -- which helps to ensure that the predictions made by the heads are consistent with one another.
Sharing a single representation for all three tasks improves efficiency and is beneficial during learning, as shown by our experiments in \cref{sec:results:encoder}.
An overview of the architecture of our proposed model is shown in \cref{fig:network}.

\subsubsection{Part Segmentation Head}  \label{sec:method:network:parts}
According to previous work \citep{de2021part}, the grouping of parts yields better results.
We have verified this finding for our architecture in \citep{jagadeesh2023jppf} and consequently follow the same principle and group semantically identical parts, \eg the windows of cars and buses are grouped into a single window class.
The grouping of elements allows the network to learn without ambiguity and provides more data per class for training.
Additionally, we represent all non-partitionable semantic classes as a single background class within our part head.
This avoids redundant predictions across different heads and further balances the learning of parts versus other classes.
Both groupings of classes (semantic grouping of parts, as well as grouping of the background) can later be reverted into class-specific parts by the additional information of the other prediction heads to obtain a fine-grained panoptic-part segmentation.

\subsection{Joint Panoptic-Part Fusion} \label{sec:method:fusion}
\begin{figure*}[t]
    \centering
    \includegraphics[width=\linewidth]{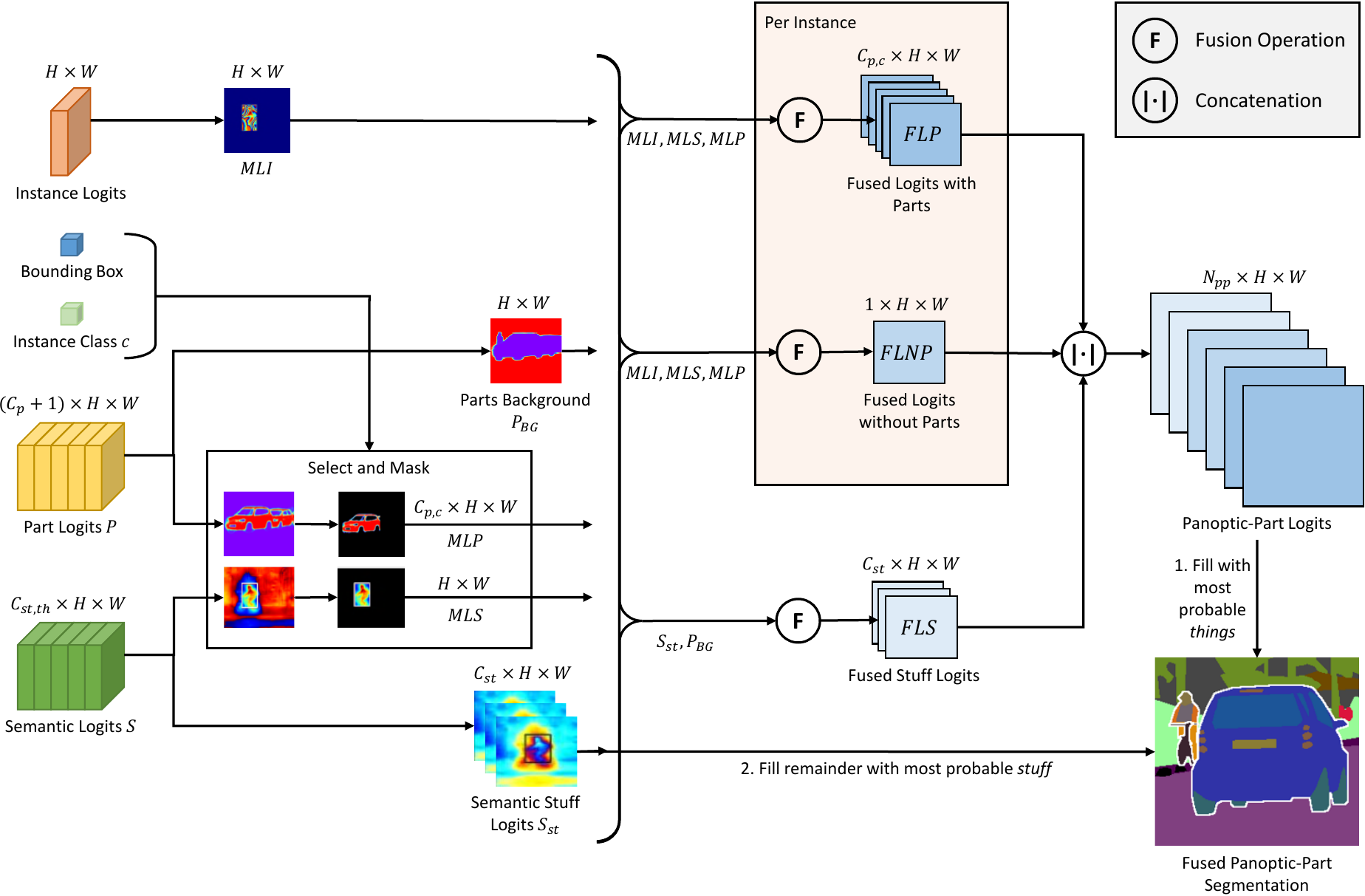}
     \caption{Illustration of our proposed joint fusion module. For simplicity, we illustrate the process for a single instance object. Semantic, instance, and part predictions are equally balanced and combined}
     \label{fig:fusion}
\end{figure*}

The mutual combination of the predictions for semantic segmentation, instance segmentation, and part segmentation are the core of our work.
Inspired by the panoptic fusion module of EfficientPS \citep{mohan2021efficientps}, we propose a module that jointly fuses the individual results of the three heads by giving each prediction equal priority and thoroughly exploiting coherent predictions.
Given the definition of panoptic-part segmentation, we identified four possible cases for fusion: Partitionable and non-partitionable \stuff{}, and partitionable and non-partitionable \things{}.
In the following, we will first describe the required input for our fusion module and then describe the three combinations which actually occur in the existing datasets (partitionable \stuff{} is not included).
However, our approach generalized to the missing case as well.
\cref{fig:fusion} depicts our JPPF module.

\begin{figure*}
    \centering
    \includegraphics[width=\linewidth]{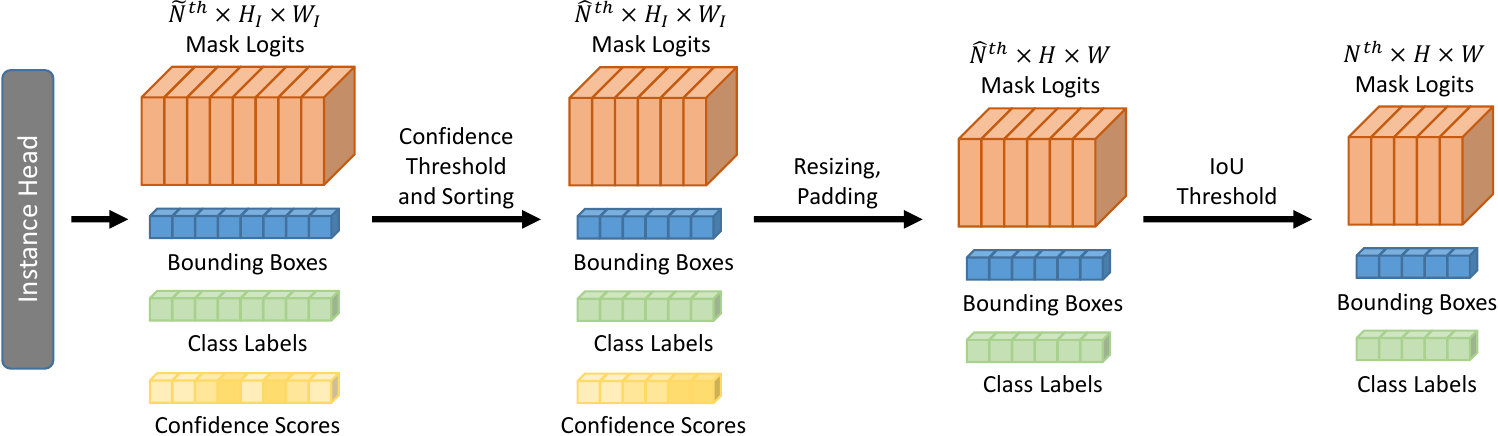}
    \caption{Illustration of the pre-processing steps in \citep{mohan2021efficientps} for predictions from the instance head. The remaining instances serve as input for our fusion}
    \label{fig:preprocessing}
\end{figure*}

\paragraph{Input and Pre-Processing}
The input for our fusion are the individual dense predictions for semantics, instances, and parts.
In our complete architecture, these are obtained from the three prediction heads using the shared backbone, but it could be any other source that satisfies the preconditions.
More precisely, we require three input components:
\begin{enumerate}
    \item A map of semantic logits $S \in \mathrm{R}^3$ of shape $C_{st,th} \times H \times W$ in the interval $[0,1]$ (\eg via softmax activation), in which $H$ and $W$ are spatial dimensions of the input image (potentially resized) and $C_{st,th} = C_{st} + C_{th}$ is the total number of semantic classes.
    \item A set of instance predictions for the \things{} classes, each consisting of: 
    \begin{enumerate}
        \item A softmax-activated map of logits $M$ of shape $H_I \times W_I$ representing the object mask.
        \item An axis-aligned 2D bounding box.
        \item A class label $c$ for this object.
        \item A confidence score in the interval $[0,1]$.
    \end{enumerate}
    \item A map of part logits $P \in \mathrm{R}^3$ of shape $(C_p+1) \times H \times W$ in the interval $[0,1]$, in which $C_p$ is the number of (grouped) part classes.
\end{enumerate}

Before actual fusion, the instance objects are pre-processed, following the steps in \citep{mohan2021efficientps}.
This includes confidence thresholding, confidence based sorting, spatial resizing and padding of the instance-specific mask logits and box coordinates to the relevant input size, \ie from $H_I \times W_I$ to $H \times W$, and a non-maximum suppression based on the overlap and confidence of boxes.
After filtering, there remain $N^{th}$ instances.
The pre-processing is illustrated in \cref{fig:preprocessing}.

\paragraph{Fusion for Things}
For the fusion of \things{}, all three input components are considered, even if the specific class is not further partitionable.
In this case, the generic background class of the part head, can still support this hypothesis during fusion.
The individual instance objects guide our fusion process, however during actual fusion, all three predictions are treated equally.

Given a single one out of the $N^{th} = N^{th}_{np} + N^{th}_p$ pre-processed \things{} instance of class $c$ with its mask logits $MLI$, we first use the resized bounding box to mask the corresponding prediction from the semantic head.
Precisely, class $c$ is sliced out of the semantic logits $S$ and all values outside of the bounding box are set to zero to obtain the masked semantic logits $MLS$.

In case class $c$ is partitionable, then the corresponding subset of size $C_{p,c}$ of the part logits $P$ is selected from the part segmentation head, \eg for an instance of class $c=person$, the part logits for head, torso, legs, and arms ($C_{p,human}=4$) are selected.
These logits are again masked by the corresponding bounding box to produce the third masked logits for parts $MLP$.
If class $c$ can not be segmented into parts, the background class from the part logits is selected instead and masked likewise.
In order to make the fusion operation feasible, we replicate $MLS$ and $MLI$ to match the number of channels in $MLP$.
\Eg, a person instance contains four parts (head, arms, torso, legs), thus $MLP$ is of shape  $4 \times W \times H$. Therefore, $MLS$ and $MLI$ are replicated 4 times to match the shape of $MLP$.
If the instance is not partitionable, $MLP$ consists of the background class only and therefore $MLS$ and $MLI$ are not replicated.

At this point, we have obtained three sets of masked logits.
We are now fusing these individual logits to obtain the fused logits for classes with parts $FLP$ and class without parts $FLNP$.

\paragraph{Fusion Operation}
To compute the fused logits for any of the cases, we propose a uniformed fusion operation.
This operation computes the sum of the sigmoid of the masked logits and the sum of the masked logits and calculates the Hadamard product of both.
The procedure is formalized in \cref{eq:fusion}:
\begin{equation}
FL\left(MLL\right)=\left(\sum_{l \in MLL}\sigma(l)\right) \odot \left(\sum_{l \in MLL} l\right)
\label{eq:fusion}
\end{equation}
In this equation, $\sigma(\cdot)$ denotes the sigmoid function, $\odot$ denotes the Hadamard product, and $MLL$ is a set of equally shaped masked logits which are supposed to be fused, \eg $MLL = \{MLS, MLI, MLP\}$.
This equation describes a generalized version of the fusion proposed by \citet{mohan2021efficientps} that handles arbitrarily many logits.

\paragraph{Fusion for Stuff}
To generate the fused logits $FLS$ for the \stuff\ classes, each of the $C_{st}$ channels from the semantic head are selected and fused with the background channel of the part head in the same manner, \ie according to \cref{eq:fusion}, but this time with only two sets of logits (no instance information).
As mentioned, the same concept would also apply for \stuff{} that is partitionable, \ie selecting the corresponding parts, replicating the \stuff{} logits, followed by pair-wise fusion.

\paragraph{Overall Fusion}
All three fused logits, $FLP$, $FLNP$, and $FLS$, are concatenated along the channel dimension to obtain the intermediate logits, in which each of the $N_{pp}$ channels represents a valid panoptic-part label (see \cref{sec:method:pps}.
The total number of $N_{pp}$ label candidates depends on the number of things $N^{th}$ predicted by the instance head and the number of parts of their classes $C_{p,c}$. 
We produce an intermediate panoptic-part prediction by taking the \textit{argmax} of these intermediate logits.
Precisely, during fusion there will be $N_{pp} = C_{st} + N^{th}_{np} + \sum_{c \in N^{th}_{p}}{C_{p,c}}$ candidate logits.
Finally, we fill an empty canvas with the most probable panoptic-part label for all \things{} and the remaining areas are filled with the prediction for \stuff{} classes extracted from the semantic segmentation head.
During fusion, the fused score increases if the predictions of all three heads are consistent, and likewise it is decreased if the predictions do not match with each other.

\paragraph{Post-Processing}
Areas of \stuff{} classes below a minimum threshold $min_{st}=2048$~pixels are filtered out, as in \citep{mohan2021efficientps}.

\section{Experiments and Results} \label{sec:results}
In this section, we will first introduce the relevant datasets and then provide more details on the implementation and training of our model.
Afterwards, we compare our JPPF to previous work and investigate our design choices in ablative experiments.

\paragraph{Datsets}
For most of our experiments, we use the recently introduced Cityscapes Panoptic Parts (CPP) and Pascal Panoptic Parts (PPP) datasets \citep{meletis2020cityscapes,de2021part}.
CPP provides pixel-level annotations for 19 semantic categories, of which 11 are \stuff\ and 8 are \things\ classes.
Out of the 8 \things{}, 5 classes include annotations at the part level.
There are 2975 images for training and 500 for validation in this finely annotated dataset.
PPP consists of 20 \things\ and 80 \stuff\ classes.
Part-level annotations are provided for 16 of the 20 \things.
As in previous work \citep{meletis2020cityscapes}, we only consider a subset of 59 object classes for training and evaluation, including 20 \things\, 39 \stuff\ classes, and 58 part classes.
These parts are detailed by \citet{michieli2020gmnet} and \citet{zhao2019multi}.
PPP consists of a total of 10103 images which are divided into 4998 images for training and 5105 for validation.
Next to CPP and PPP, we perform some experiments on a variety of other datasets to demonstrate how our method generalizes across domains.

\paragraph{Metrics}
For the evaluation of individual semantic and part segmentations, we apply the  mean Intersection-over-Union (mIoU), and the mean Average Precision (mAP) for instance segmentations. 
For the complete evaluation of combined panoptic-part segmentation, we use the Part Panoptic Quality (PartPQ) \citep{de2021part}, which is an extension of the Panoptic Quality (PQ) that was proposed by  \citet{kirillov2019panoptic}.
Because the authors of \citep{li2023panopticpartformer} identified limitations in the expressiveness and interpretability of the PartPQ metric, they have introduced the Part-Whole Quality (PWQ), which we will also consider in our experiments.

\paragraph{Training and Implementation Details}
For the Cityscapes data, we use images of the original resolution, \ie $1024 \times 2048$ pixels, and resize the input images of PPP to $384\times512$ pixels for training.
We perform data augmentation, scaling and hyperparameter initialization as in EfficientPS \citep{mohan2021efficientps}.
We use a multi-step learning rate ($lr$) and train our network by Stochastic Gradient Descent (SGD) with a momentum of $0.9$.
For the CPP and PPP, we use an initial $lr$ of 0.07 and 0.01, respectively.
We begin the training with a warm-up phase in which the $lr$ is increased linearly from $\frac{1}{3} \cdot lr$ up to $lr$ within $200$ iterations.
The weights of all InPlace-ABN layers \citep{bulo2018place} are frozen, and we train the model for $10$ additional epochs with a fixed learning rate of $10^{-4}$. Finally, we unfreeze the weights of the InPlace-ABN layers and train the model for $50k$ iterations beginning with $lr$ of 0.07 (CPP) and 0.01 (PPP), and reduce $lr$ after $32k$ and $44k$ iteration by a factor of 10.
Four GPUs are used for the training with a batch size of $2$ per GPU for CPP and $8$ per GPU for PPP.
Our feature backbone is the most recent version of EfficientNet -- EfficientNet-L2 \citep{xie2020self}.
This is in contrast to our previous work \citep{jagadeesh2023jppf}, in which we have used the preliminary EfficientNet-B5 \citep{tan2019efficientnet}.
We initialize the backbone with weights pre-trained on COCO \citep{lin2014microsoft}.
The impact of this initialization is quantified in \cref{tab:pretraining}.

\begin{table}[t]
    \centering
    \caption{Comparison of our updated model on CPP with and without pre-trained weights}
    \label{tab:pretraining}
    \begin{tabular}{c|cc}
        Pretrained & PartPQ & PWQ\Bstrut\\
        \hline
        no & 61.4 & 66.7\Tstrut\\
        COCO \citep{lin2014microsoft} & 61.4 & 67.3
    \end{tabular}
\end{table}  

\begin{table}[t]
    \centering
    \caption{Comparison of EfficientPS trained on Panoptic Cityscapes and on Cityscapes Panoptic Parts (CPP)}
    \label{tab:baselinecomparision}
    
    \begin{tabular}{c|c|ccc}
        Network & Data & PQ\Bstrut\\
        \hline
        \multirow{2}{*}{EfficientPS \cite{mohan2021efficientps} } & Panoptic CS \cite{cordts2016cityscapes} & 63.9\Tstrut\\
         & CPP \citep{de2021part} & 62.2
    \end{tabular}
\end{table}

\subsection{Comparison to State-of-the-Art} \label{sec:results:sota}
Our comparison to previous work and state-of-the-art considers the initially introduced baseline in \citep{de2021part} and the more recent unified transformer-based architecture of \citet{li2022panoptic, li2023panopticpartformer}.
The baseline by \citet{de2021part} uses the panoptic labels of the Cityscapes dataset \citep{cordts2016cityscapes} to train a panoptic segmentation network.
Since this data is slightly different from the actual panoptic-part dataset (CPP), a direct, fair comparison is not possible.
This deviation is indicated in \cref{tab:baselinecomparision} that shows the results of EfficientPS \citep{mohan2021efficientps} trained on Cityscapes \vs CPP.
To make the baseline comparable in terms of data, we re-implement the baseline and train it on the same data.
The re-implementation consists of EfficientPS \citep{mohan2021efficientps} for panoptic segmentation, and our part segmentation network with a separate backbone (\cf \cref{sec:method:network:parts}).
Top-down merging is then used to combine the two independent results into a \pps.
The re-implementation and results are in line with our previous work in \citep{jagadeesh2023jppf}.

Finally, we compare our previous and updated model with JPPF to the reproduced baseline, the official baselines of \citet{de2021part}, and multiple variants of both versions of the Panoptic-PartFormer (PPF) \citep{li2023panopticpartformer, li2022panoptic}.
The official baseline consists of EfficientPS \citep{mohan2021efficientps} and BSANet \citep{zhao2019multi} with top-down merging.
The results of this comparison on CPP and PPP are shown in \cref{tab:sota} for single-scale and multi-scale inference.

\begin{table*}[t]
    \centering
    \caption{Comparison of results for panoptic-part segmentation on Cityscapes and Pascal Panoptic Parts \citep{meletis2020cityscapes}. \textit{P} and \textit{NP} refer to areas with and without part labels, respectively. The best result per data setting and metric is highlighted in bold. $^\ast$indicates our reproduced baseline (details in \cref{sec:results:sota})}
    \label{tab:sota}
    \begin{tabular}{llcccc}
        \multirow{2}{*}{Method} & \multirow{2}{*}{Backbone} & \multicolumn{3}{c}{PartPQ} & \multirow{2}{*}{PWQ}\\
         & & All & P & NP\Bstrut\\
         
        \hline
        \multicolumn{6}{c}{Cityscapes Panoptic Parts, Single-Scale}\Tstrut\Bstrut\\
        \hline
        Baseline$^\ast$ & EfficientNet-B5 \citep{tan2019efficientnet} & 57.7 & 44.2 & 62.5 & --\Tstrut\\
        \\[-0.7em]
        \multirow{2}{*}{\ours} & EfficientNet-B5 \citep{tan2019efficientnet} & 59.6 & 47.7 & 63.8 & 66.1 \\
         & EfficientNet-L2 \citep{xie2020self} & \textbf{61.4} & \textbf{49.5} & \textbf{65.7} & \textbf{67.3}\Bstrut\\
        
        \hline
        \multicolumn{6}{c}{Cityscapes Panoptic Parts, Multi-Scale}\Tstrut\Bstrut\\
        \hline
        Baseline \citep{de2021part} & EfficientNet \citep{tan2019efficientnet}, ResNet101 \citep{he2016deep} & 60.2 & 46.1 & 65.2 & --\Tstrut\\
        \\[-0.7em]
        \multirow{2}{*}{PPF \citep{li2022panoptic}} & ResNet50 \citep{he2016deep} & 57.4 & 43.9 & 62.2 & 60.5\\
         & Swin \citep{liu2021swin} & 61.9 & 45.6 & 68.0 & 65.3\\
        \\[-0.7em]
        \multirow{3}{*}{PPF++ \citep{li2023panopticpartformer}} & ResNet50 \citep{he2016deep} & 59.2 & 42.5 & 65.1 & 62.1\\
         & Swin \citep{liu2021swin} & 62.3 & 46.0 & 68.2 & 65.7\\
         & ConvNext \citep{liu2022convnet} & 63.1 & 46.4 & \textbf{69.1} & 66.5\\
        \\[-0.7em]
        \multirow{2}{*}{\ours} & EfficientNet-B5 \citep{tan2019efficientnet} & 61.8 & 50.8 & 65.7 & \textbf{68.2}\\
         & EfficientNet-L2 \citep{xie2020self} & \textbf{63.3} & \textbf{52.2} & 67.2 & \textbf{68.2}\Bstrut\\
        
        \hline
        \multicolumn{6}{c}{Pascal Panoptic Parts, Single-Scale}\Tstrut\Bstrut\\
        \hline
      Baseline-1 \citep{de2021part} & ResNet50 \citep{he2016deep} & 31.4 & 47.2 & 26.0 & --\Tstrut\\
        Baseline-2 \citep{de2021part} & ResNeSt269 \citep{zhang2022resnest} & 38.3 & 51.6 & 33.8 & --\\
        \\[-0.7em]
        \multirow{2}{*}{PPF \citep{li2022panoptic}} & ResNet50 \citep{he2016deep} & 37.8 & -- & -- & 40.2\\
         & ResNet101 \citep{he2016deep} & 39.3 & -- & -- & 41.3\\
        \\[-0.7em]
        \multirow{4}{*}{PPF++ \citep{li2023panopticpartformer}} & ResNet50 \citep{he2016deep} & 42.2 & -- & -- & 45.2\\
         & ResNet101 \citep{he2016deep} & 42.4 & -- & -- & 46.0\\
         & Swin \citep{liu2021swin} & \textbf{49.3} & -- & -- & 52.7\\
         & ConvNext \citep{liu2022convnet} & 48.6 & -- & -- & \textbf{54.2}\\
        \\[-0.7em]
        \multirow{2}{*}{\ours} & EfficientNet-B5 \citep{tan2019efficientnet} & 32.3 & 48.3 & 26.9 & 45.6\\
         & EfficientNet-L2 \citep{xie2020self} & 40.5 & \textbf{58.5} & \textbf{34.4} & 53.7 
    \end{tabular}
\end{table*}

\begin{figure*}[t]
    \centering
    \begin{subfigure}{0.245\linewidth}
\includegraphics[width=\linewidth]{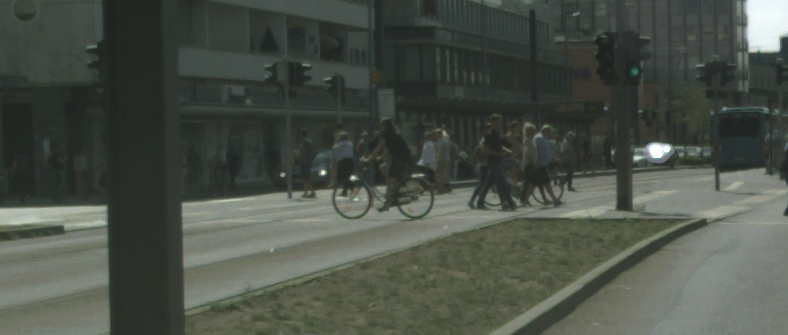}
    \end{subfigure}\hspace*{\fill}
     \begin{subfigure}{0.245\linewidth}
\includegraphics[width=\linewidth]{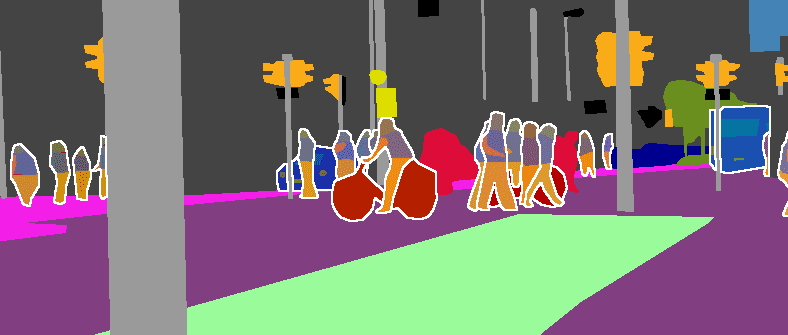}
    \end{subfigure}\hspace*{\fill}
    \begin{subfigure}{0.245\linewidth}
\includegraphics[width=\linewidth]{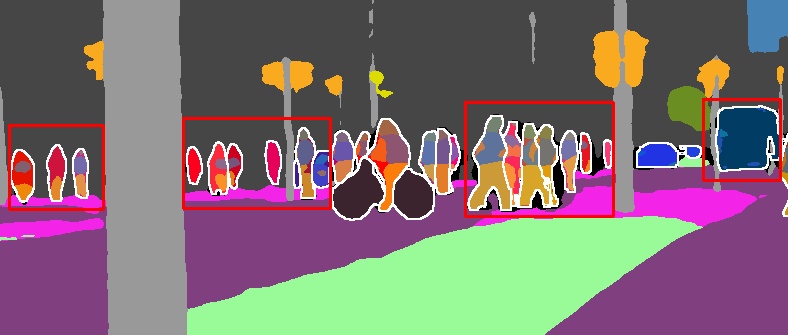}
    \end{subfigure}\hspace*{\fill}
     \begin{subfigure}{0.245\linewidth}
\includegraphics[width=\linewidth]{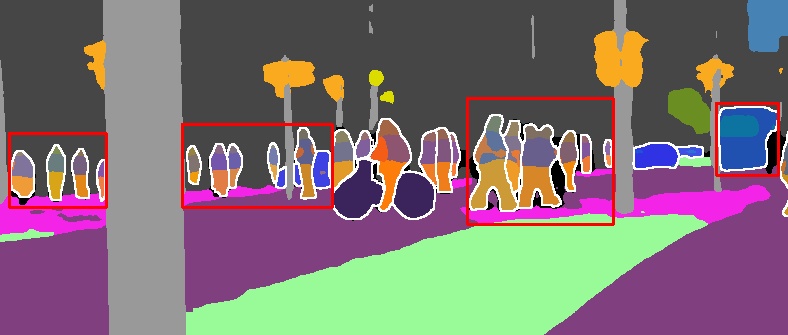}
    \end{subfigure}
    
    \vspace*{\fill}
    
    \begin{subfigure}{0.245\linewidth}
\includegraphics[width=\linewidth]{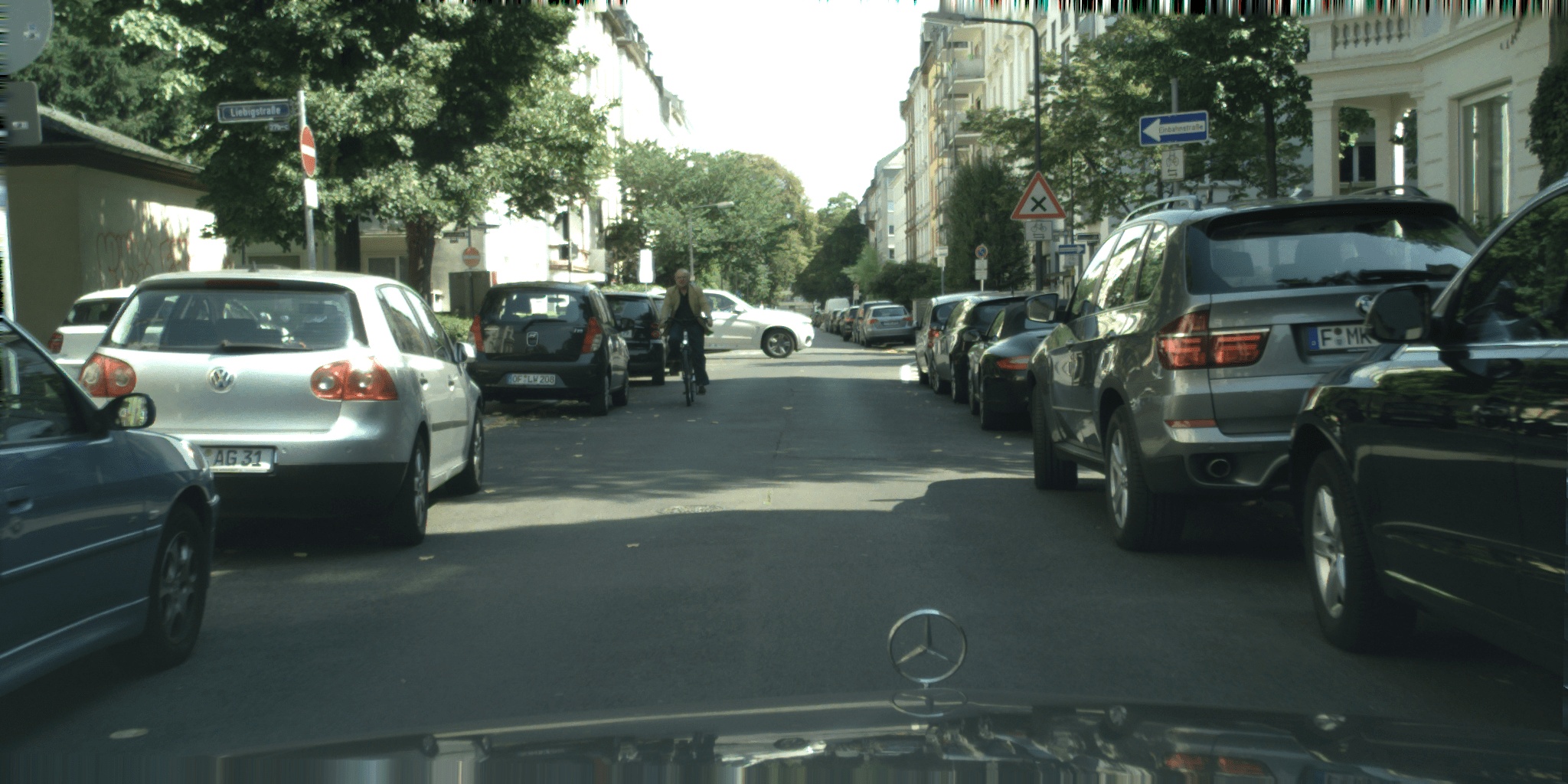}
    \end{subfigure}\hspace*{\fill}
     \begin{subfigure}{0.245\linewidth}
\includegraphics[width=\linewidth]{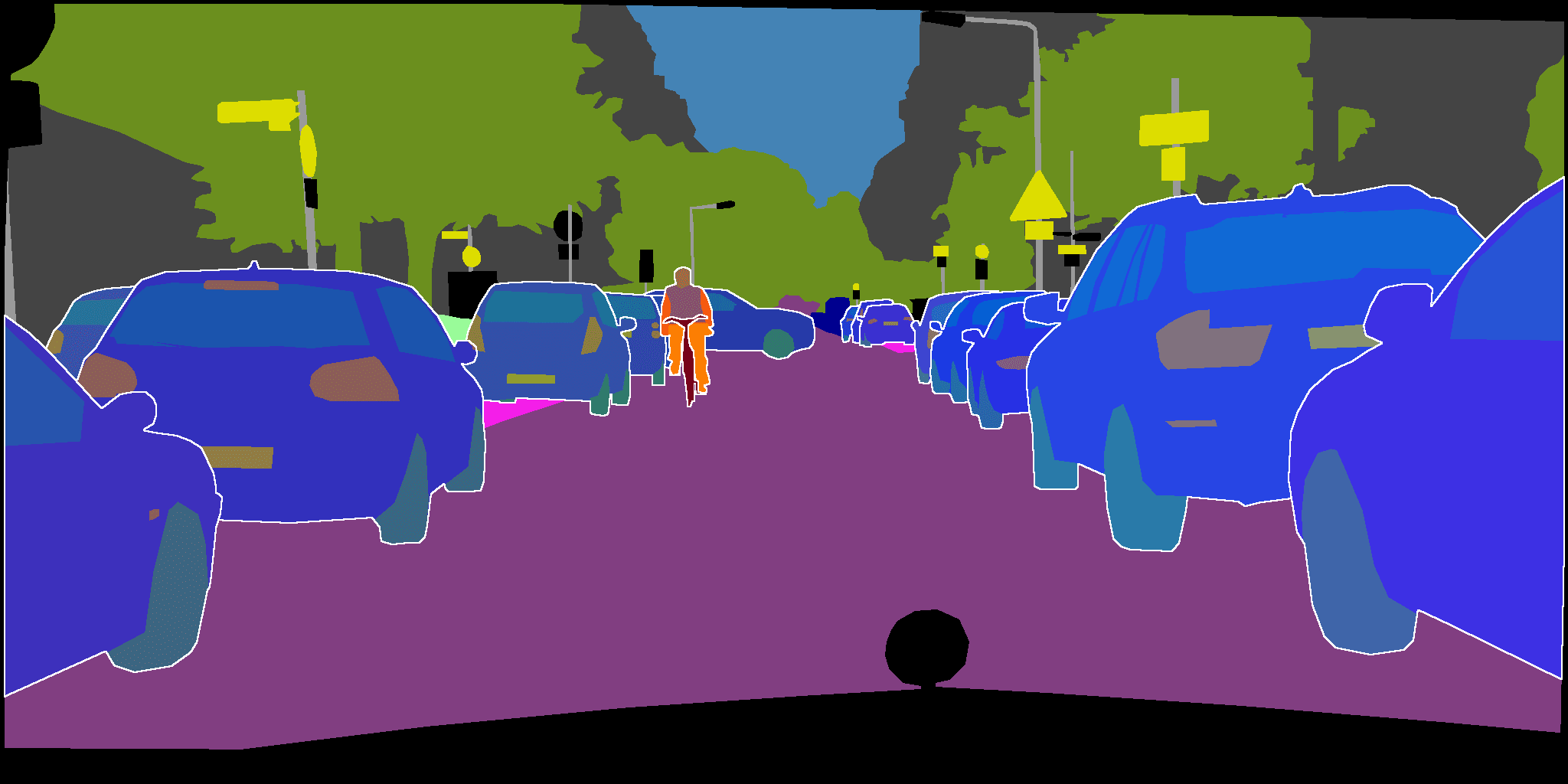}
    \end{subfigure}\hspace*{\fill}
    \begin{subfigure}{0.245\linewidth}
\includegraphics[width=\linewidth]{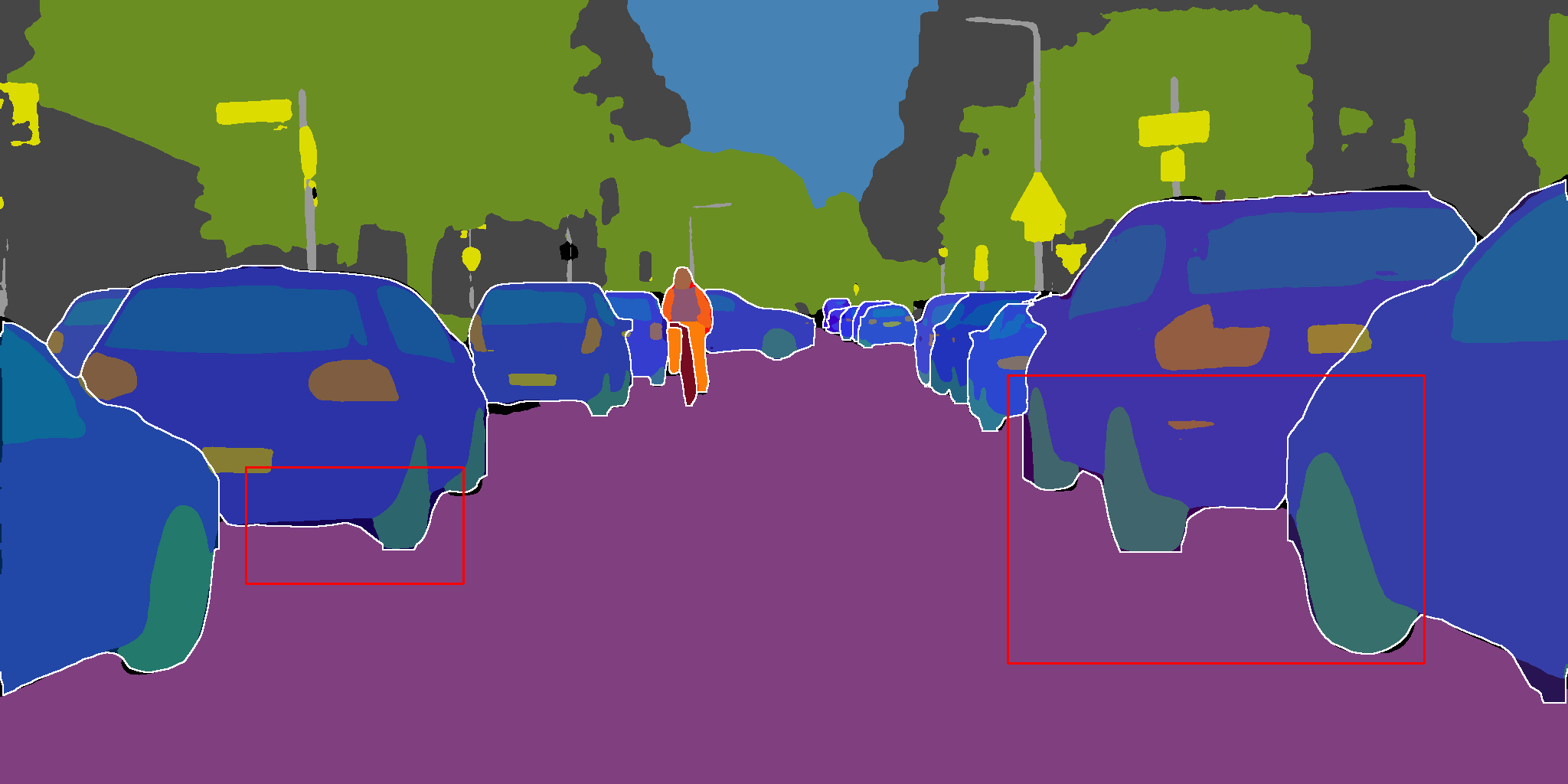}
    \end{subfigure}\hspace*{\fill}
     \begin{subfigure}{0.245\linewidth}
\includegraphics[width=\linewidth]{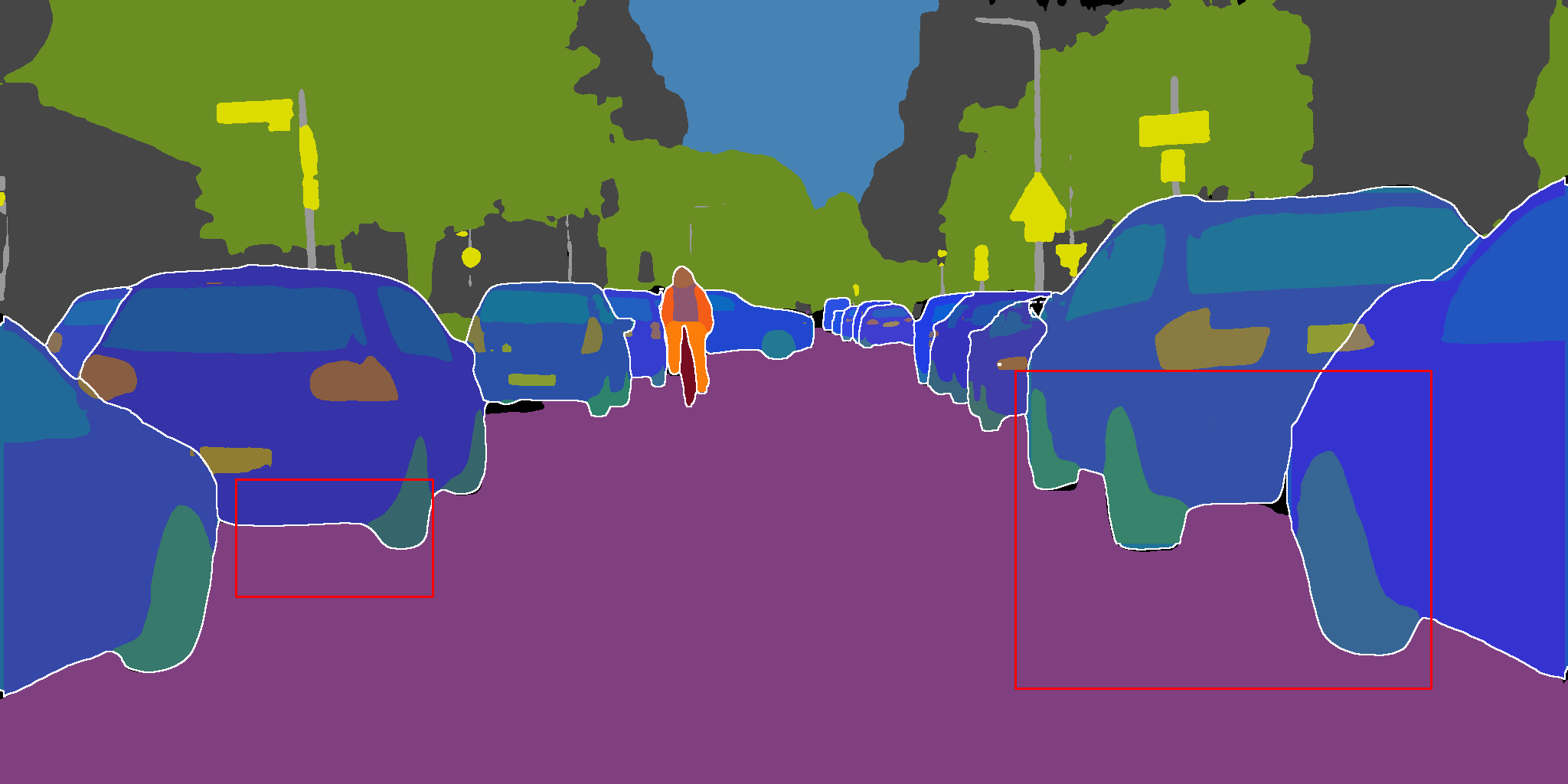}
    \end{subfigure}
    
    \vspace*{\fill}
    \begin{subfigure}{0.245\linewidth}
\includegraphics[width=\linewidth]{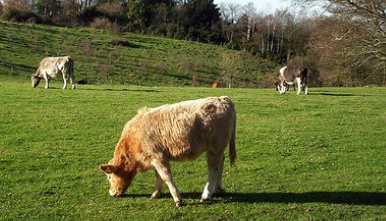}

    \end{subfigure}\hspace*{\fill}
     \begin{subfigure}{0.245\linewidth}
\includegraphics[width=\linewidth]{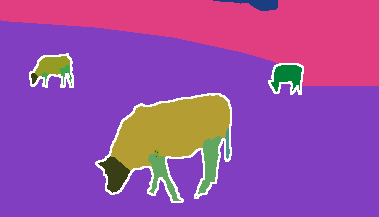}

    \end{subfigure}\hspace*{\fill}
    \begin{subfigure}{0.245\linewidth}
\includegraphics[width=\linewidth]{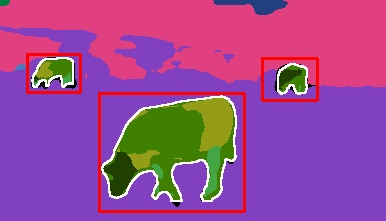}

    \end{subfigure}\hspace*{\fill}
    \begin{subfigure}{0.245\linewidth}
\includegraphics[width=\linewidth]{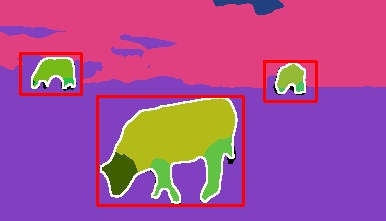}
    \end{subfigure}

    \vspace*{\fill}
   \begin{subfigure}{0.245\linewidth}
\includegraphics[width=\linewidth]{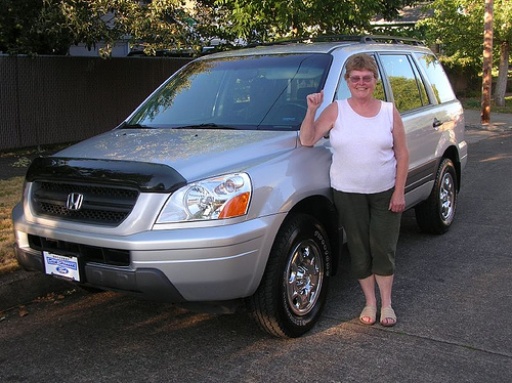}
\subcaption*{Original Image}
    \end{subfigure}\hspace*{\fill}
     \begin{subfigure}{0.245\linewidth}
\includegraphics[width=\linewidth]{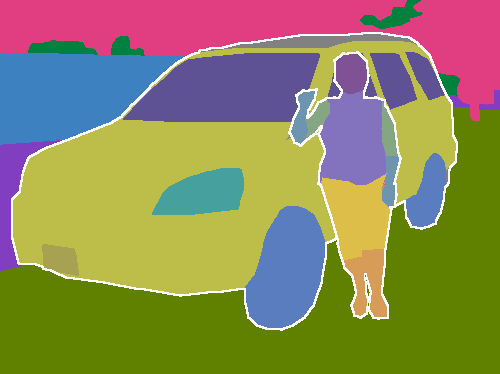}
\subcaption*{Ground-truth}
    \end{subfigure}\hspace*{\fill}
    \begin{subfigure}{0.245\linewidth}
\includegraphics[width=\linewidth]{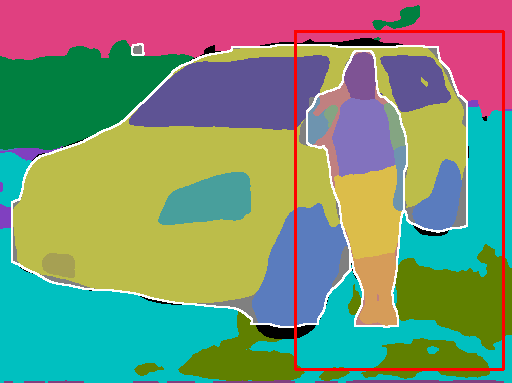}
\subcaption*{Baseline$^\ast$}
    \end{subfigure}\hspace*{\fill}
     \begin{subfigure}{0.245\linewidth}
\includegraphics[width=\linewidth]{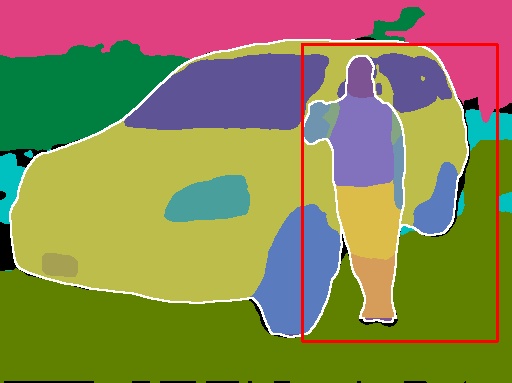}
\subcaption*{JPPF \cite{jagadeesh2023jppf}}
    \end{subfigure}
    
    \caption{Qualitative results of our proposed model on Citscapes Panoptic Parts (first two rows) and Pascal Panoptic Parts (last two rows) compared to our reproduced baseline, ground-truth and the reference image. $^\ast$indicates the reproduced baseline which is detailed in \cref{sec:results:sota}. The results for our JPPF are obtained with the backbone of the previous version. The graphic is adopted from \citep{jagadeesh2023jppf}. More visual examples with our updated backbone for both datasets are provided in the appendix in \cref{fig:appendix:visualcityscapes,fig:appendix:visualpascal}}
    \label{fig:visual}
\end{figure*}

On CPP with single-scale testing, JPPF improves the accuracy significantly compared to the reproduced baseline.
We surpass the reproduced baseline by 3.7 percentage points (pp) in overall PartPQ and by 5.3 pp in $\text{PartPQ}_\text{P}$ with our updated backbone.
Similarly for multi-scale testing, our updated model outperforms the baseline by 3.1 pp and 6.1 pp in PartPQ and $\text{PartPQ}_\text{P}$, respectively.
Our JPFF even outperforms the strong transformer-based competitor PPF and the non-peer reviewed extension PPF++ in terms of PartPQ and PWQ by a small margin.
Especially for areas that can be segmented into parts, we achieve more accurate results, indicating the increased consistency after our fusion and leading to a higher PWQ metric.
Interestingly, for these two metrics ($\text{PartPQ}_\text{P}$ and PWQ), even our single-scale results better over the multi-scale results of any competitor.

For PPP, our model outperforms the top-down combination of DeepLabV3+ \citep{chen2018b} and Mask RCNN \citep{HeGDG17} (\textit{Baseline-1}), even though this baseline was trained with the original Pascal parts and Pascal panoptic segmentation datasets, which provide more annotations.
\textit{Baseline-2} (top-down merging of DeepLabV3-ResNeSt269 \citep{ChenPSA17, zhang2022resnest}, DetectoRS \citep{qiao2020detectors}, and BSANet \citep{zhao2019multi}) yields even better results because of the more advanced backbones, and hence has a higher representational capacity.
Similarly, the more sophisticated transformer of PPF (and PPF++) together with powerful backbone models achieves the best results for PartPQ and PWQ.
However, in partionable areas ($PartPQ_P$), we significantly outperform the baselines on PPP. We believe that this advantage can be attributed to the balanced integration of parts in our fusion module.
In comparison to top-down merging, our design is also slightly favorable in terms of density, as presented in \cref{tab:fusion}.

From \cref{fig:visual}, we can see that our proposed fusion is able to segment the parts of very small and distant object classes reliably.
Also, our proposed fusion solves some typical problems of top-down merging, which are the bifurcation of \things{} by \stuff{} and the inconsistent parts within \things.
As illustrated in \cref{fig:visual}, our fusion gets rid of unknown regions within objects by giving equal priority to all three individual predictions.
In \cref{fig:appendix:visualcityscapes,fig:appendix:visualpascal} we provide more examples and a visual comparison to PPF \cite{li2022panoptic}.
There, we also present failure cases of our model to provide insights into its limitations.
In some cases, especially on PPP, PPF produces finer details compared to our approach. In cluttered areas where small objects occlude each other, our JPPF seems to perform favorably.

\begin{table*}[t]
    \centering
    \caption{Comparison of three independent encoders to our design with a shared feature encoder with different backbones on Cityscapes Panoptic Parts}
    \label{tab:encoder}
    \begin{tabular}{ccccc}
       \multirow{2}{*}{Method} & \multirow{2}{*}{Backbone} & Semantic & Instance & Part\\
        & & mIoU & AP & mIoU\Bstrut\\
        \hline
        Independent Networks & EfficientNet-B5 \citep{tan2019efficientnet} & 78.1 & 37.3 & 74.5\Tstrut\\
        \\[-0.7em]
        \multirow{2}{*}{Shared Features (Ours)} & EfficientNet-B5 \citep{tan2019efficientnet} & 80.5 & 37.9 & 77.0\\
        & EfficientNet-L2 \citep{xie2020self} & 81.7 & 40.7 & 76.6\\
    \end{tabular}
\end{table*}

\begin{table*}[t]
    \centering
    \caption{Comparison between the uni-directional top-down merge \citep{de2021part} and our proposed joint fusion module using various input sources on Cityscapes and Pascal Panoptic Parts \citep{meletis2020cityscapes}}
    \label{tab:fusion}
    \resizebox{\linewidth}{!}{
    \begin{tabular}{lll|ccc|cccc|c}
        \multirow{3}{*}{Model} & \multirow{3}{*}{Backbone} & \multirow{2}{*}{Merging/} & \multicolumn{3}{c|}{Before Merge/Fusion} & \multicolumn{4}{c|}{After Merge/Fusion} & \multirow{2}{*}{Density}\\
         & & \multirow{2}{*}{Fusion} & Sem. & Inst. & Part & \multicolumn{3}{c}{PartPQ} & \multirow{2}{*}{PWQ} & \multirow{2}{*}{[\%]}\\
         & & & mIoU & AP & mIoU & All & P & NP &\Bstrut\\
         
        \hline
        \multicolumn{11}{c}{Cityscapes Panoptic Parts, Single-Scale}\Tstrut\Bstrut\\
        \hline
        \multirow{4}{*}{JPPF} & \multirow{2}{*}{EfficientNet-B5 \citep{tan2019efficientnet}} & Top-Down  & \multirow{2}{*}{80.5} & \multirow{2}{*}{37.9} & \multirow{2}{*}{77.0} & 59.5 & 47.5 & 63.7 & 66.0 & 99.1\Tstrut\\ 
         &  & JPPF &  &  &  & 59.6 & 47.7 & 63.8 & 66.1 & 99.3\\
         & \multirow{2}{*}{EfficientNet-L2 \citep{xie2020self}} & Top-Down & \multirow{2}{*}{81.7} & \multirow{2}{*}{40.7} & \multirow{2}{*}{76.6} & 61.1 & 48.7 & 65.5 & 67.1 & 99.3\\
         &  & JPPF &  &  &  & 61.4 & 49.5 & 65.7 & 67.3 & 99.5\\
        
        \hline
        \multicolumn{11}{c}{Cityscapes Panoptic Parts, Multi-Scale}\Tstrut\Bstrut\\
        \hline
        \multirow{4}{*}{JPPF} & \multirow{2}{*}{EfficientNet-B5 \citep{tan2019efficientnet}} & Top-Down & \multirow{2}{*}{81.8} & \multirow{2}{*}{41.3} & \multirow{2}{*}{78.5} & 61.6 & 50.7 & 65.5 & 68.2 & 99.2\Tstrut\\
         & & JPPF &  &  &  & 61.8 & 50.8 & 65.7 & 68.2 & 99.5\\
         & \multirow{2}{*}{EfficientNet-L2 \citep{xie2020self}} & Top-Down & \multirow{2}{*}{80.0} & \multirow{2}{*}{40.3} & \multirow{2}{*}{76.3} & 62.7 & 50.7 & 67.0 & 67.8 & 99.2\\
         & & JPPF &  &  &  & 63.3 & 52.2 & 67.2 & 68.2 & 99.5\\
        
        \hline
        \multicolumn{11}{c}{Pascal Panoptic Parts, Single-Scale}\Tstrut\Bstrut\\
        \hline
        \multirow{4}{*}{JPPF} & \multirow{2}{*}{EfficientNet-B5 \citep{tan2019efficientnet}} & Top-Down & \multirow{2}{*}{46.0} & \multirow{2}{*}{39.1} & \multirow{2}{*}{54.4} & 29.0 & 37.8 & 26.0 & 42.3 & 89.6\Tstrut\\
         & & JPPF &  &  &  & 32.3 & 48.3 & 26.9 & 45.6 & 92.1\\
         & \multirow{2}{*}{EfficientNet-L2 \citep{xie2020self}} & Top-Down & \multirow{2}{*}{52.3} & \multirow{2}{*}{47.3} & \multirow{2}{*}{62.2} & 30.6 & 43.6 & 26.2 & 49.2 & 92.7\\
         & & JPPF &  &  &  & 40.5 & 58.5 & 34.4 & 53.7 & 92.7\\
        
    \end{tabular}
    }
\end{table*}

\subsection{A Single Shared Encoder} \label{sec:results:encoder}
As part of our contribution, we aim to unify semantic, instance, and part segmentation and jointly learn all three in a single, unified model.
We are validating that these three tasks benefit from a shared feature representation by comparing the individual predictions before fusion to three separate equivalent networks that have been trained individually with different encoders.
As shown in \cref{tab:encoder}, both models with a single, shared encoder surpass the individual models in all three tasks.
To no surprise, the more recent version of EfficientNet \citep{tan2019efficientnet} produces already better initial results for our fusion.
This experiment clearly indicates that using a shared encoder enables the learning of a common feature representation, resulting in more accurate individual outcomes of each head, which are also more consistent by design due to the shared representation.

\subsection{Joint Fusion} \label{sec:results:fusion}
Next, we compare our joint fusion module to the previously presented top-down merging strategy \citep{de2021part} in \cref{tab:fusion}.
It is important to note, that even the recently published state-of-the-art method PPF \citep{li2022panoptic, li2023panopticpartformer} uses this merging strategy. 
The proposed fusion module surpasses the top-down merge in terms of PartPQ, $\text{PartPQ}_\text{P}$, $\text{PartPQ}_\text{NP}$, and PWQ on all datasets and settings.
Even though our proposed fusion is admittedly only slightly better in some cases, the joint fusion produces also denser results than the uni-directional merge, indicating the improved consistency.
The advantages of our fusion are mainly reflected for the results in areas that are partitionable.
Since the \things{} with part labels are limited in CPP, the impact is best observed on the PPP dataset.
On this data, our proposed fusion module is significantly better.
Specifically for our design, $\text{PartPQ}_\text{P}$ is improved by 10.5 pp and 14.9 pp for the different backbones.
That is a relative improvement of about 28~\% and 44~\%.

\begin{table*}[t]
    \centering
    \caption{Detailed run-time analysis of our JPPF and the baseline on full resolution images of  Cityscapes Panoptic Parts usisng a Nvidia A100 GPU. $^\ast$indicates the reproduced baseline which is detailed in \cref{sec:results:sota}}
    \label{tab:runtime}
    \resizebox{\linewidth}{!}{%
    \begin{tabular}{cc|c|c|ccc|c}
        \multirow{3}{*}{Method} & \multirow{3}{*}{Backbone} & Feature & Individual & \multicolumn{3}{c|}{Fuse/Merge [ms]} & Total\\ 
         & & Extraction & Predictions & Panoptic & \multirow{2}{*}{Merge} & \multirow{2}{*}{JPPF} & Inference\\
         & & [ms] & [ms] & Fusion & & & [ms]\Bstrut\\ 
        \hline
        Baseline$^\ast$ & EfficientNet-B5 \citep{tan2019efficientnet} & 52 & 202 & 118  & 484  & --  & 856\Tstrut\\
        \multicolumn{2}{l|}{} & \multicolumn{1}{c|}{} & \multicolumn{1}{c|}{} & \multicolumn{3}{c|}{}\\[-0.7em]
        \multirow{2}{*}{\ours} & EfficientNet-B5 \citep{tan2019efficientnet} & 26 & 202 & --  & -- & 161 & 389\\
         & EfficientNet-L2 \citep{xie2020self} & 59 & 202 & --  & -- & 161 & 422
    \end{tabular}
    }
\end{table*}

\begin{table}[t]
    \centering
    \caption{Comparison of model complexity for an input of size $1200 \times 800$ pixels}
    \label{tab:complexity}
    \begin{tabular}{ll|cc}
        \multirow{2}{*}{Method} & \multirow{2}{*}{Backbone} & Params & FLOPs\\
         & & [M] & [G]\Bstrut\\
        \hline
        \multirow{2}{*}{PPF \citep{li2022panoptic}} & ResNet50 \citep{he2016deep} & \phantom{1}37.4 & 185.8\Tstrut\\
         & Swin \citep{liu2021swin} & 100.3 & 408.5\\
        \multicolumn{2}{l|}{}\\[-0.7em]
        \multirow{2}{*}{PPF++ \citep{li2023panopticpartformer}} & ResNet50 \citep{he2016deep} & \phantom{1}45.6 & 215.4 \\
         & ConvNext \citep{liu2022convnet} & 120.2 & 519.5 \\
        \multicolumn{2}{l|}{}\\[-0.7em]
        \multirow{2}{*}{\ours} & Eff.Net-B5 \citep{tan2019efficientnet} & \phantom{1}44.2 & 211.6 \\
         & Eff.Net-L2 \citep{xie2020self} & 406.2 & 889.7
    \end{tabular}
\end{table}

\subsection{Density, Run Time, and Model Size} \label{sec:results:complexity}

Our JPPF produces results, which are at least as dense as the top-down merging (see \cref{tab:fusion}).
We further assessed the inference time of our proposed model with JPPF, and the results are displayed in \cref{tab:runtime}.
It is evident that the top-down merging requires more than twice the time compared to our proposed fusion.
To obtain panoptic-part segmentation as proposed by \citet{de2021part}, one must first perform a panoptic fusion and then combine it with the part segmentation, which adds an extra overhead.
\Cref{tab:complexity} shows that our approach, in terms of model size and number of floating point operations (FLOPs), is comparable to previous work for the smaller backbones (ResNet50 \cite{he2016deep} and EffificentNet-B5 \cite{tan2019efficientnet}).
Our updated backbone (EfficientNet-L2 \cite{xie2020self}) is significantly larger and more complex than the initial version, but adds only little overhead in terms of run time (see \cref{tab:runtime}).

\begin{figure*}
    \centering

    \begin{subfigure}{0.495\linewidth}
        \includegraphics[width=\linewidth]{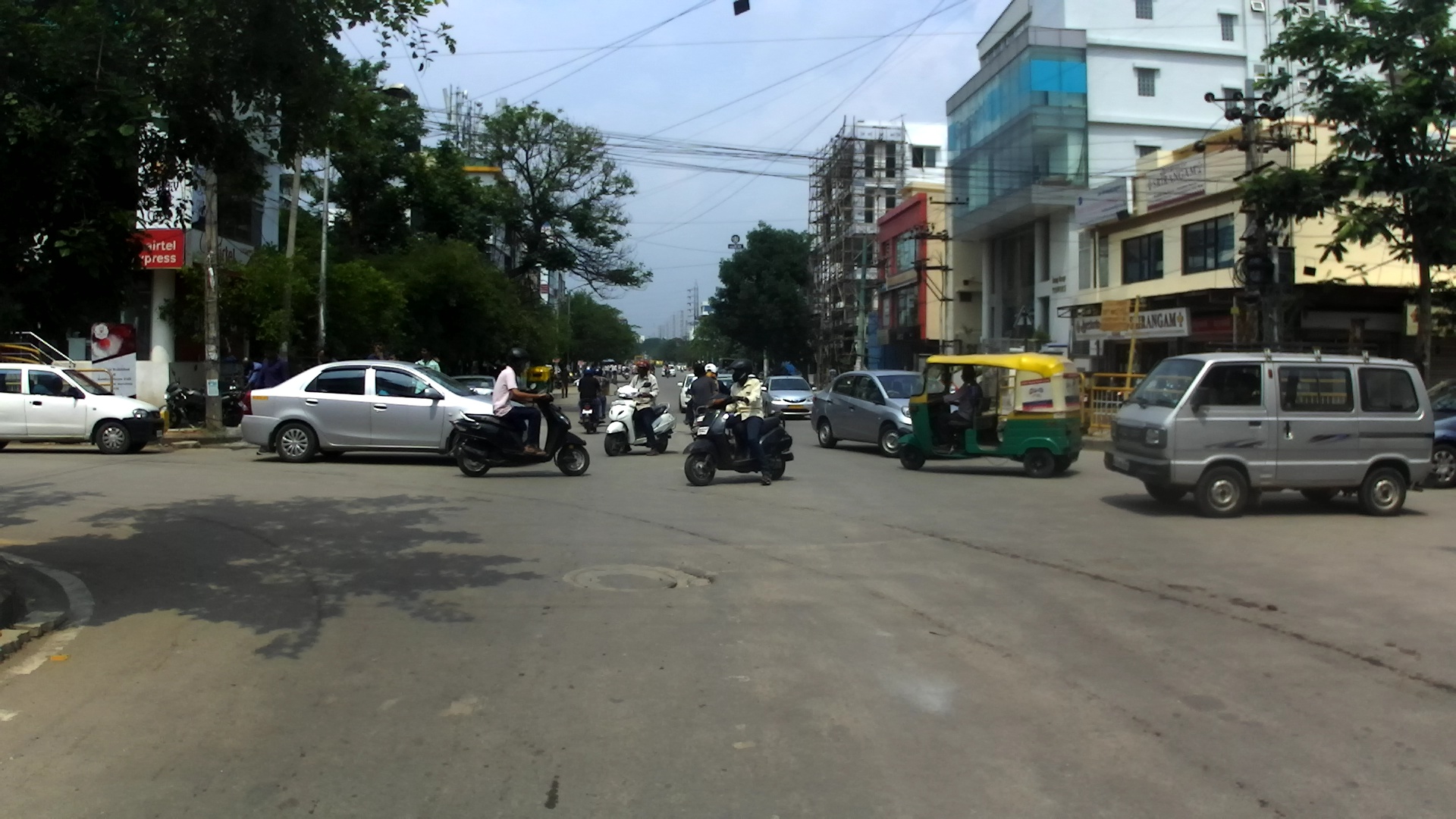}
    \end{subfigure}\hspace*{\fill}
    \begin{subfigure}{0.495\linewidth}
        \includegraphics[width=\linewidth]{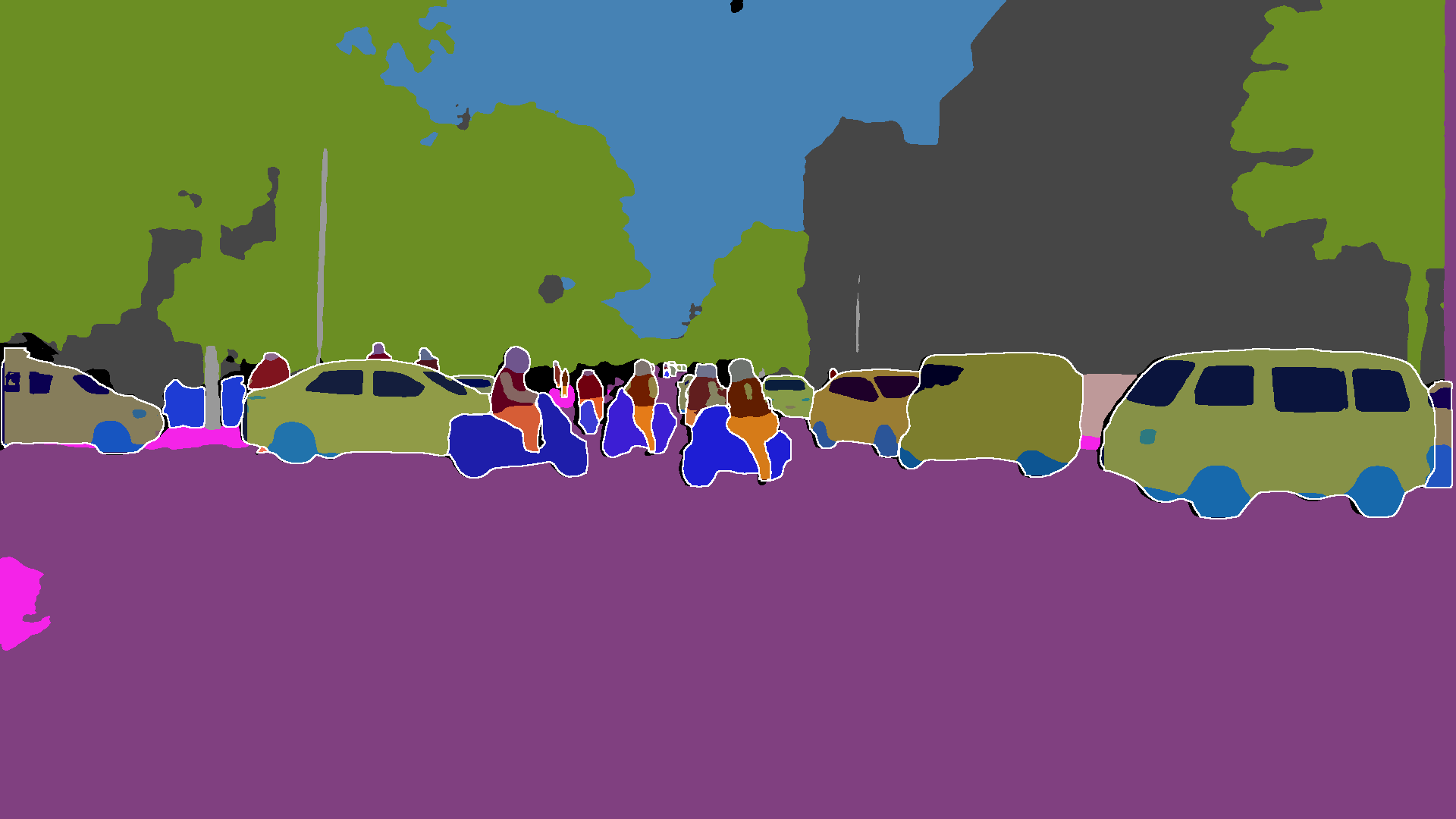}
    \end{subfigure}
    
    \begin{subfigure}{0.495\linewidth}
        \includegraphics[width=\linewidth]{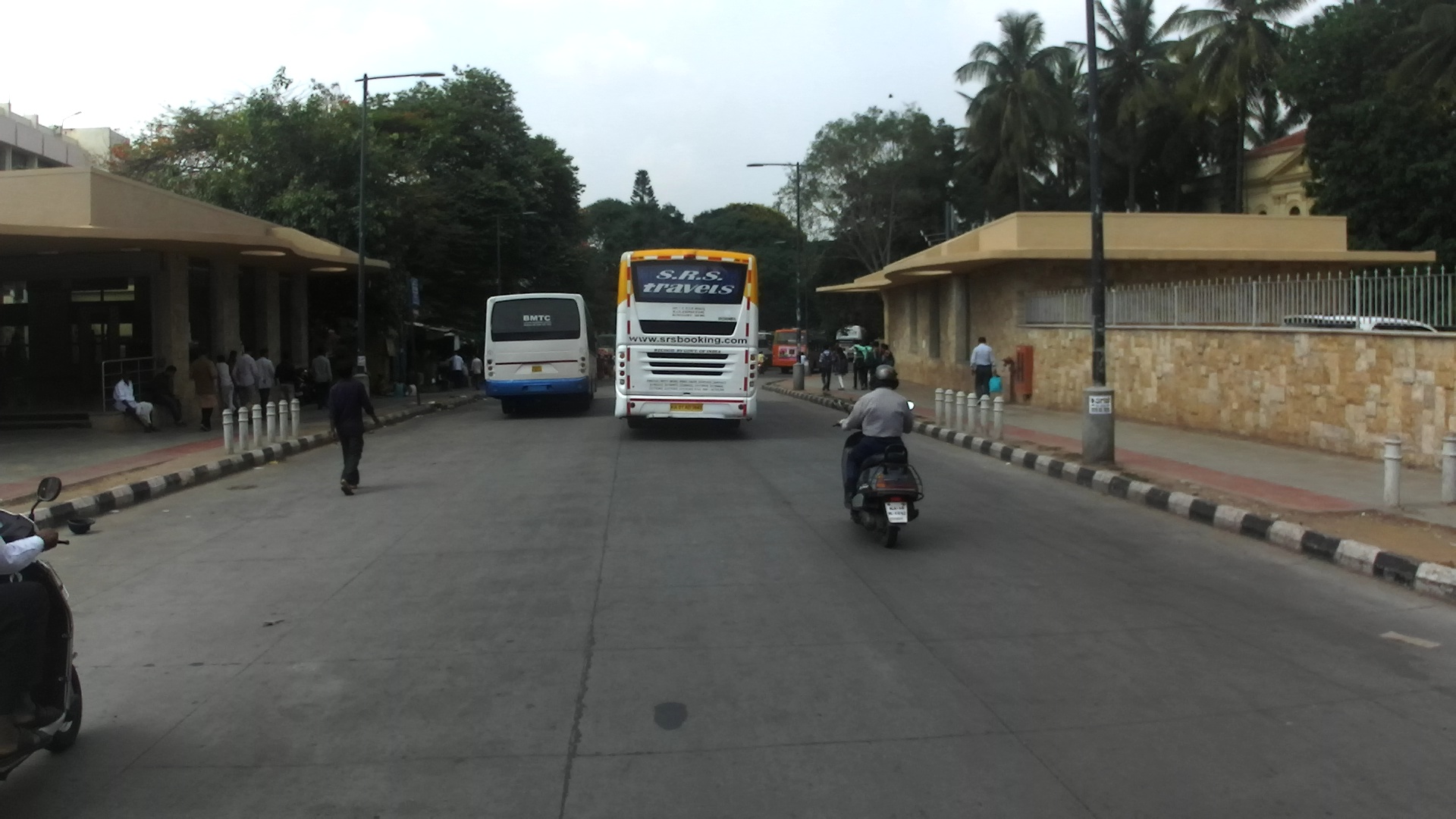}
        \subcaption*{Input Images}
    \end{subfigure}\hspace*{\fill}
    \begin{subfigure}{0.495\linewidth}
        \includegraphics[width=\linewidth]{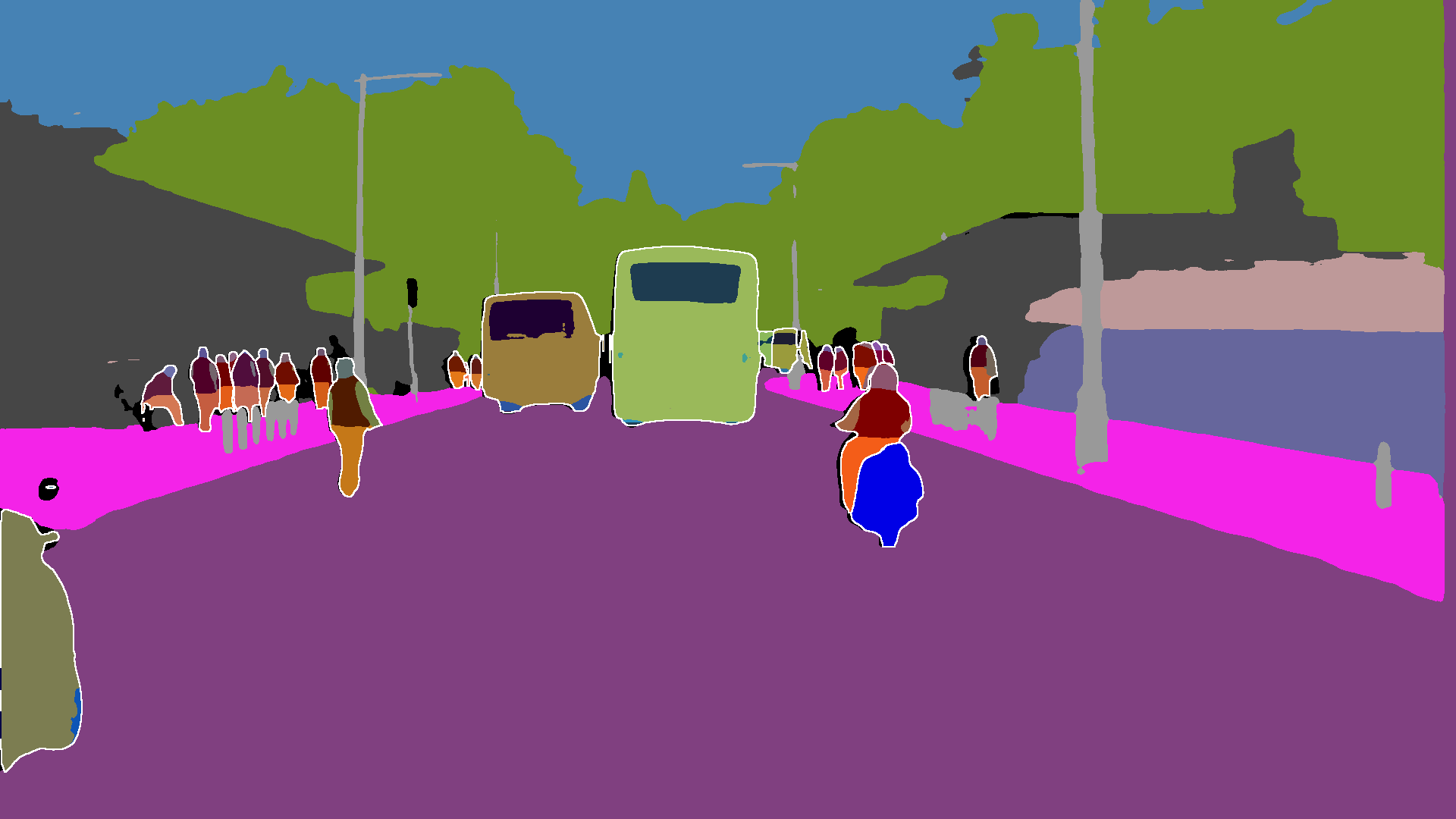}
        \subcaption*{Results of JPPF}
    \end{subfigure}
    
    \caption{Visual results of our JPPF on the Indian Driving Dataset (IDD) \citep{varma2019idd} without fine-tuning}
    \label{fig:generalization}
\end{figure*}

\subsection{Generalization} \label{sec:results:generalization}
Since our fusion module is free of learned parameters (\ie there are no trainable layers involved), it is independent of its input and supposed to exhibit a good generalization to unseen domains.
However, the entire model (including the backbone and individual prediction heads) is restricted by the typical rigidity of deep neural networks and their sensitivity to shifts in the distribution.
Yet, our complete model generalizes well to other datasets, as demonstrated in \cref{fig:generalization} even though it has only been trained on CPP for this experiment.
We show the generalization for a more extensive set of various other datasets without fine-tuning in \cref{fig:appendix:bdd,fig:appendix:acdc,fig:appendix:kitti,fig:appendix:vistas}, including a typical failure case.
A qualitative comparison in terms of generalization between PPF++ \cite{li2023panopticpartformer} and our method is provided in \cref{fig:appendix:generalization}.
For all our results, we have resized the input images to fit the size of the original CPP dataset, \ie $1024 \times 2048$ pixels.

\section{Limitations} \label{sec:limitations}
During the thorough evaluation of our approach, we have identified some remaining limitations, which we discuss here.
Though, our fusion operation treats the logits of the three prediction heads equally, the overall process is mainly guided by the prediction of the instance branch.
\Ie the detected \things{} and their classes and bounding boxes control the information flow during fusion, mostly.
With respect to balance and importance of the individual predictions, this is a limitation.
Additionally, as a side effect of this fact, the fusion of \things{} is limited to the area within each bounding box.
Thus, for very large objects that are not fully covered by the bounding box, the fusion can not compensate the initially too small estimated area of these objects.
Furthermore, we have identified a theoretical limitation in the fusion operation in \cref{eq:fusion} itself.
Our generalized version is indeed able to handle an arbitrarily sized set of input logits, however there is no explicit mechanism to balance (normalize) the fused output for different numbers of inputs.
In practice, highly confident inputs produce similar highly confident outputs when they are consistent, independent of the number of inputs (\eg 2 or 3).
For less confident areas, the imbalance between the fused \stuff{} (two input logits) and the fused \things{} (three input logits) might be an issue.
Finally, we notice that post-processing step (filtering out small \stuff{} areas) is the remaining factor that hinders fully dense predictions, \ie a valid (not necessarily correct) panoptic-part label for every pixel of the input image.

\section{Conclusion} \label{sec:conclusion}
JPPF is a versatile fusion operation that combines semantic, instance, and part segmentation effectively into a consistent panoptic-part segmentation.
It consistently outperforms uni-directional top-down merging for various input sources, \eg our previous and updated model.
Our design with the updated backbone and joint fusion module surpasses the baseline on all datasets, achieves state-of-the-art results on Cityscapes Panoptic Parts, and ranks in between the first and second versions of the Panoptic-PartFormer on Pascal Panoptic Parts.
The advantages of our proposed approach become most visible for partitionable areas.
The increased consistency in the prediction of our model is highlighted by its increased density.
We leave it for future work to find suitable solutions for the limitations that have been discussed.

\bmhead{Acknowledgments}
This work was partially funded by the Federal Ministry of Education and Research Germany under the project DECODE (01IW21001).

%

\bibliographystyle{sn-basic}
\bibliography{bib}

\newpage
\begin{appendices}

\section{Additional Visualizations}\label{appendix:A}

\begin{figure*}[b!]
    \centering
    \begin{subfigure}{0.245\linewidth}
\includegraphics[width=\linewidth]{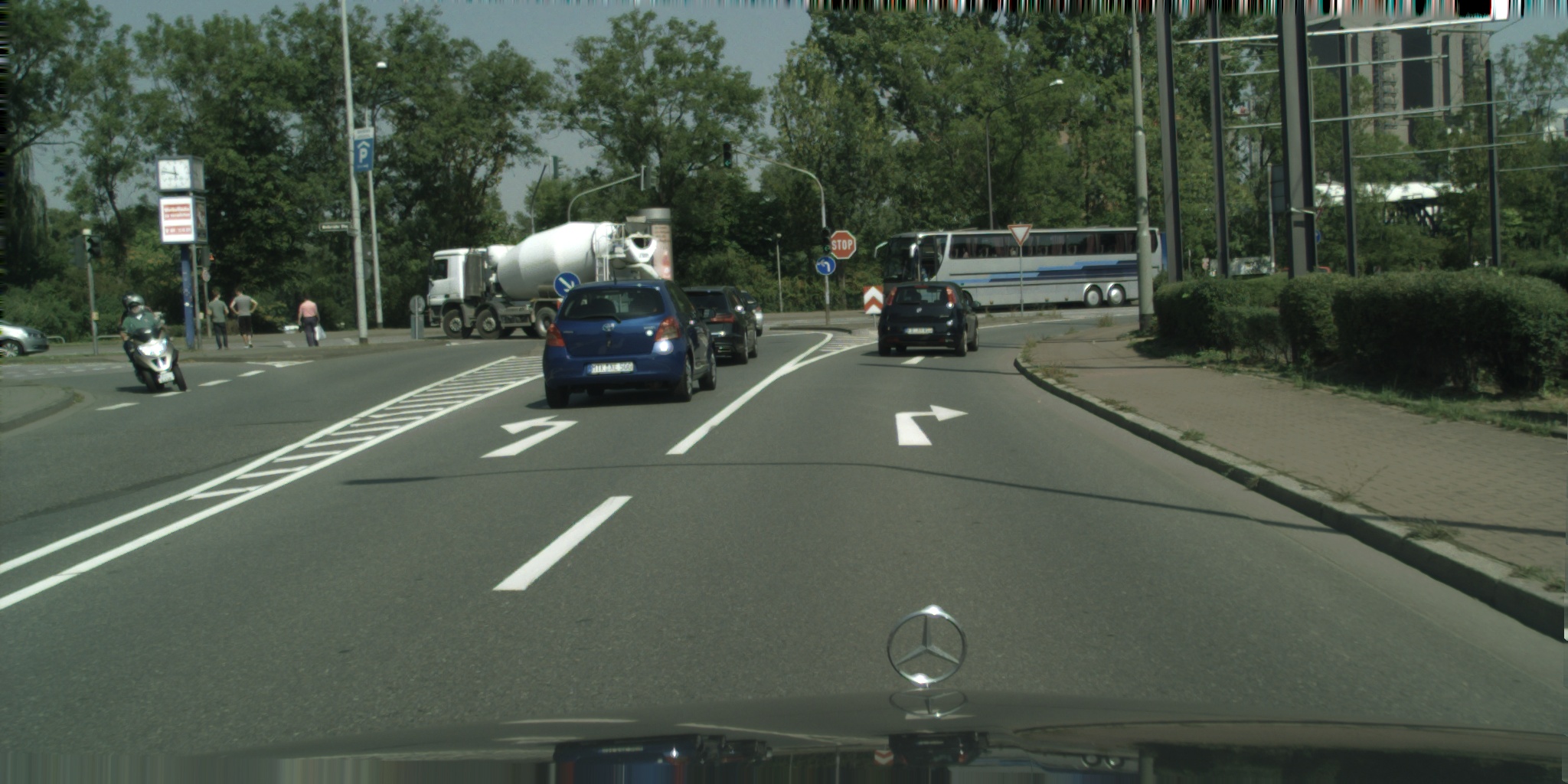}
    \end{subfigure}\hspace*{\fill}
     \begin{subfigure}{0.245\linewidth}
\includegraphics[width=\linewidth]{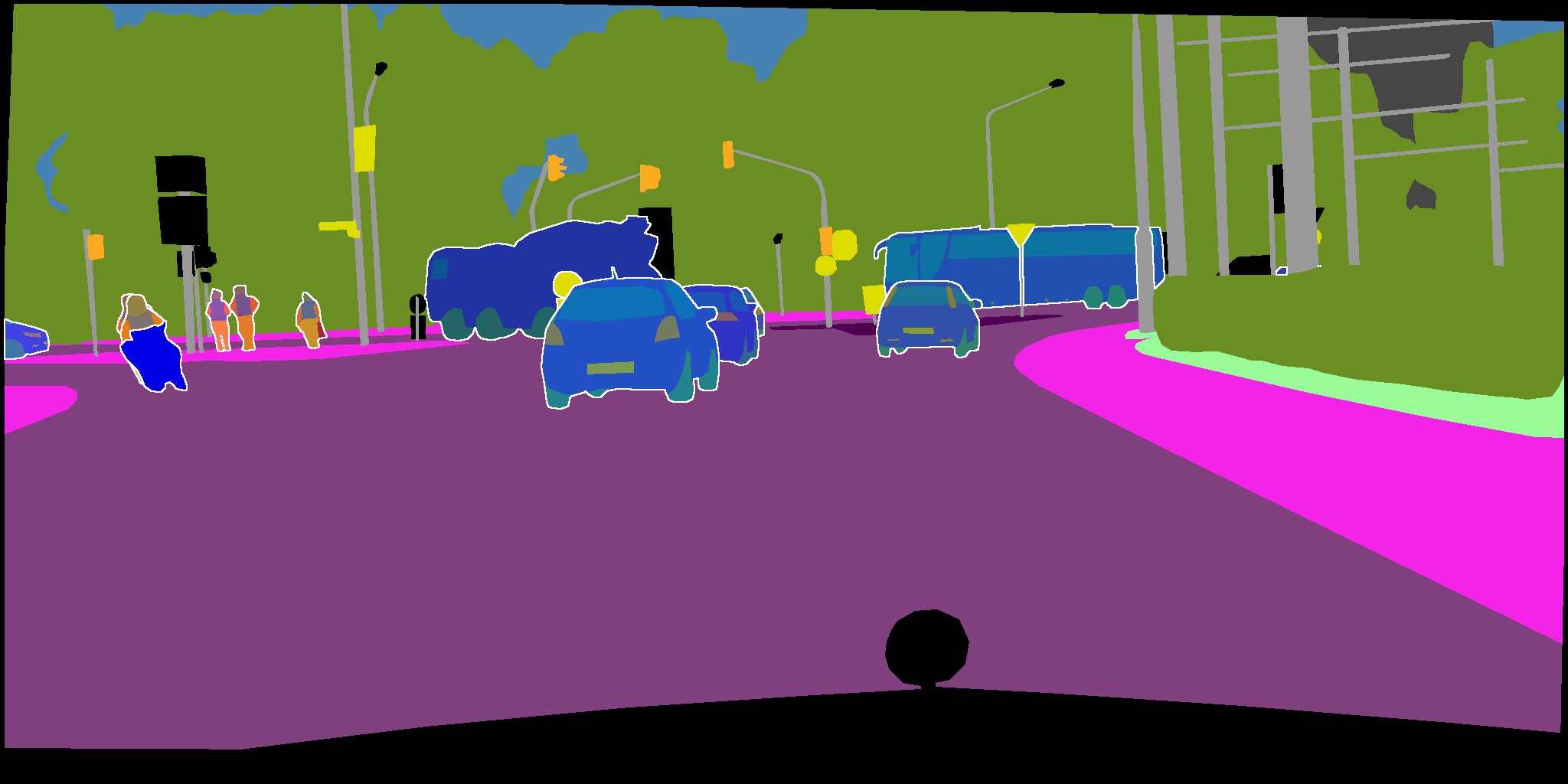}
    \end{subfigure}\hspace*{\fill}
     \begin{subfigure}{0.245\linewidth}
\includegraphics[width=\linewidth]{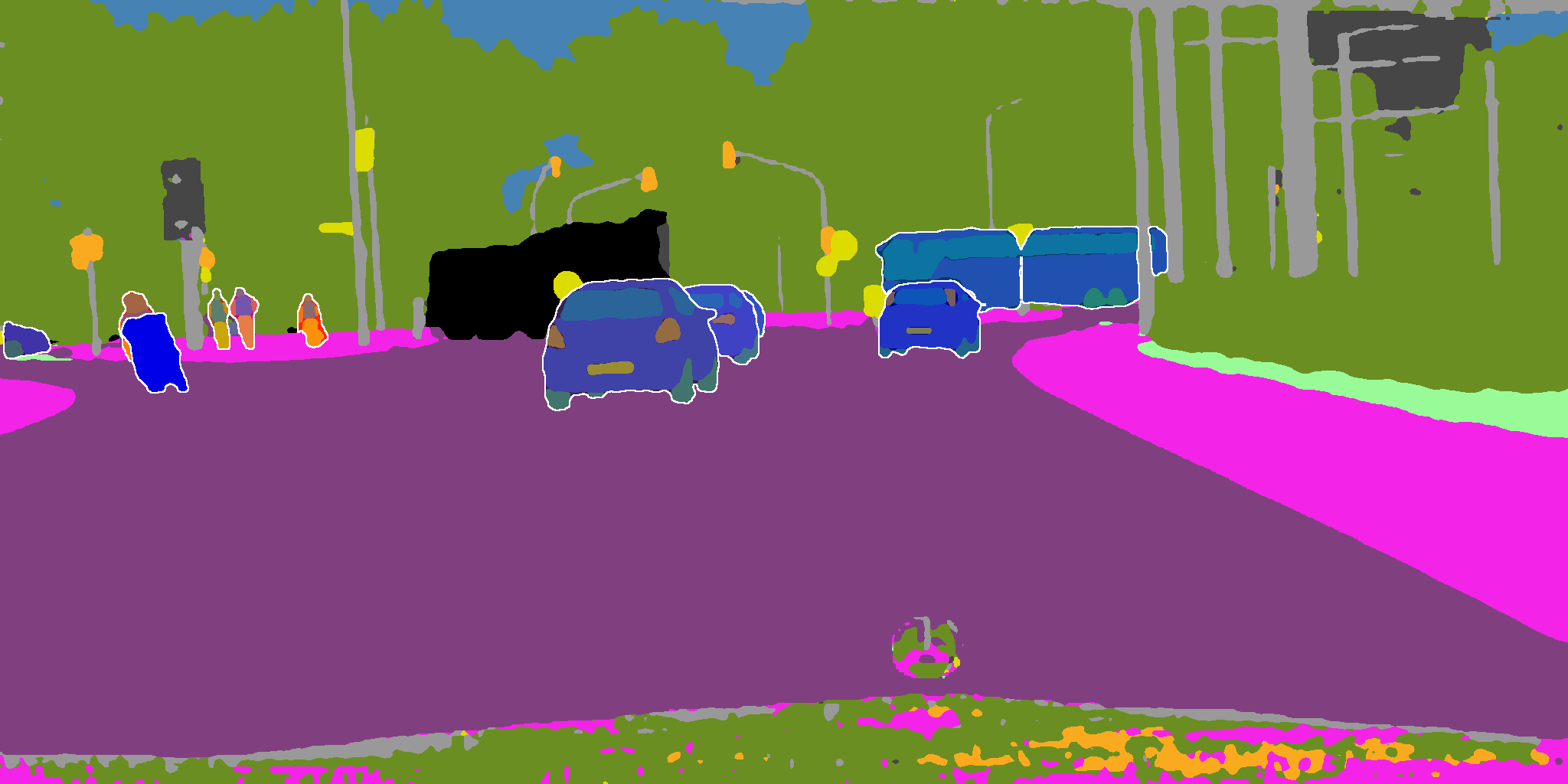}
    \end{subfigure}\hspace*{\fill}
     \begin{subfigure}{0.245\linewidth}
\includegraphics[width=\linewidth]{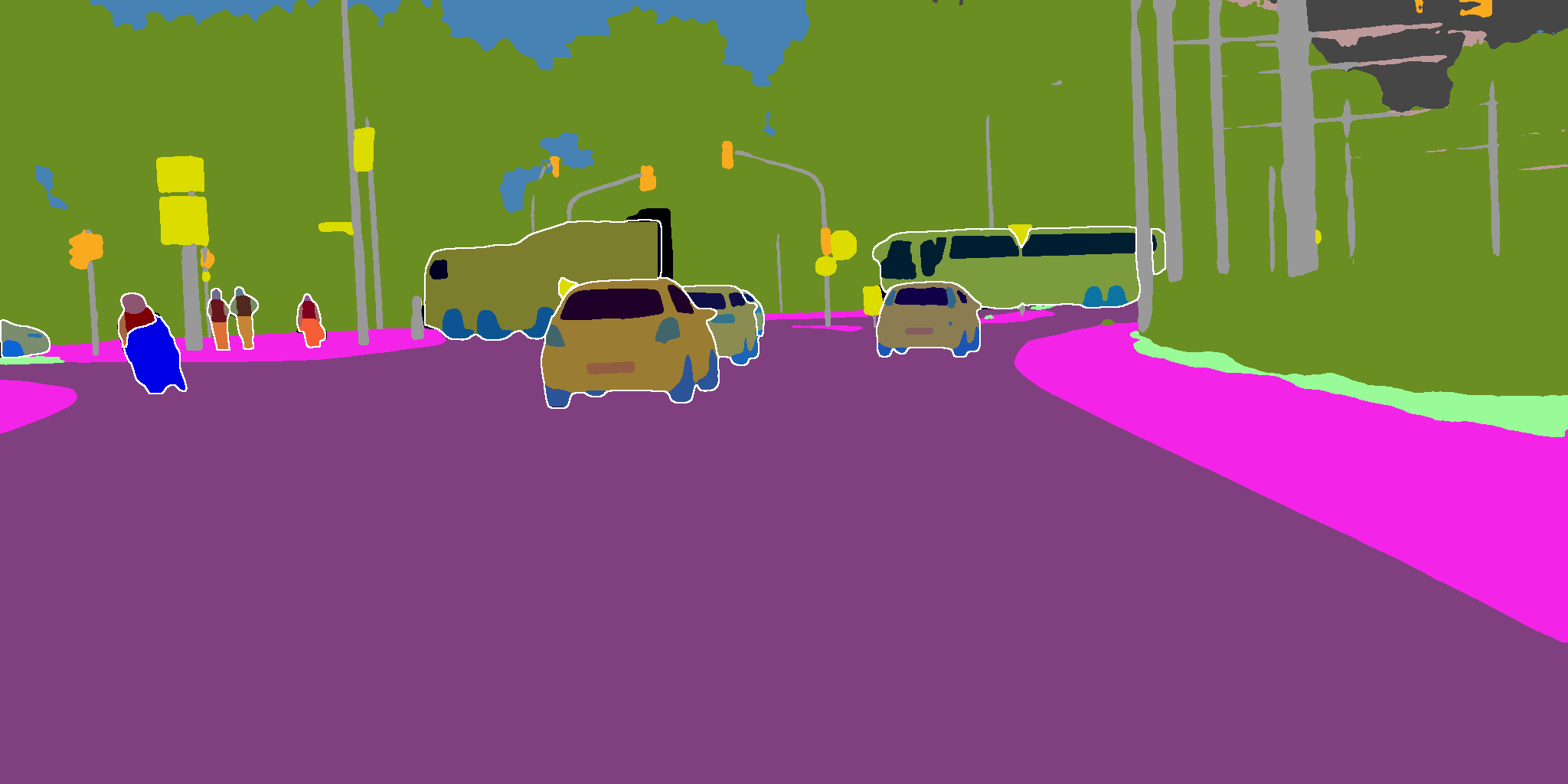}
    \end{subfigure}
    
    \vspace*{\fill}

            \begin{subfigure}{0.245\linewidth}
\includegraphics[width=\linewidth]{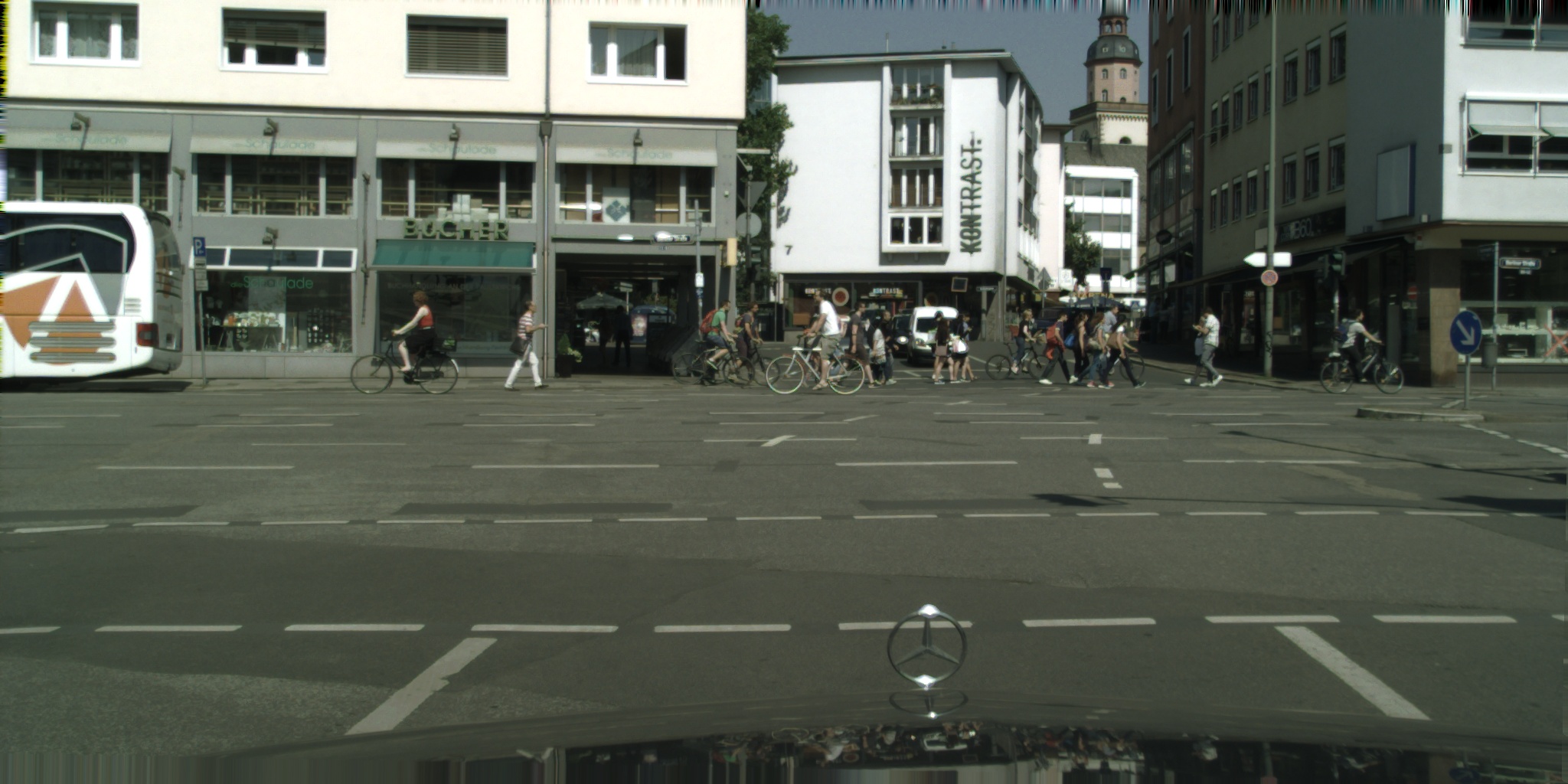}
    \end{subfigure}\hspace*{\fill}
     \begin{subfigure}{0.245\linewidth}
\includegraphics[width=\linewidth]{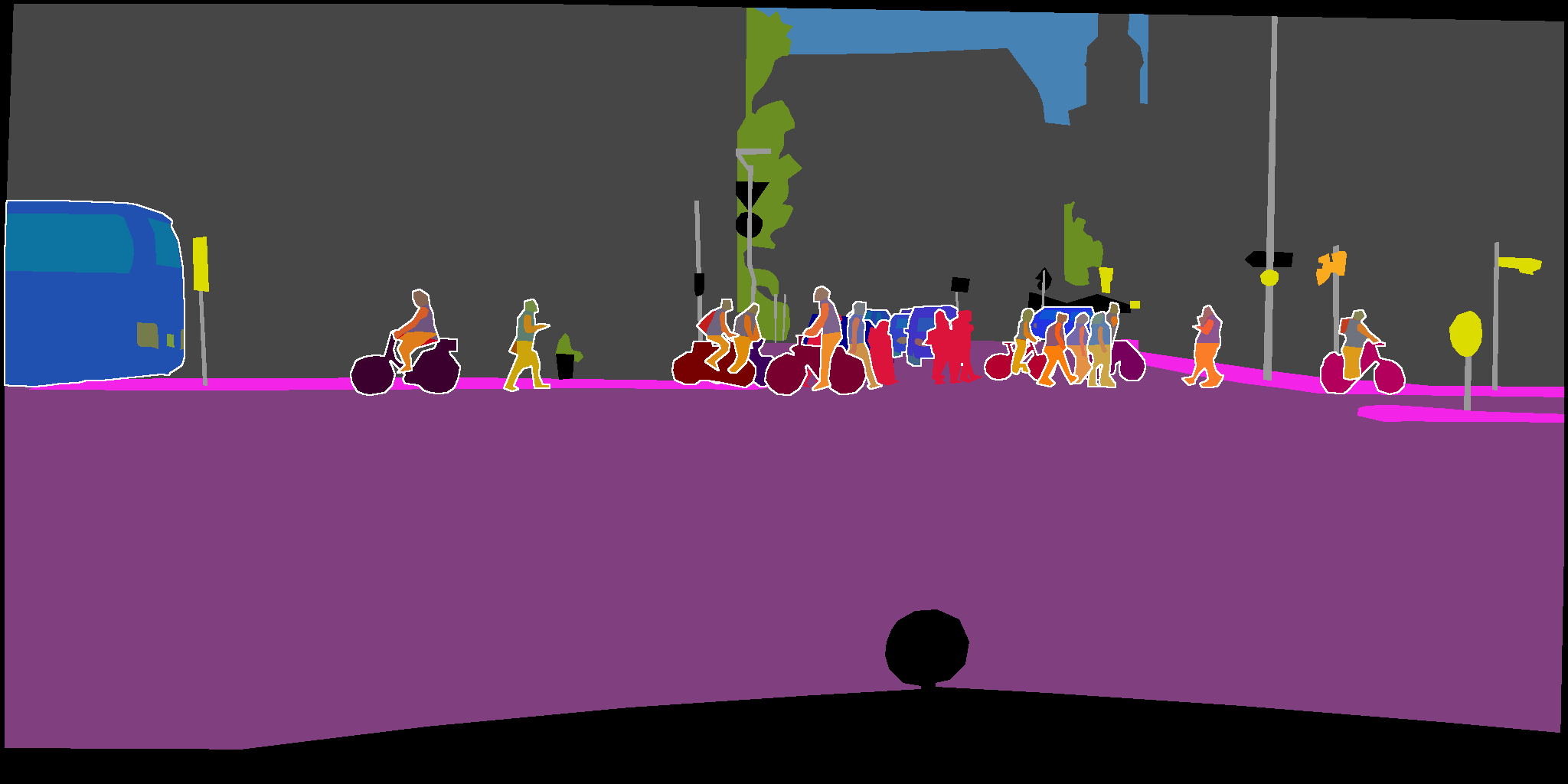}
    \end{subfigure}\hspace*{\fill}
     \begin{subfigure}{0.245\linewidth}
\includegraphics[width=\linewidth]{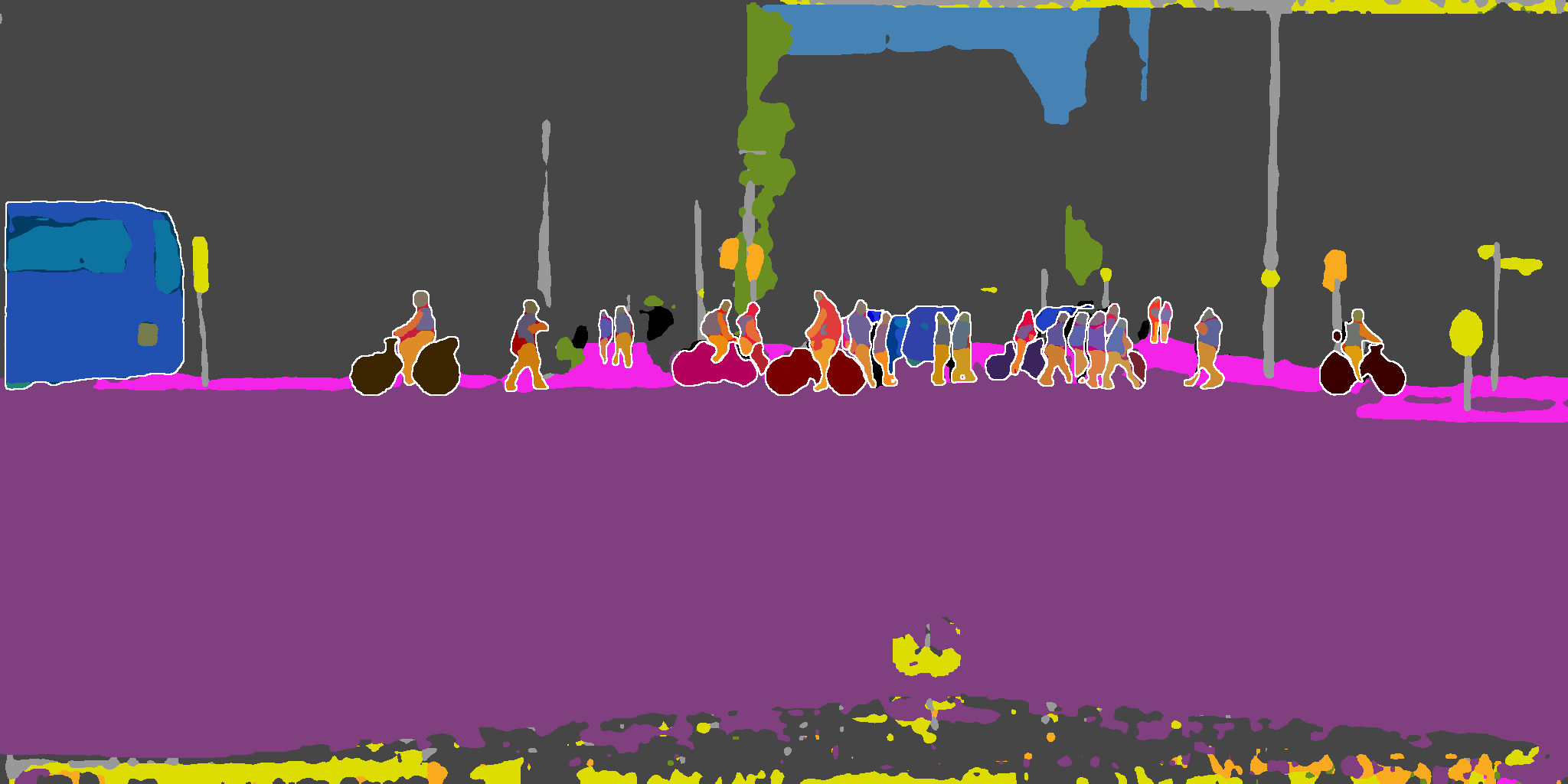}
    \end{subfigure}\hspace*{\fill}
     \begin{subfigure}{0.245\linewidth}
\includegraphics[width=\linewidth]{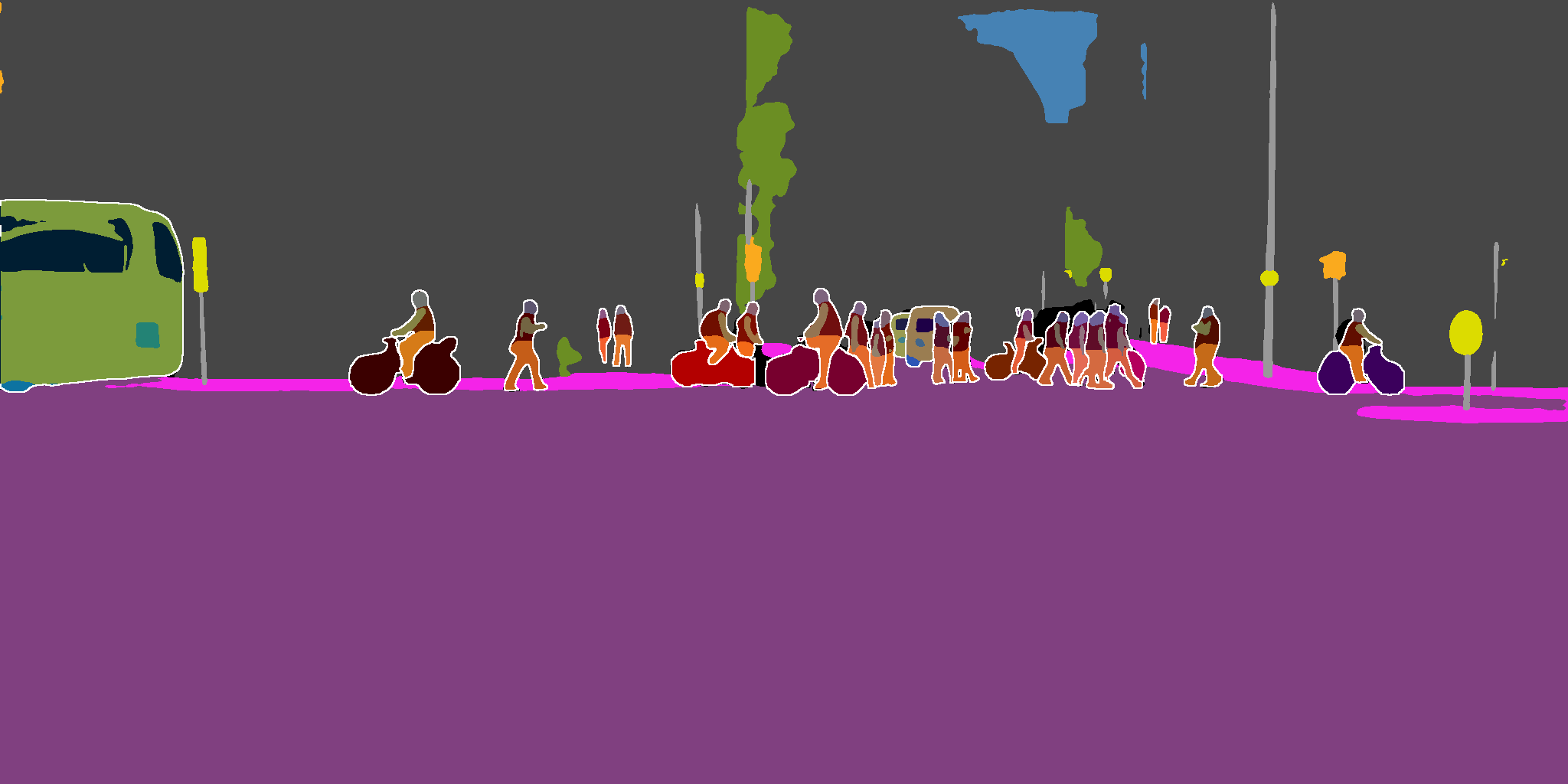}
    \end{subfigure}
    
    \vspace*{\fill}

        \begin{subfigure}{0.245\linewidth}
\includegraphics[width=\linewidth]{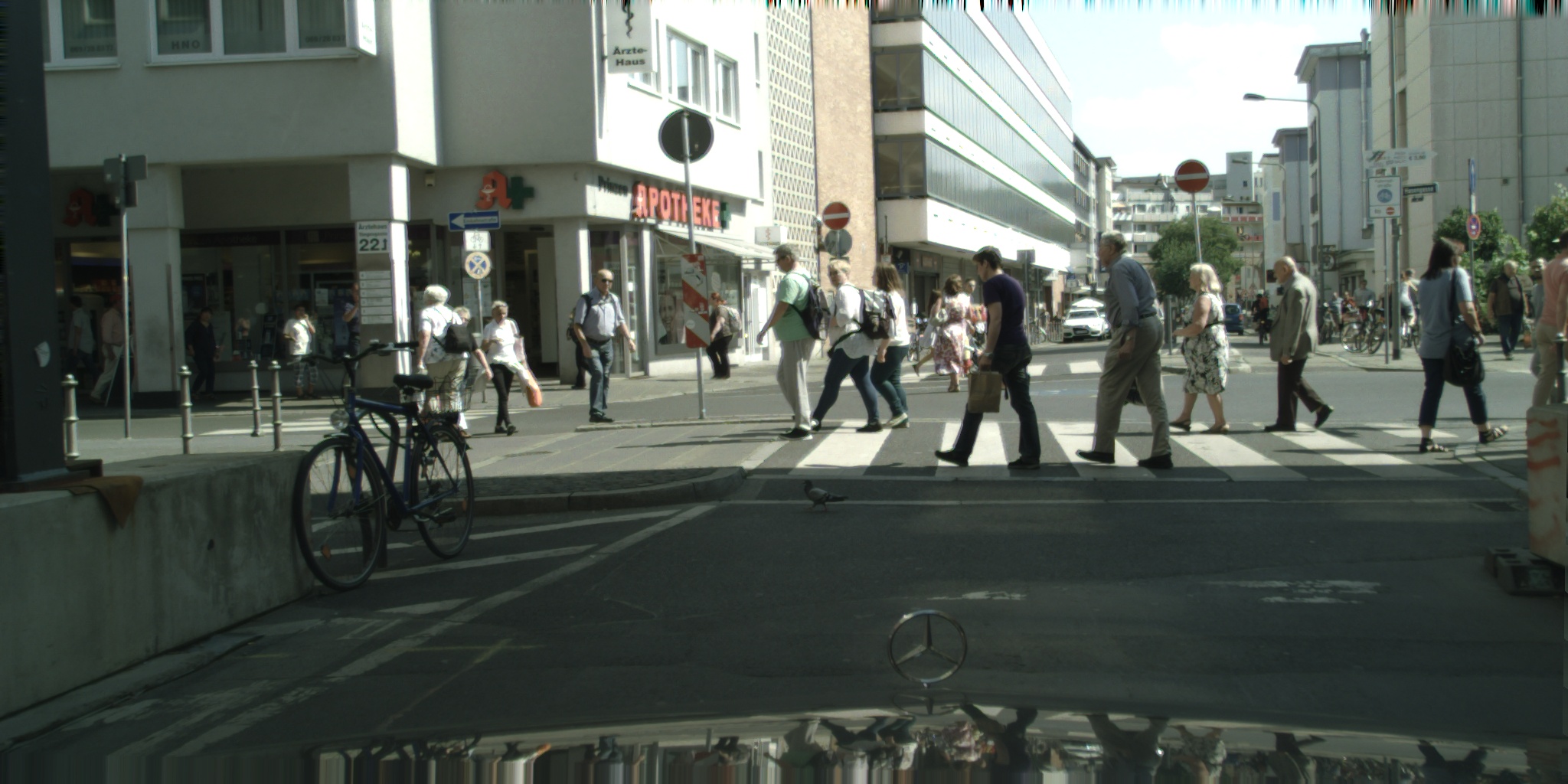}
    \end{subfigure}\hspace*{\fill}
     \begin{subfigure}{0.245\linewidth}
\includegraphics[width=\linewidth]{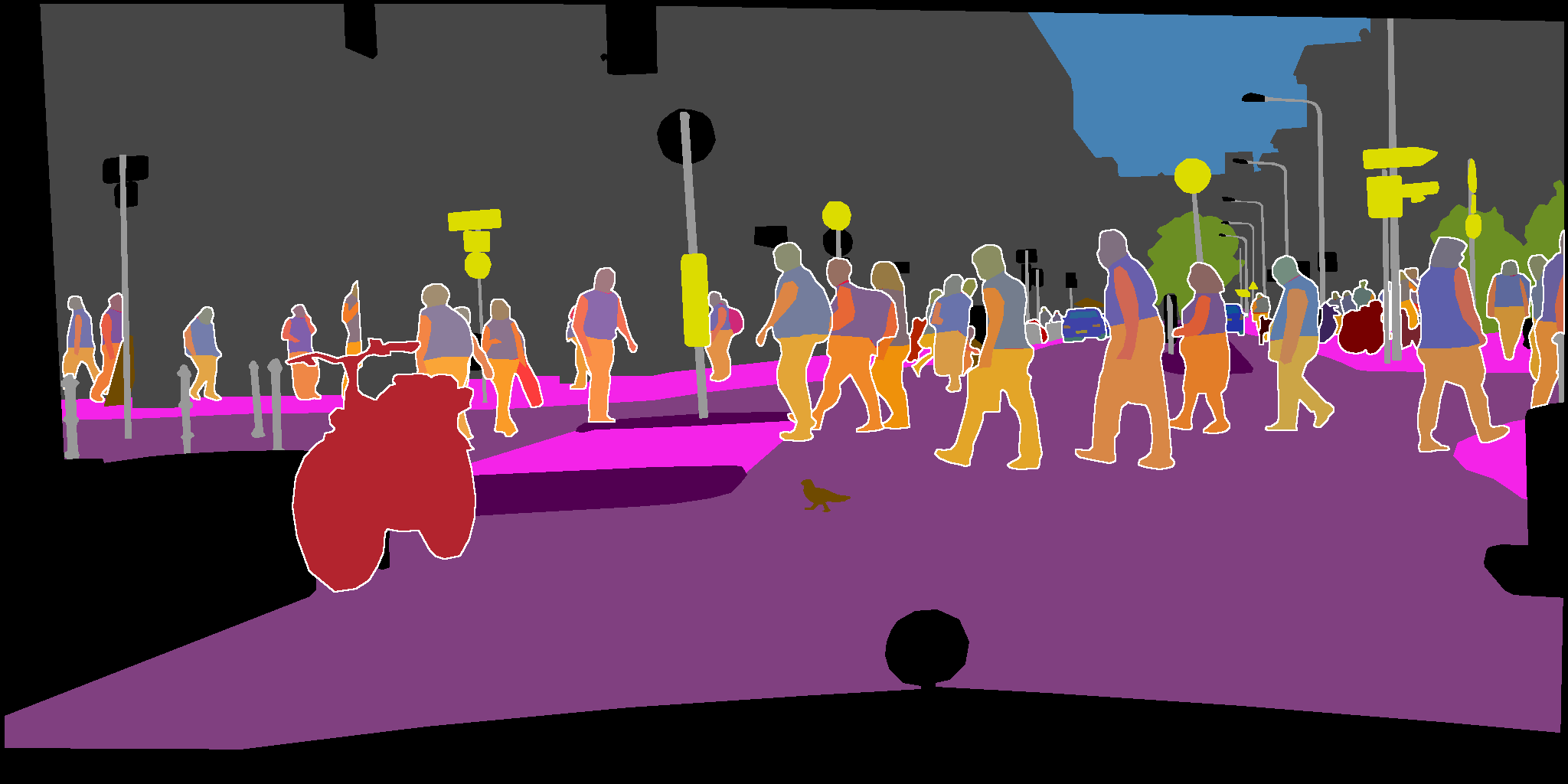}
    \end{subfigure}\hspace*{\fill}
     \begin{subfigure}{0.245\linewidth}
\includegraphics[width=\linewidth]{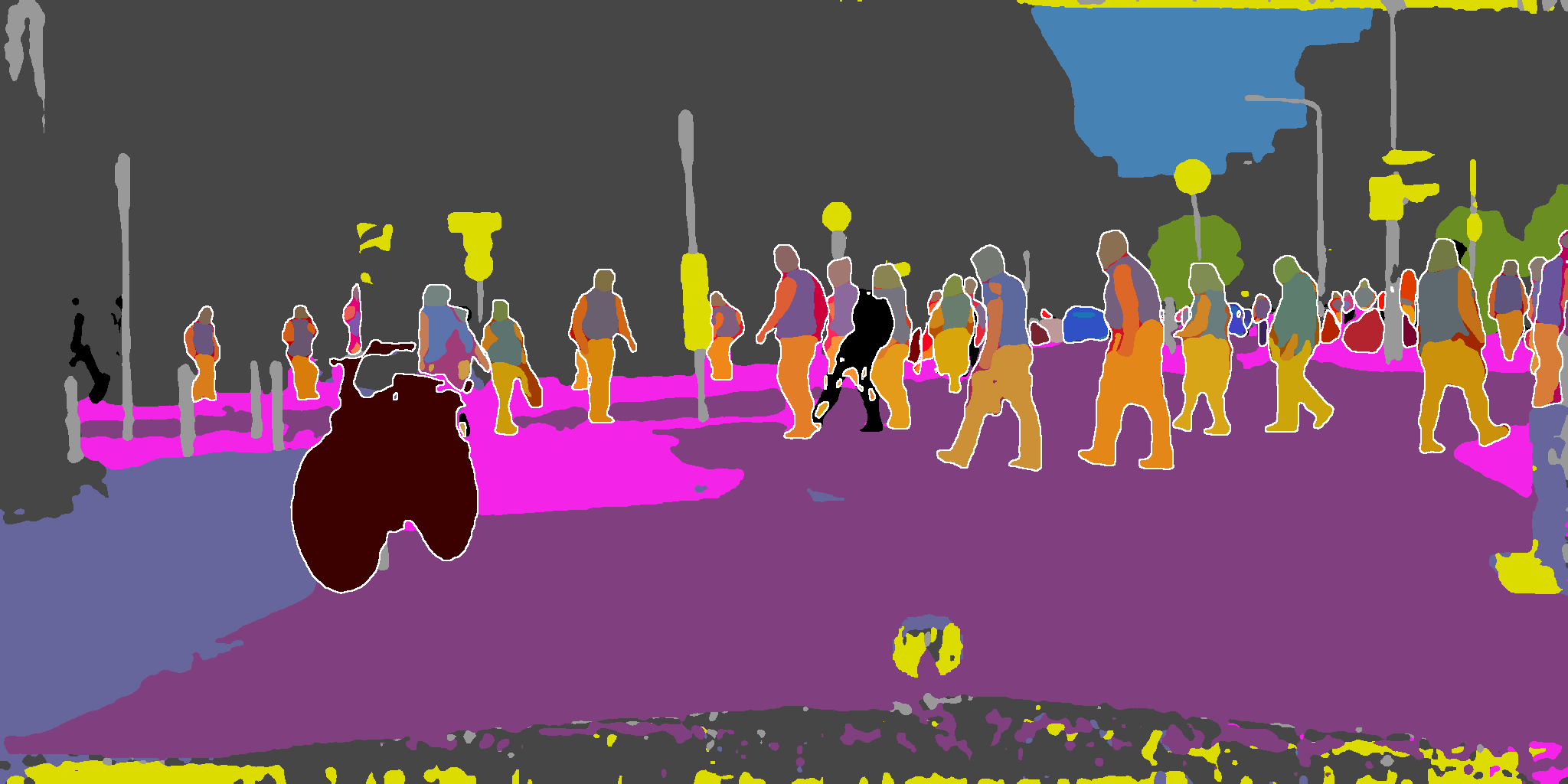}
    \end{subfigure}\hspace*{\fill}
     \begin{subfigure}{0.245\linewidth}
\includegraphics[width=\linewidth]{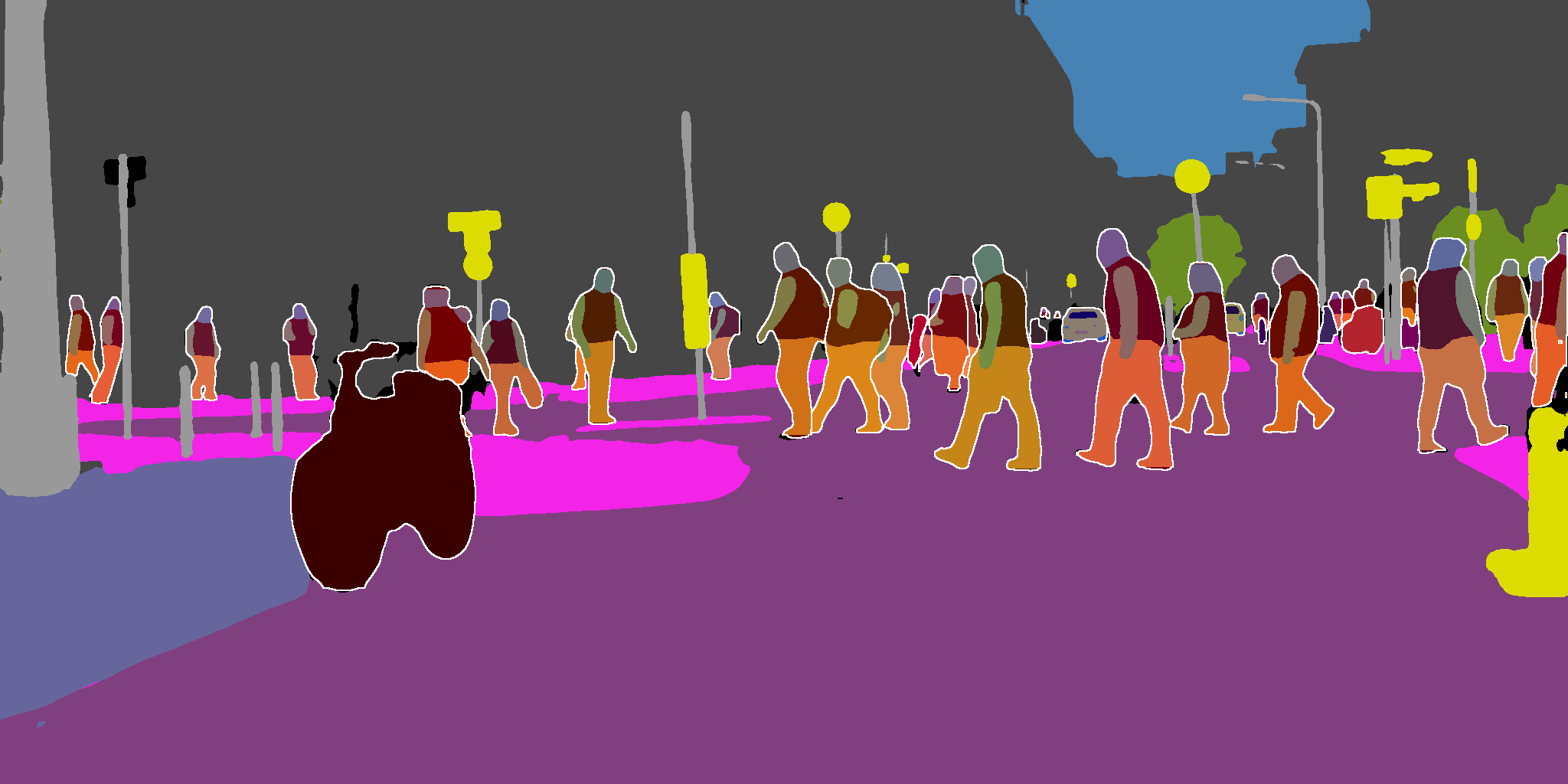}
    \end{subfigure}
    
    \vspace*{\fill}
    
        \begin{subfigure}{0.245\linewidth}
\includegraphics[width=\linewidth]{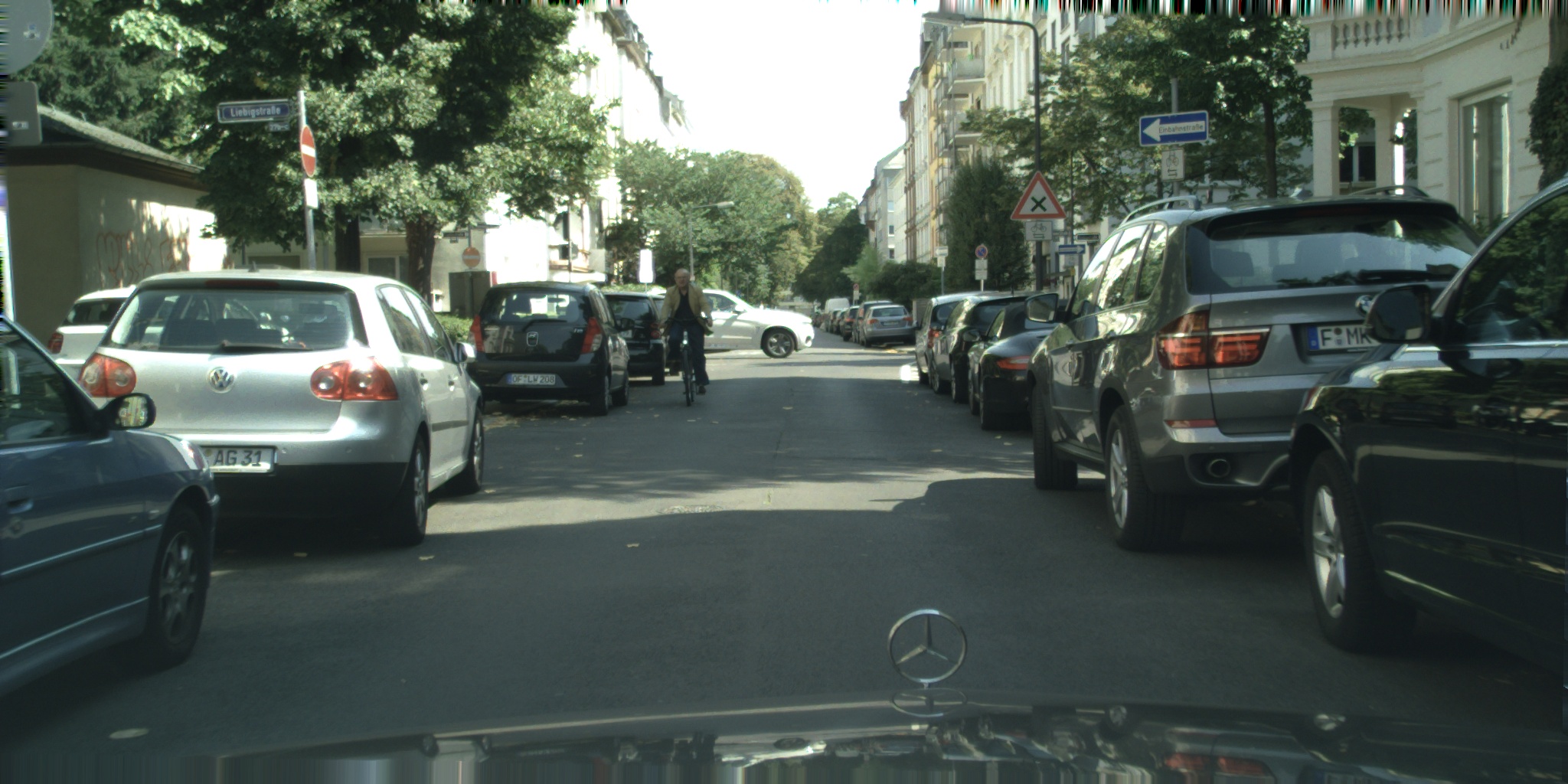}
    \end{subfigure}\hspace*{\fill}
     \begin{subfigure}{0.245\linewidth}
\includegraphics[width=\linewidth]{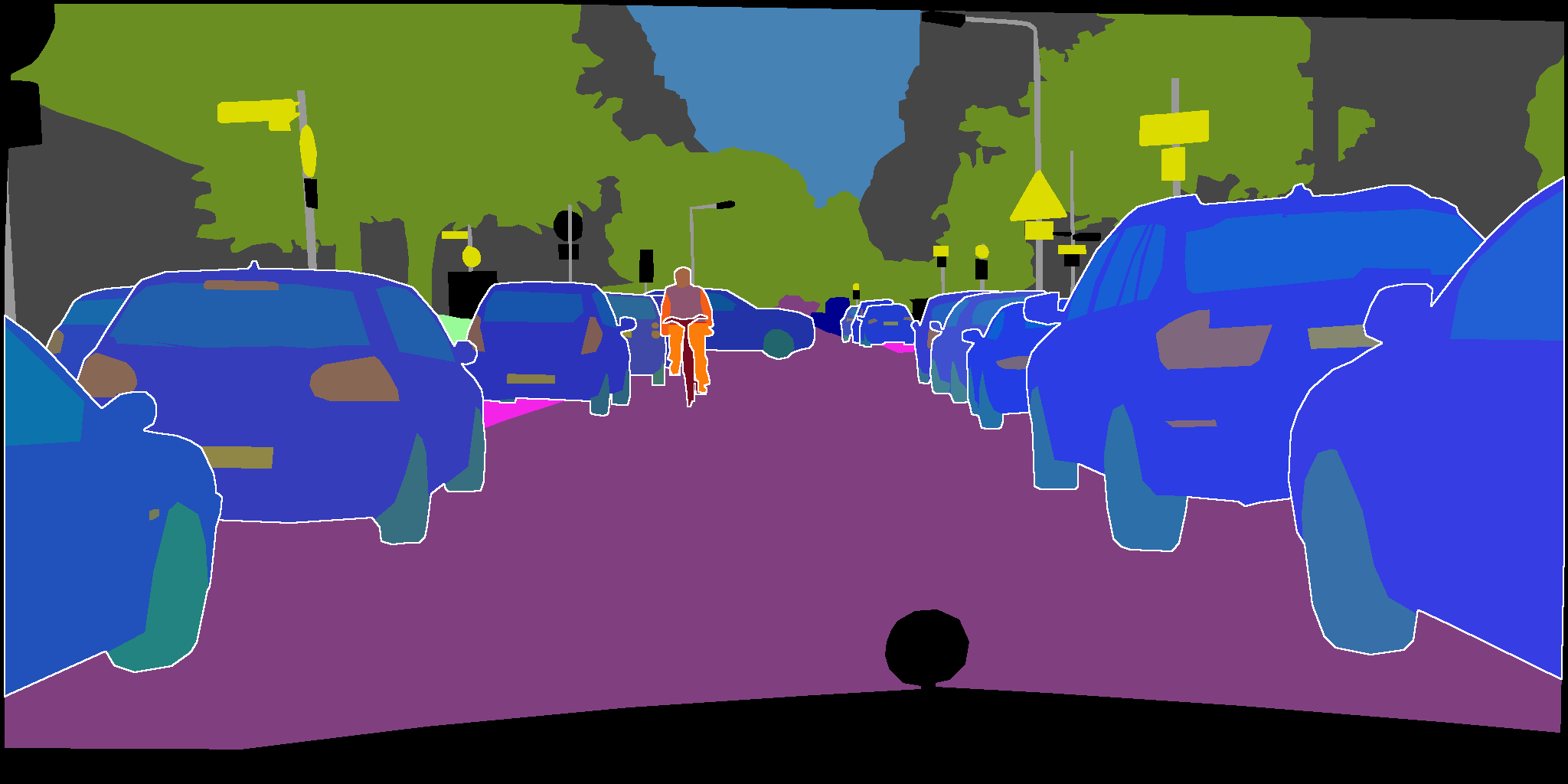}
    \end{subfigure}\hspace*{\fill}
     \begin{subfigure}{0.245\linewidth}
\includegraphics[width=\linewidth]{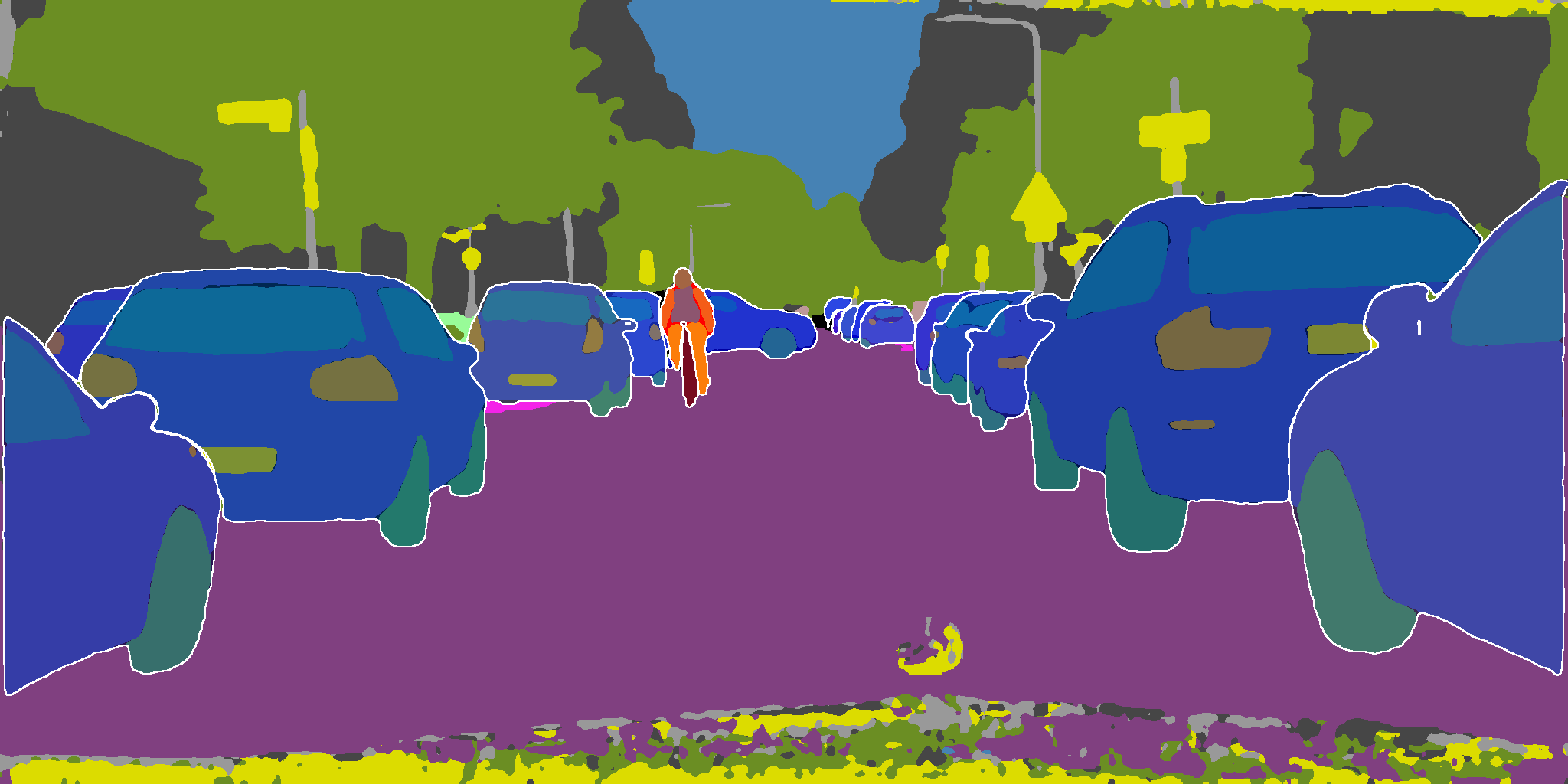}
    \end{subfigure}\hspace*{\fill}
     \begin{subfigure}{0.245\linewidth}
\includegraphics[width=\linewidth]{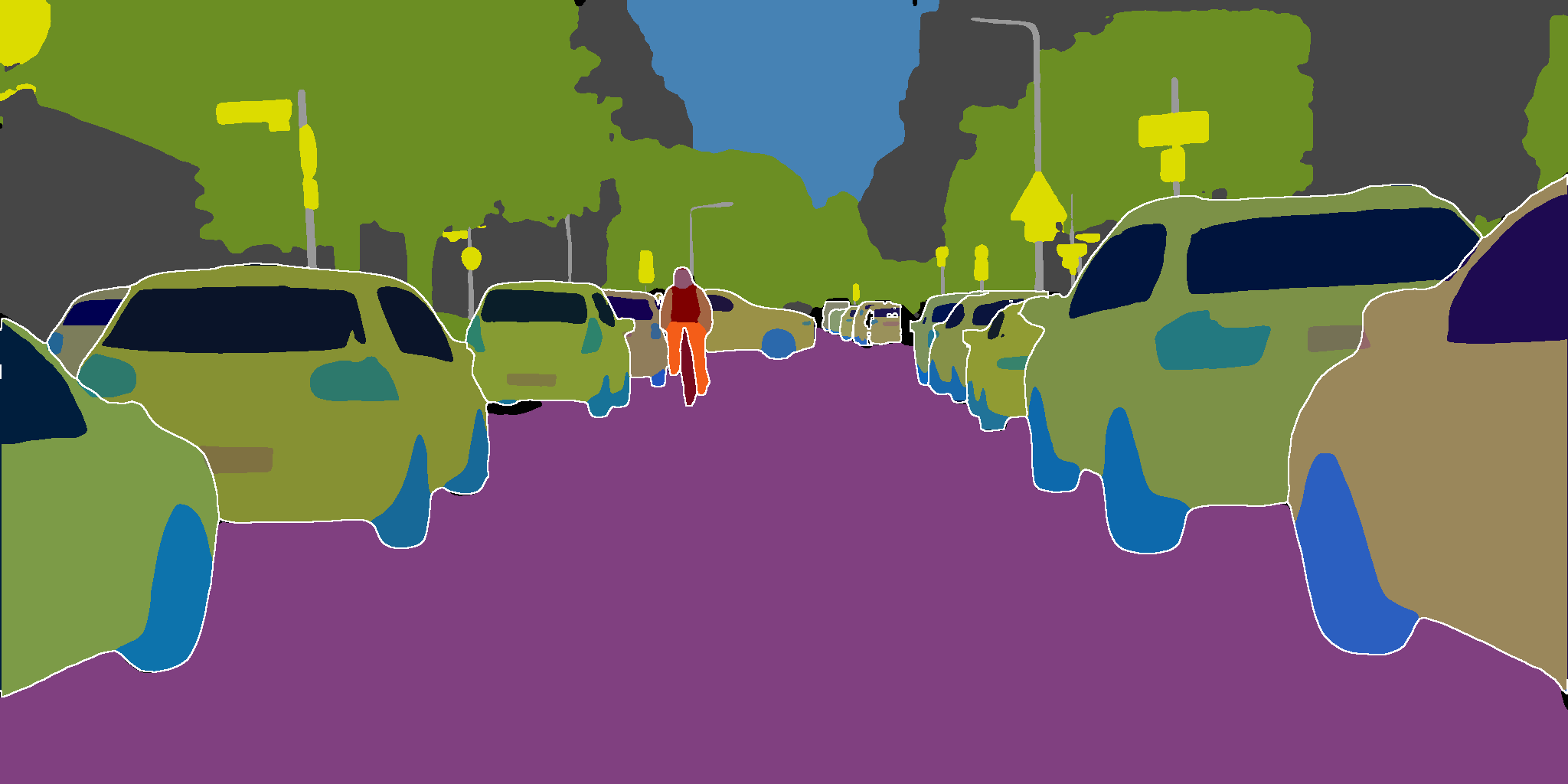}
    \end{subfigure}
    
    \vspace*{\fill}

        \begin{subfigure}{0.245\linewidth}
\includegraphics[width=\linewidth]{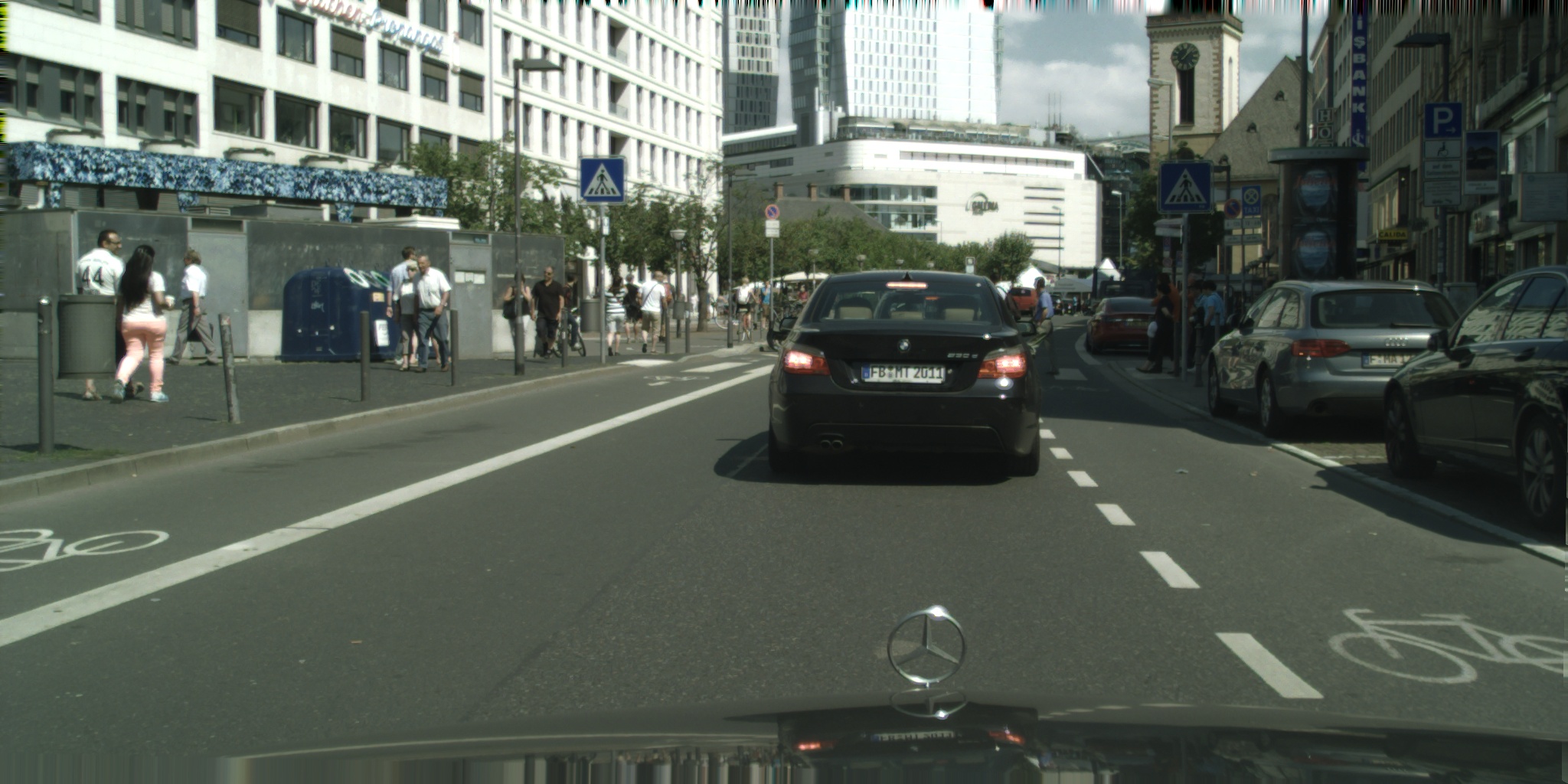}
    \end{subfigure}\hspace*{\fill}
     \begin{subfigure}{0.245\linewidth}
\includegraphics[width=\linewidth]{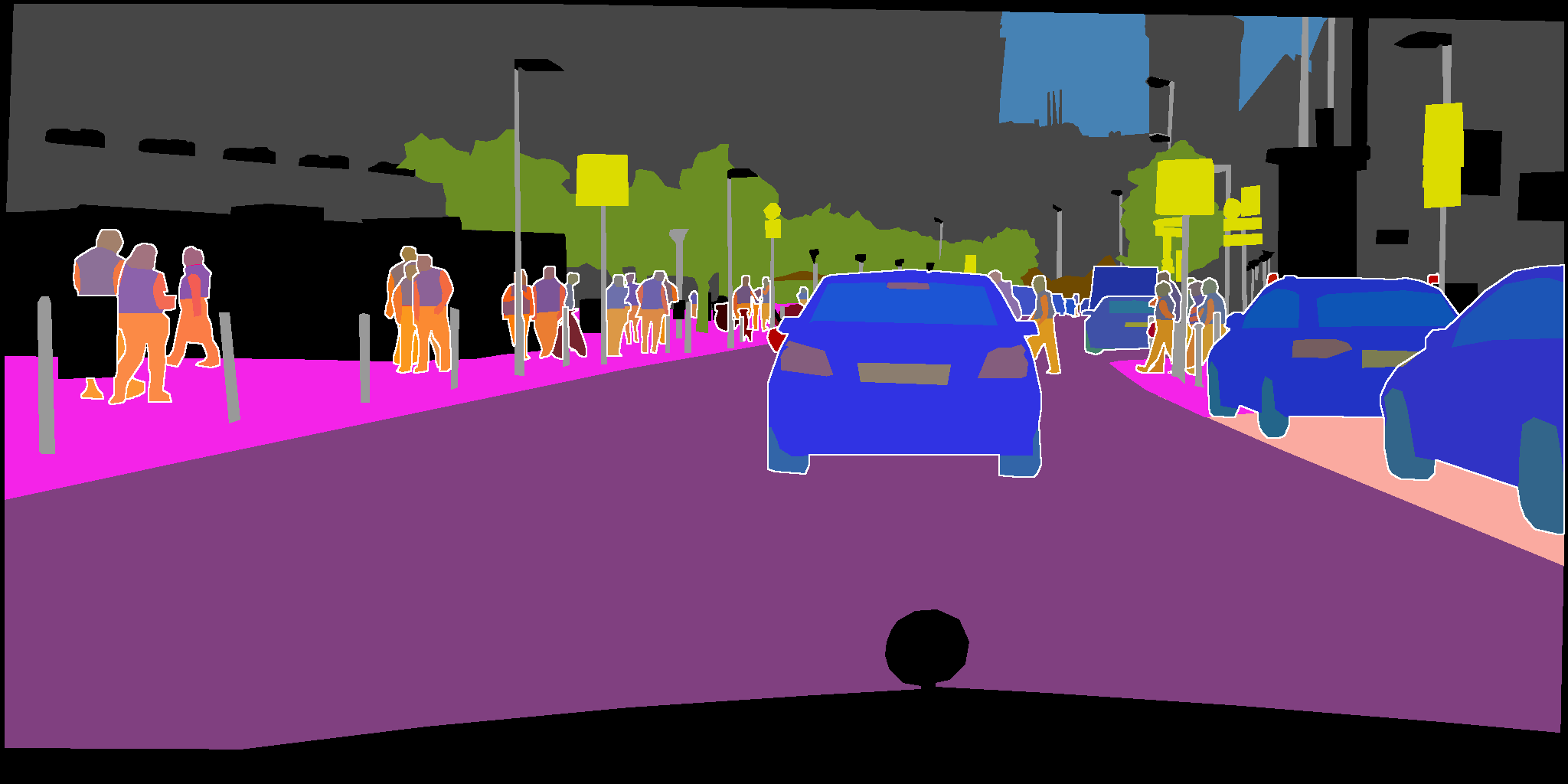}
    \end{subfigure}\hspace*{\fill}
     \begin{subfigure}{0.245\linewidth}
\includegraphics[width=\linewidth]{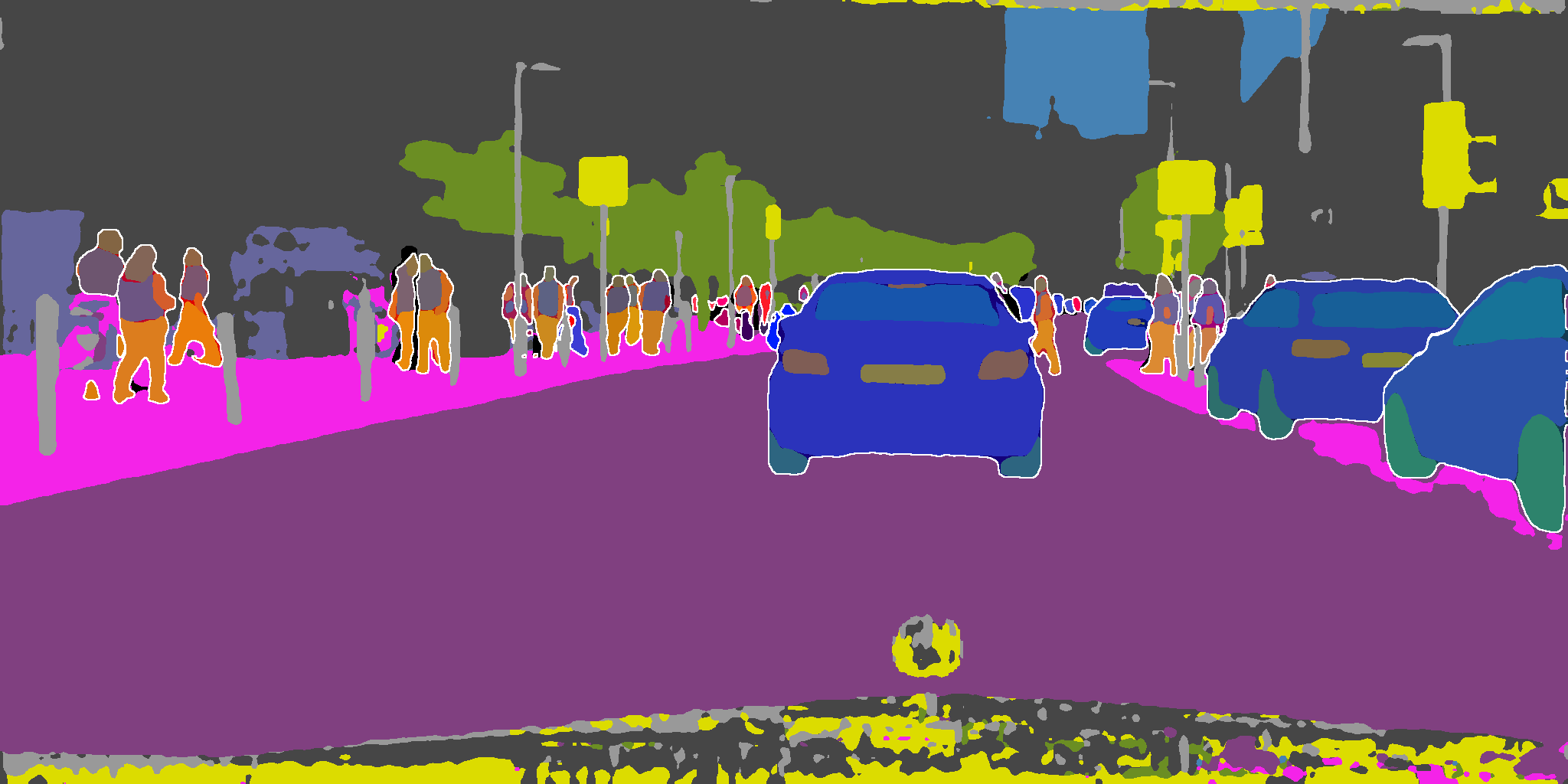}
    \end{subfigure}\hspace*{\fill}
     \begin{subfigure}{0.245\linewidth}
\includegraphics[width=\linewidth]{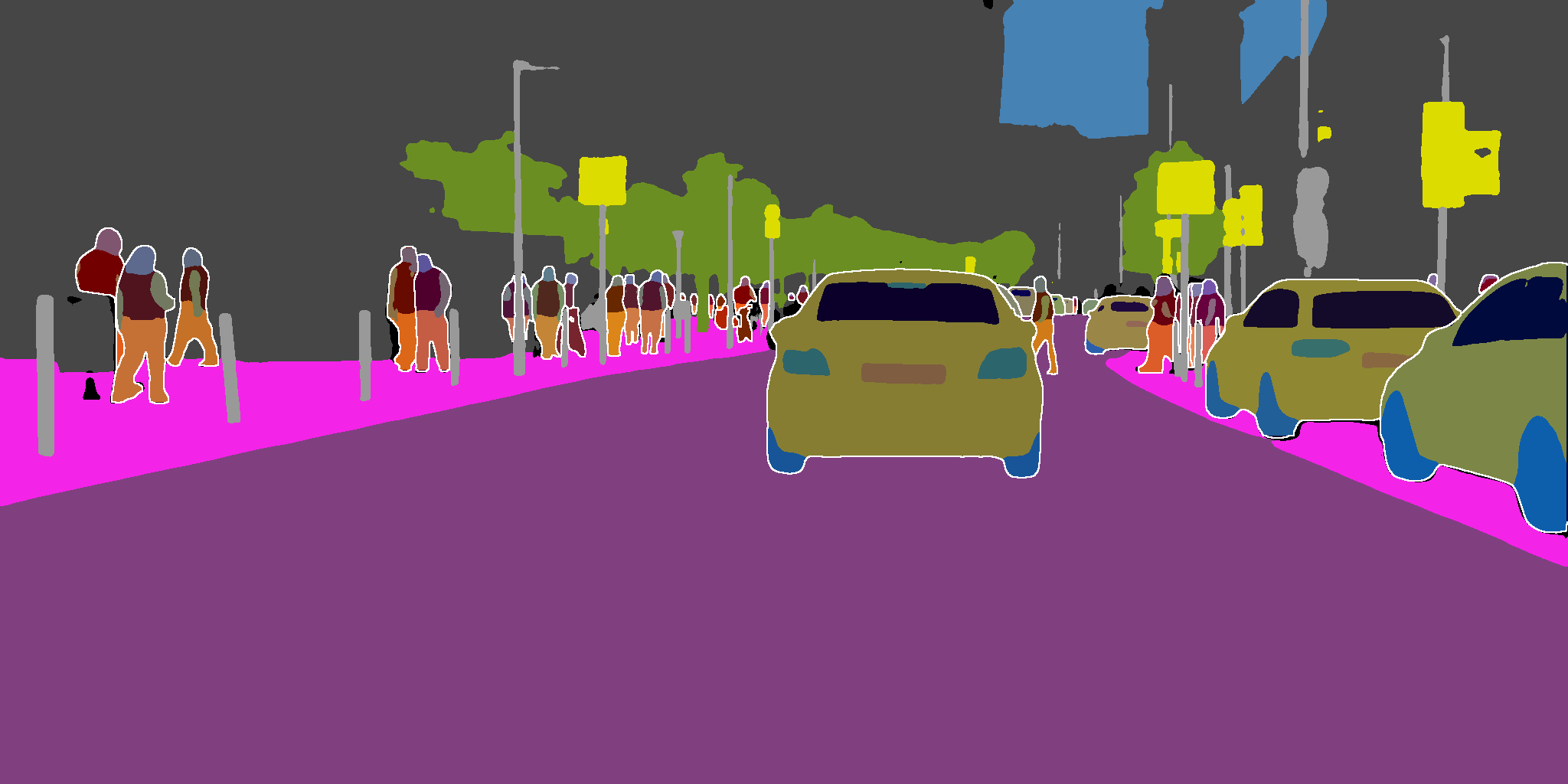}
    \end{subfigure}

    \begin{subfigure}{0.245\linewidth}
\includegraphics[width=\linewidth]{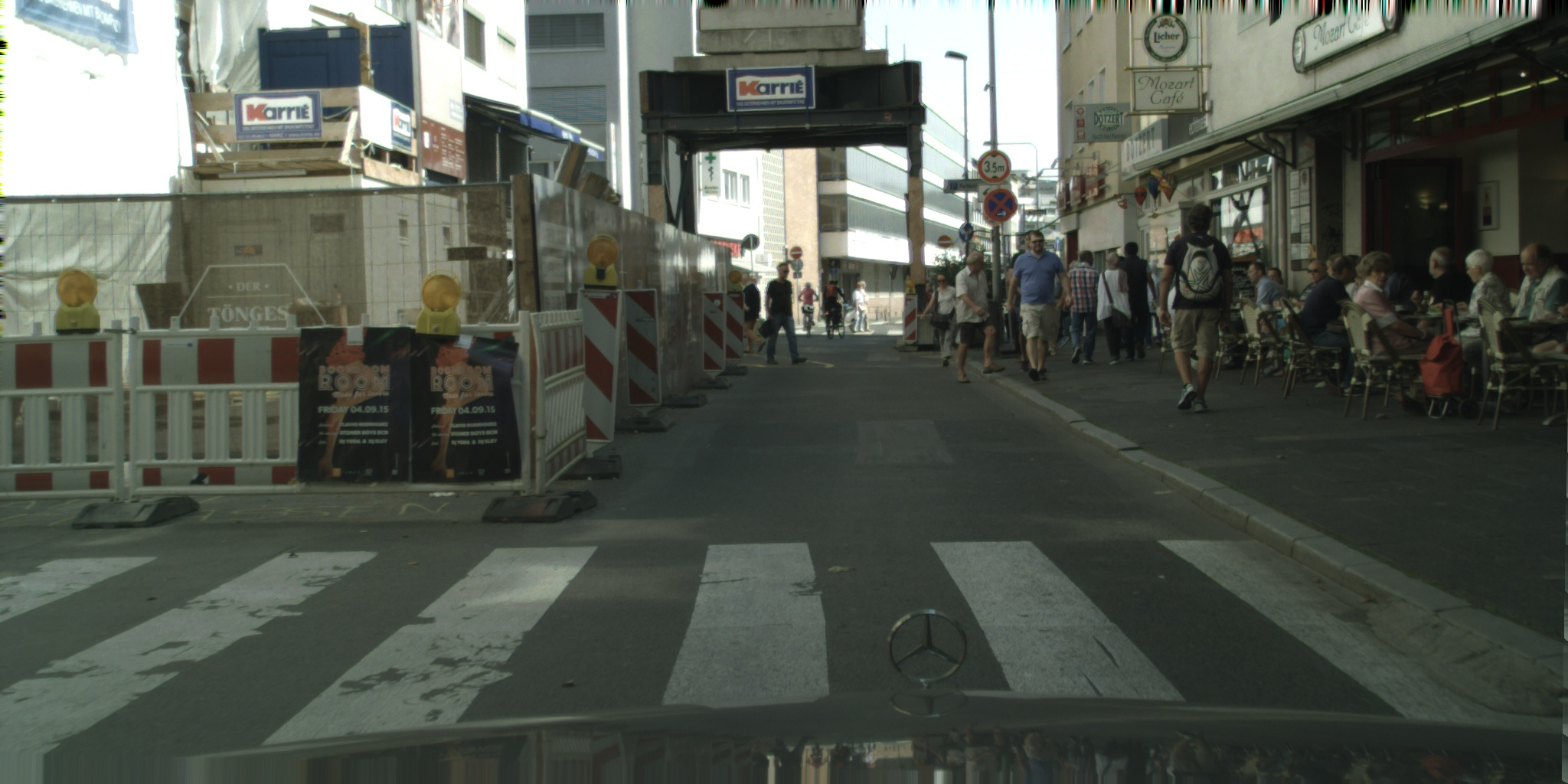}
\subcaption*{Original Image}
    \end{subfigure}\hspace*{\fill}
     \begin{subfigure}{0.245\linewidth}
\includegraphics[width=\linewidth]{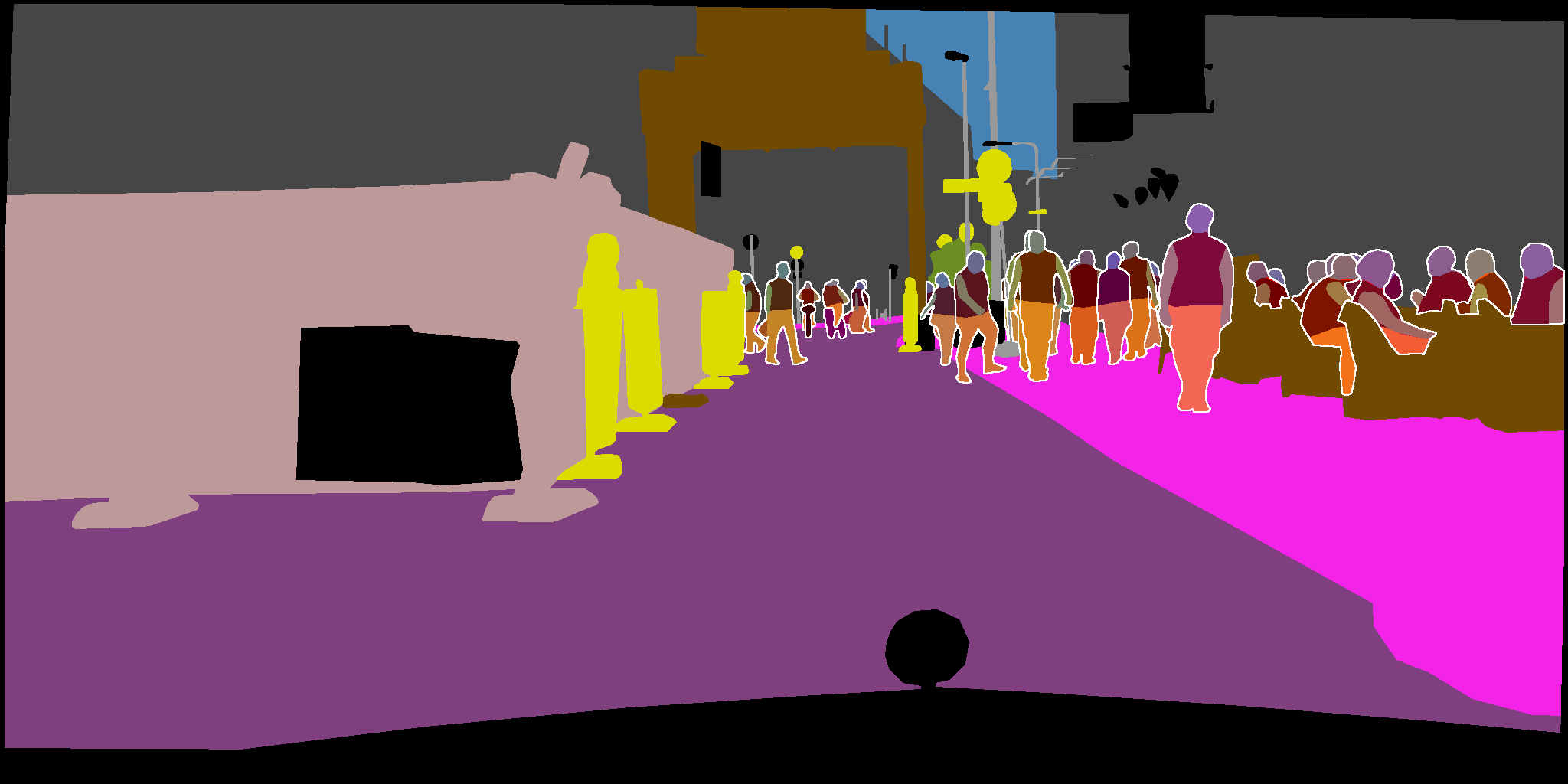}
\subcaption*{Ground-truth}
    \end{subfigure}\hspace*{\fill}
     \begin{subfigure}{0.245\linewidth}
\includegraphics[width=\linewidth]{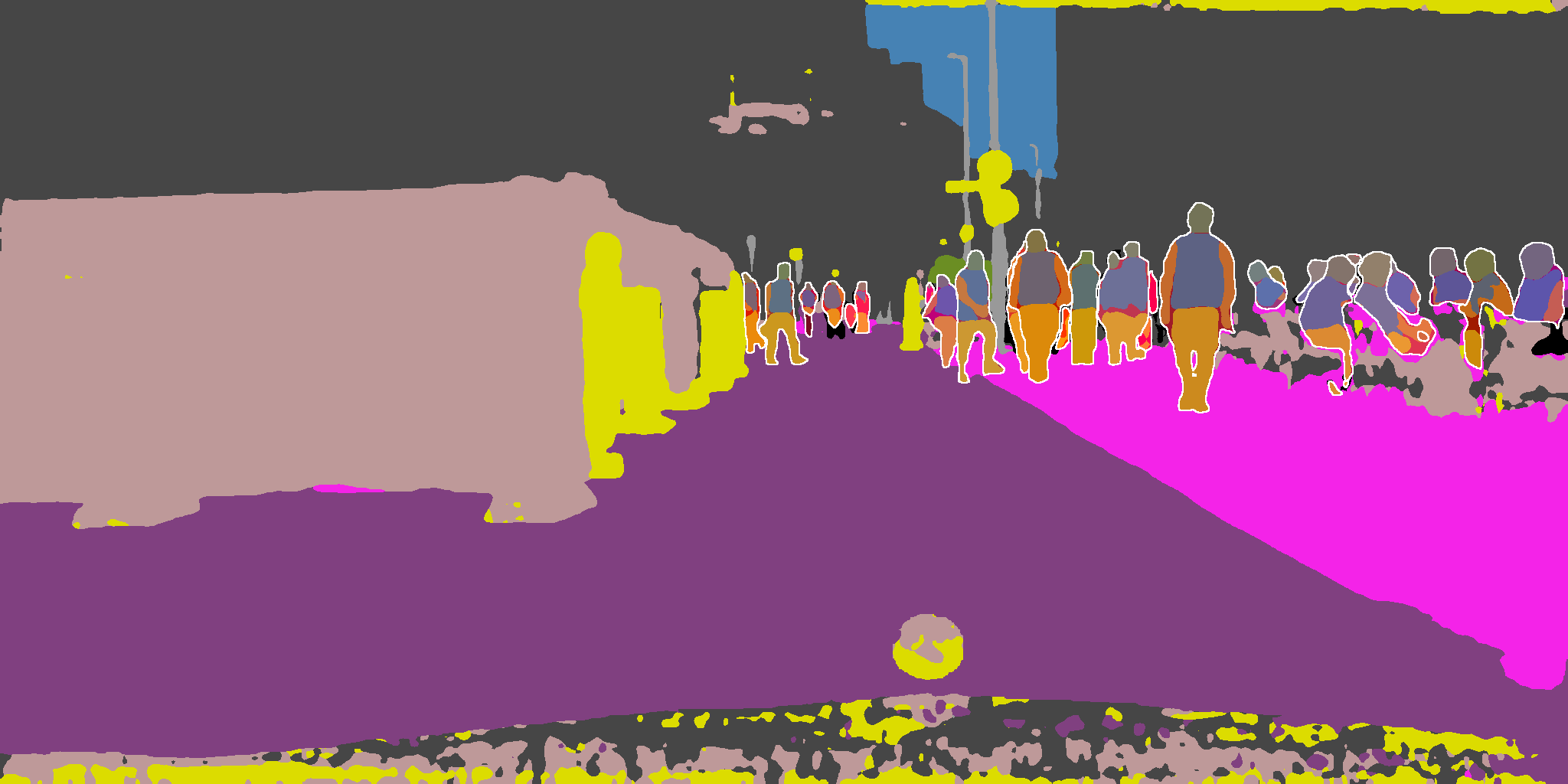}
\subcaption*{PPF \cite{li2022panoptic}}
    \end{subfigure}\hspace*{\fill}
     \begin{subfigure}{0.245\linewidth}
\includegraphics[width=\linewidth]{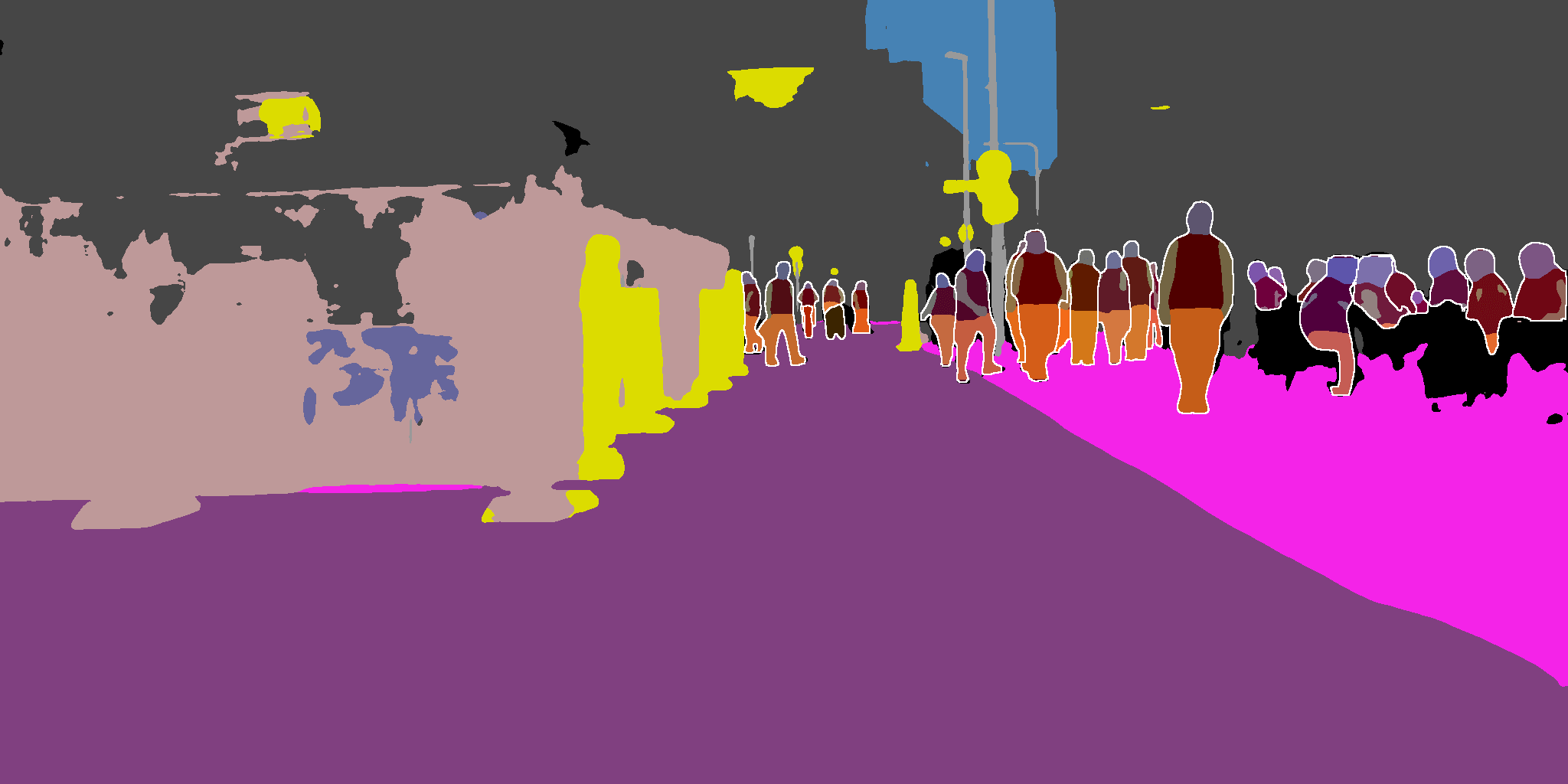}
\subcaption*{\ours}
    \end{subfigure}

    \caption{Additional qualitative results of our proposed model and PPF \cite{li2022panoptic} on CPP \citep{meletis2020cityscapes}}
    \label{fig:appendix:visualcityscapes}
\end{figure*}

\begin{figure*}
    \centering
    \begin{subfigure}{0.245\linewidth}
\includegraphics[width=\linewidth]{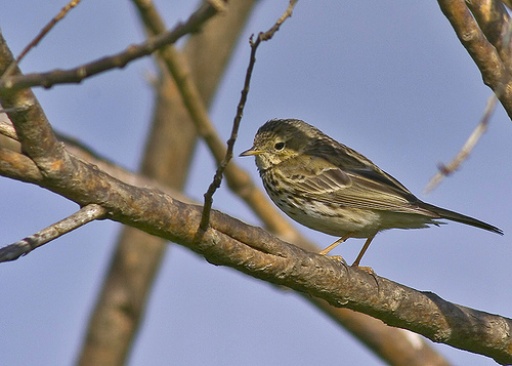}
    \end{subfigure}\hspace*{\fill}
     \begin{subfigure}{0.245\linewidth}
\includegraphics[width=\linewidth]{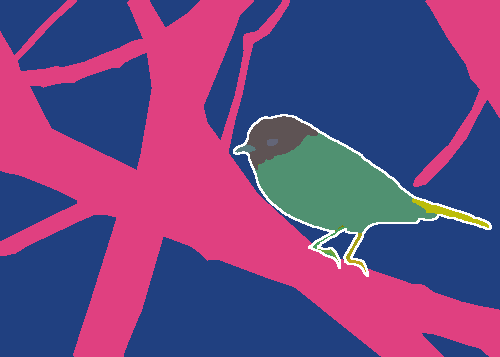}
    \end{subfigure}\hspace*{\fill}
     \begin{subfigure}{0.245\linewidth}
\includegraphics[width=\linewidth]{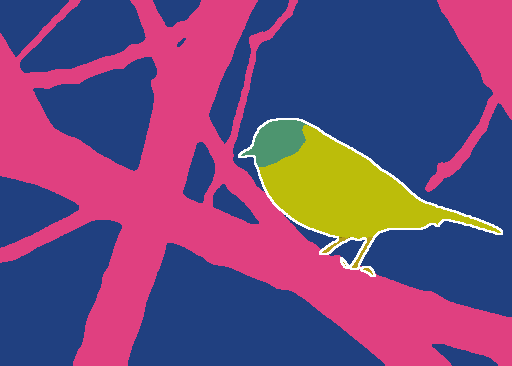}
    \end{subfigure}\hspace*{\fill}
     \begin{subfigure}{0.245\linewidth}
\includegraphics[width=\linewidth]{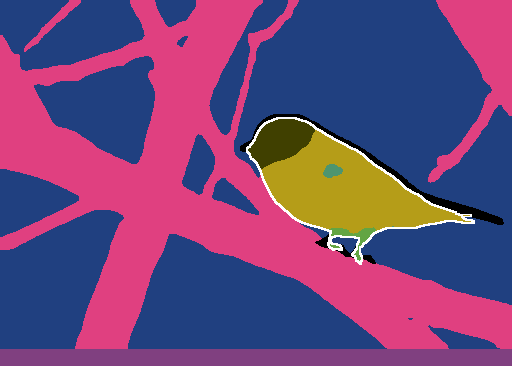}
    \end{subfigure}
    
    \vspace*{\fill}
    
        \begin{subfigure}{0.245\linewidth}
\includegraphics[width=\linewidth]{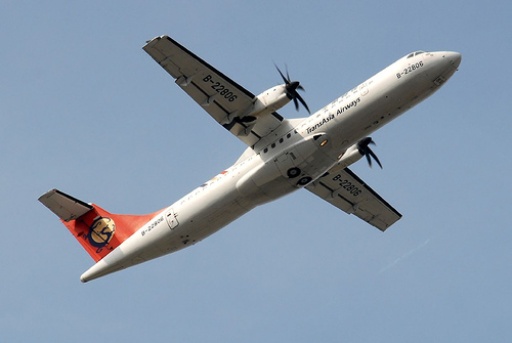}
    \end{subfigure}\hspace*{\fill}
     \begin{subfigure}{0.245\linewidth}
\includegraphics[width=\linewidth]{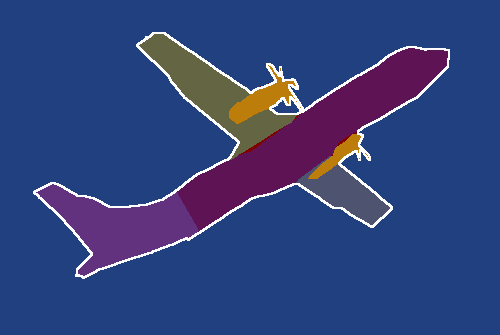}
    \end{subfigure}\hspace*{\fill}
     \begin{subfigure}{0.245\linewidth}
\includegraphics[width=\linewidth]{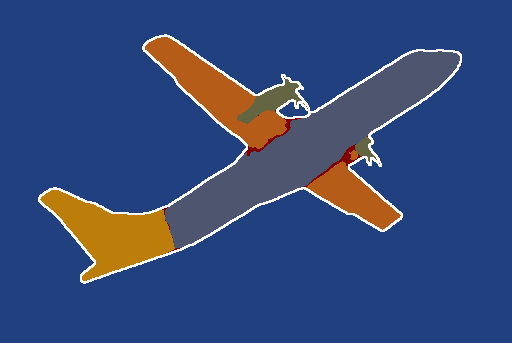}
    \end{subfigure}\hspace*{\fill}
     \begin{subfigure}{0.245\linewidth}
\includegraphics[width=\linewidth]{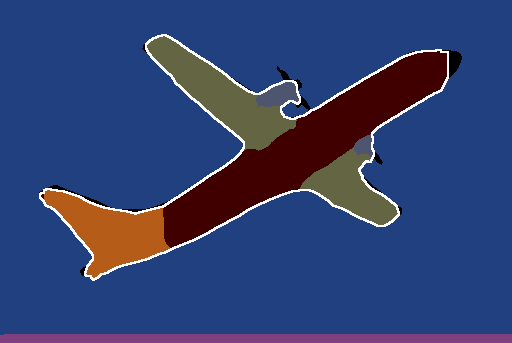}
    \end{subfigure}
    
    \vspace*{\fill}

        \begin{subfigure}{0.245\linewidth}
\includegraphics[width=\linewidth]{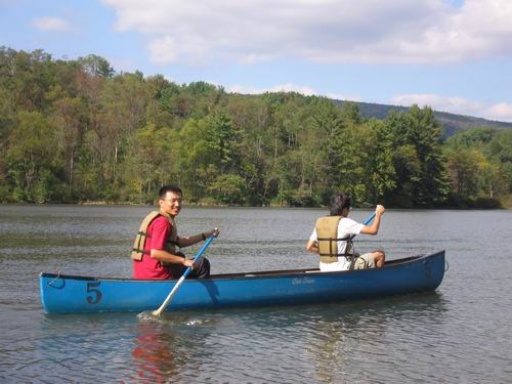}
    \end{subfigure}\hspace*{\fill}
     \begin{subfigure}{0.245\linewidth}
\includegraphics[width=\linewidth]{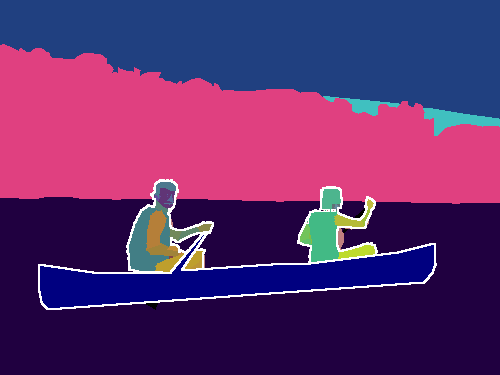}
    \end{subfigure}\hspace*{\fill}
     \begin{subfigure}{0.245\linewidth}
\includegraphics[width=\linewidth]{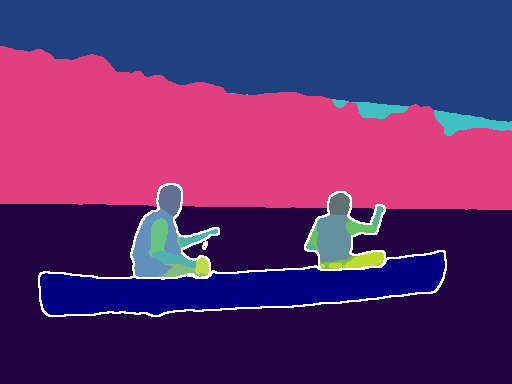}
    \end{subfigure}\hspace*{\fill}
     \begin{subfigure}{0.245\linewidth}
\includegraphics[width=\linewidth]{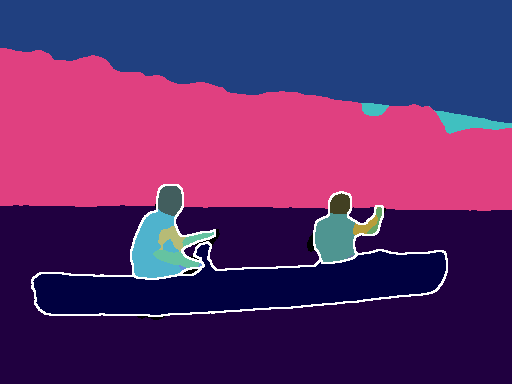}
    \end{subfigure}

    \begin{subfigure}{0.245\linewidth}
\includegraphics[width=\linewidth]{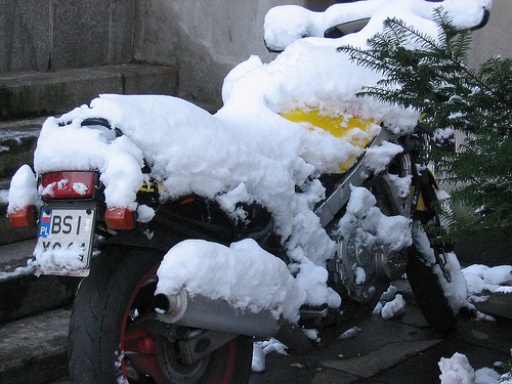}
\subcaption*{Original Image}
    \end{subfigure}\hspace*{\fill}
     \begin{subfigure}{0.245\linewidth}
\includegraphics[width=\linewidth]{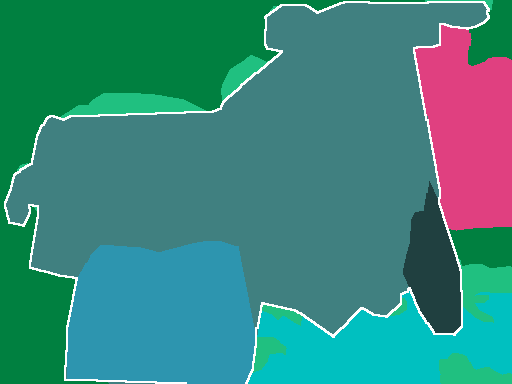}
\subcaption*{Ground-truth}
    \end{subfigure}\hspace*{\fill}
     \begin{subfigure}{0.245\linewidth}
\includegraphics[width=\linewidth]{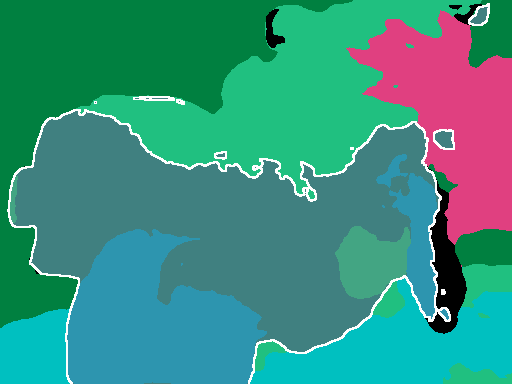}
\subcaption*{PPF \cite{li2022panoptic}}
    \end{subfigure}\hspace*{\fill}
     \begin{subfigure}{0.245\linewidth}
\includegraphics[width=\linewidth]{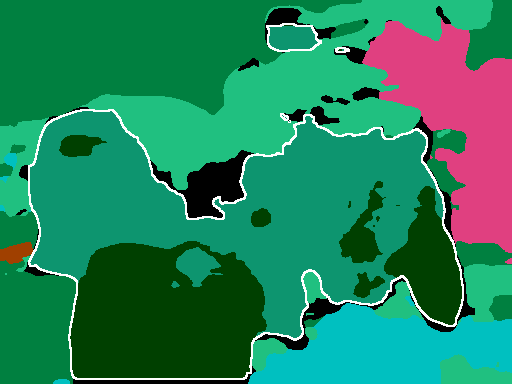}
\subcaption*{\ours}
    \end{subfigure}

    \caption{Additional qualitative results of our proposed model and PPF \cite{li2022panoptic} on PPP \citep{meletis2020cityscapes}}
    \label{fig:appendix:visualpascal}
\end{figure*}

\begin{figure*}
    \centering

    \begin{subfigure}{0.495\linewidth}
        \includegraphics[width=\linewidth]{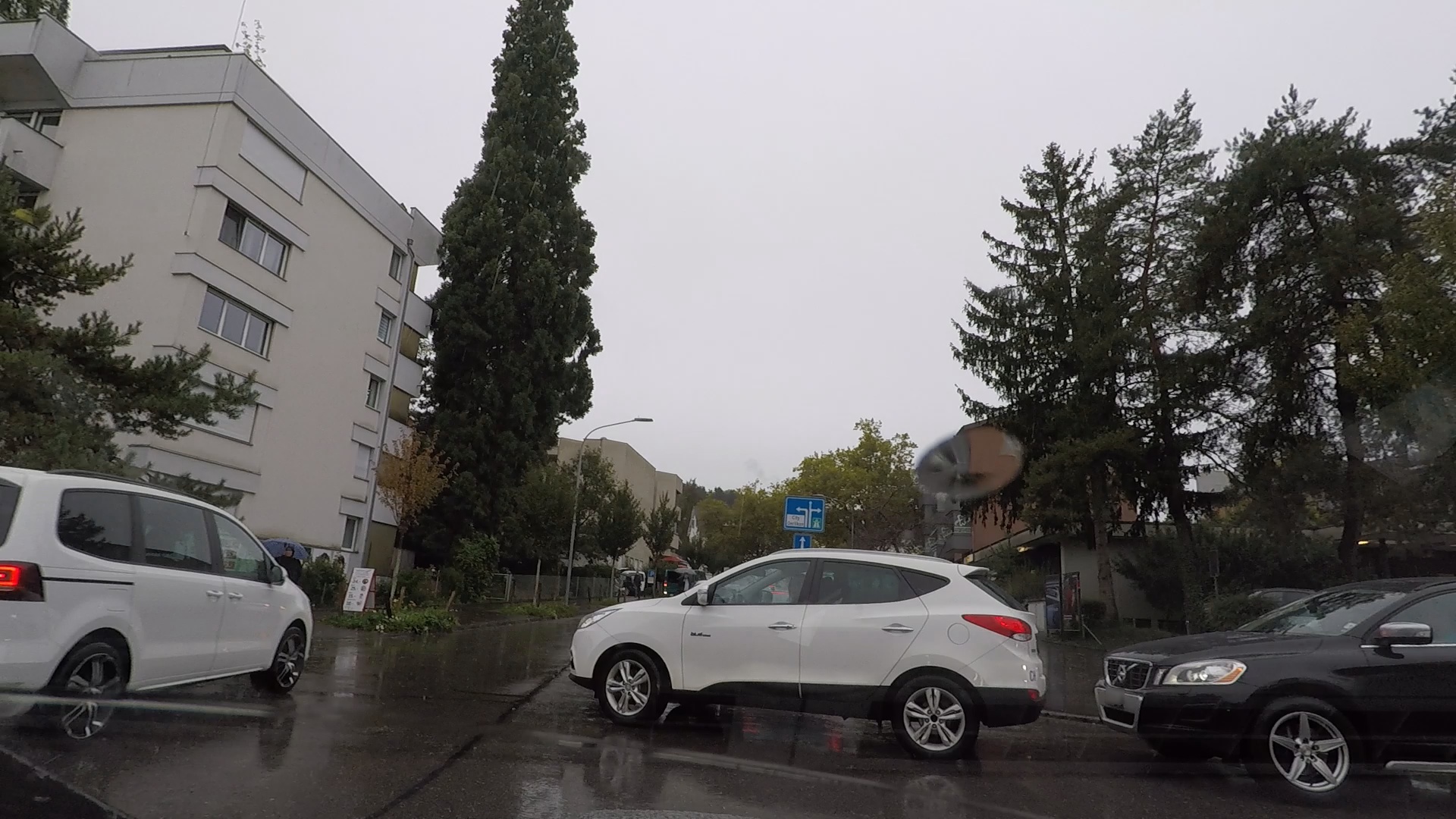}
    \end{subfigure}\hspace*{\fill}
    \begin{subfigure}{0.495\linewidth}
        \includegraphics[width=\linewidth]{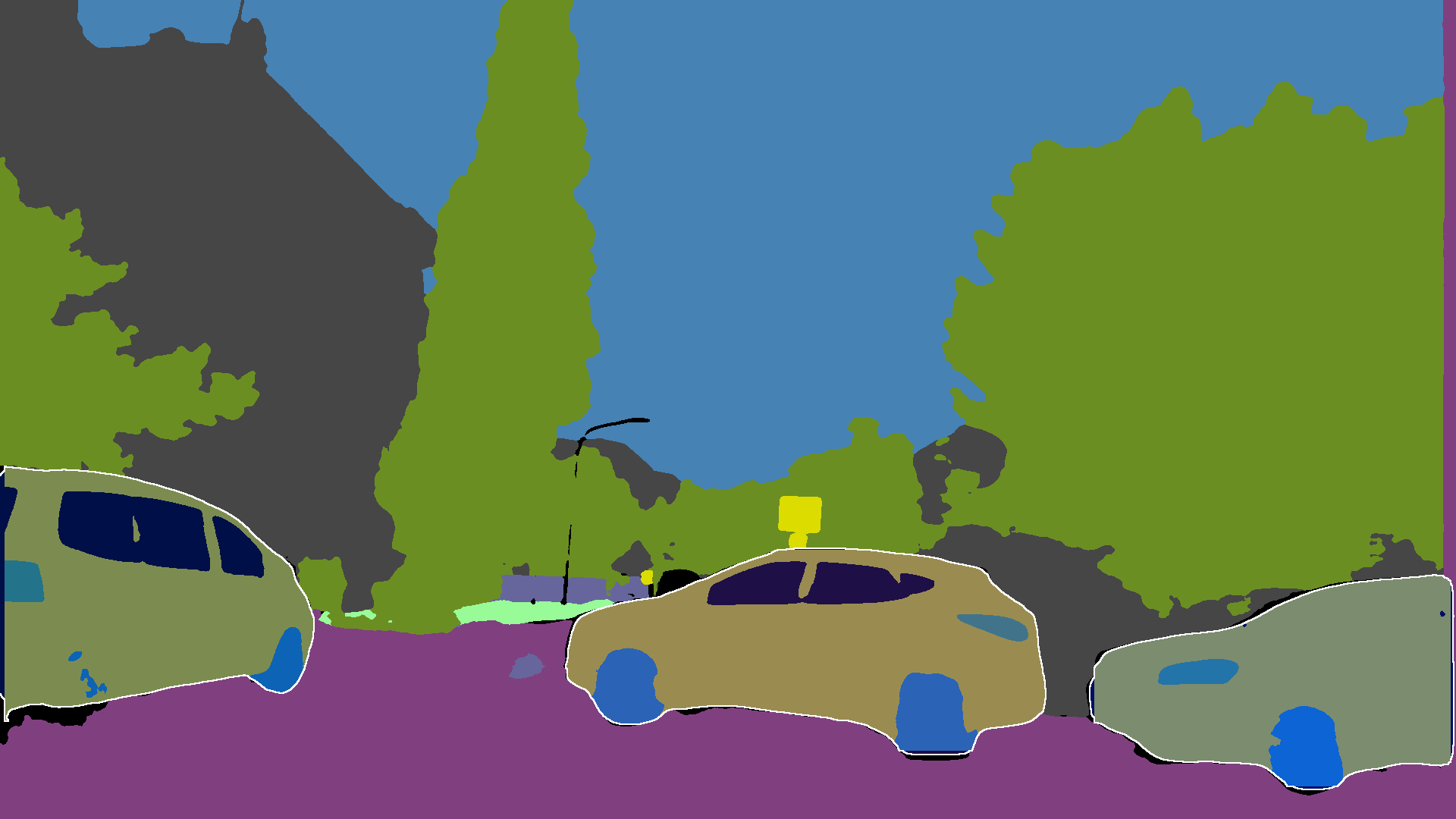}
    \end{subfigure}
    
    \begin{subfigure}{0.495\linewidth}
        \includegraphics[width=\linewidth]{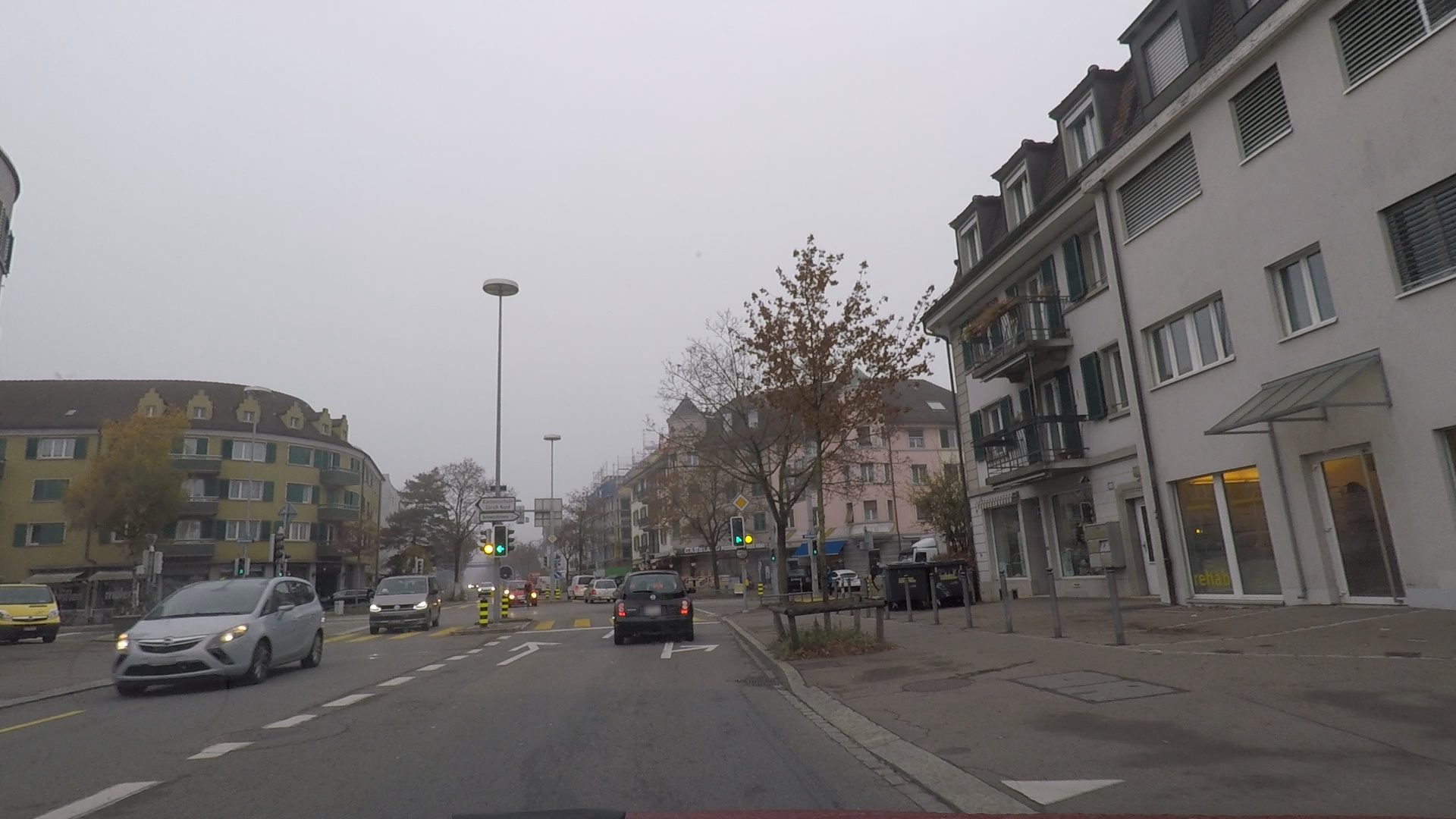}
        \subcaption*{Input Images}
    \end{subfigure}\hspace*{\fill}
    \begin{subfigure}{0.495\linewidth}
        \includegraphics[width=\linewidth]{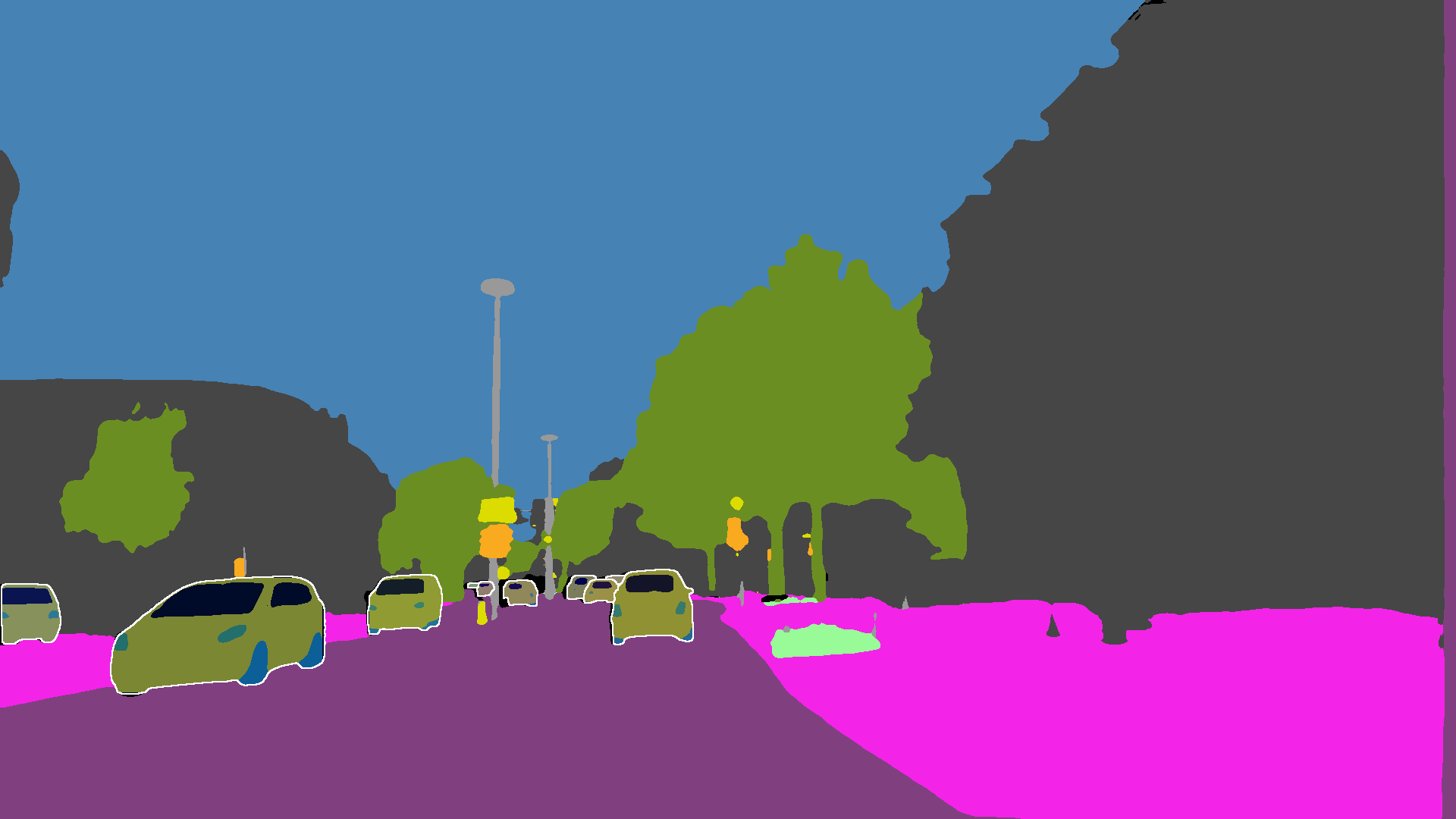}
        \subcaption*{Results of JPPF}
    \end{subfigure}
    
    \caption{Visual results of our JPPF on the ACDC dataset \citep{sakaridis2021acdc} without fine-tuning}
    \label{fig:appendix:acdc}
\end{figure*}

\begin{figure*}
    \centering

    \begin{subfigure}{0.495\linewidth}
        \includegraphics[width=\linewidth]{bdd_01}
    \end{subfigure}\hspace*{\fill}
    \begin{subfigure}{0.495\linewidth}
        \includegraphics[width=\linewidth]{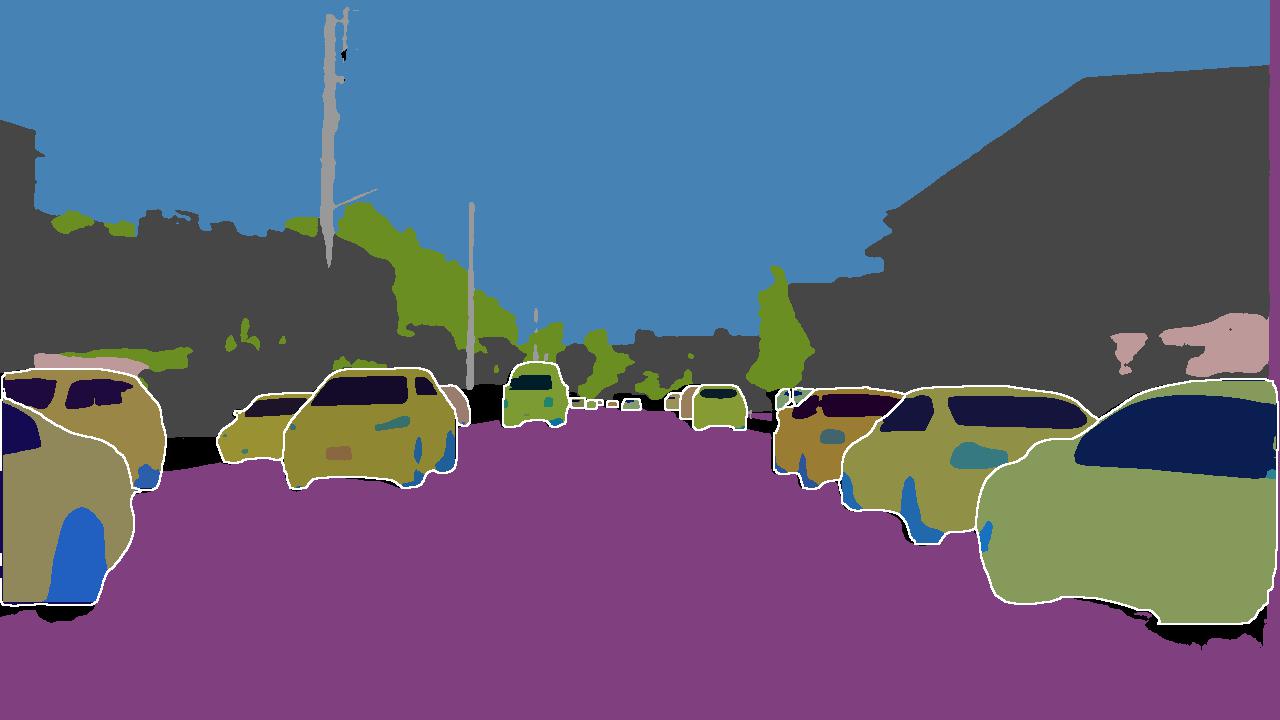}
    \end{subfigure}
    
    \begin{subfigure}{0.495\linewidth}
        \includegraphics[width=\linewidth]{bdd_fail}
        \subcaption*{Input Images}
    \end{subfigure}\hspace*{\fill}
    \begin{subfigure}{0.495\linewidth}
        \includegraphics[width=\linewidth]{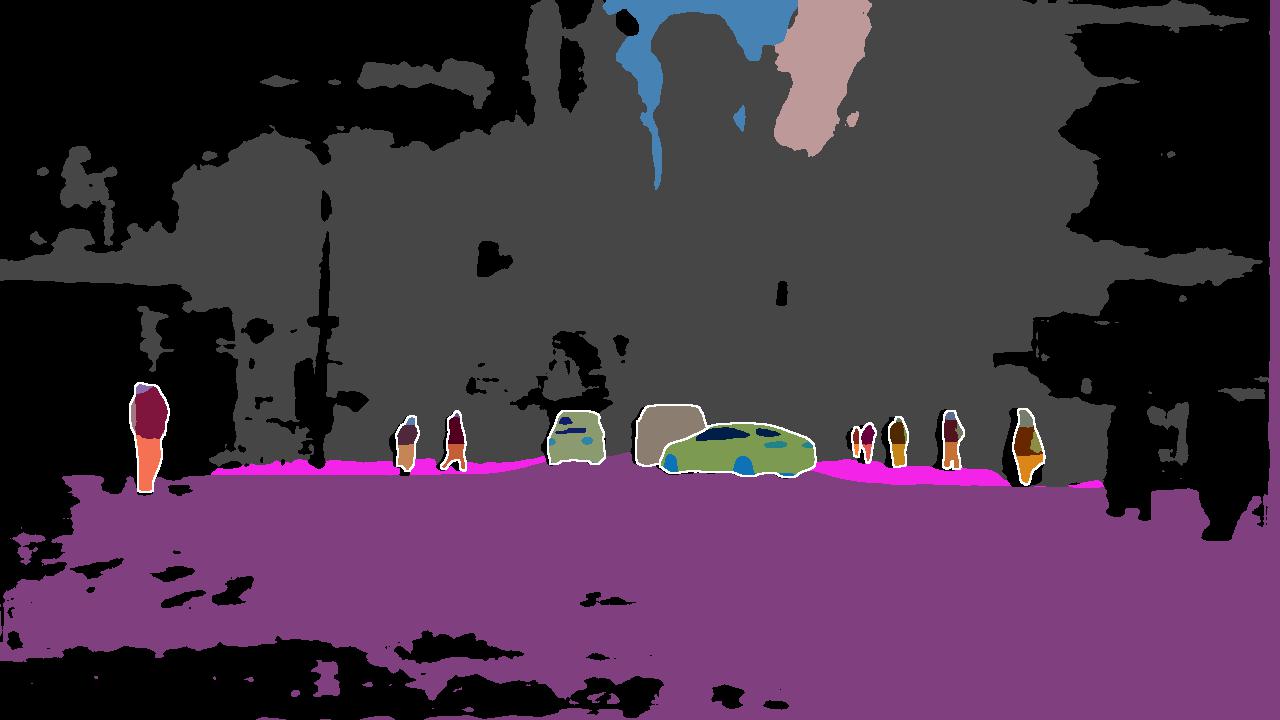}
        \subcaption*{Results of JPPF}
    \end{subfigure}
    
    \caption{Visual results of our JPPF on the BDD100K dataset \citep{yu2020bdd100k} without fine-tuning. The second example shows a failure case, in which our model is not able to properly generalize to the unseen data}
    \label{fig:appendix:bdd}
\end{figure*}

\begin{figure*}
    \centering

    \begin{subfigure}{0.495\linewidth}
        \includegraphics[width=\linewidth]{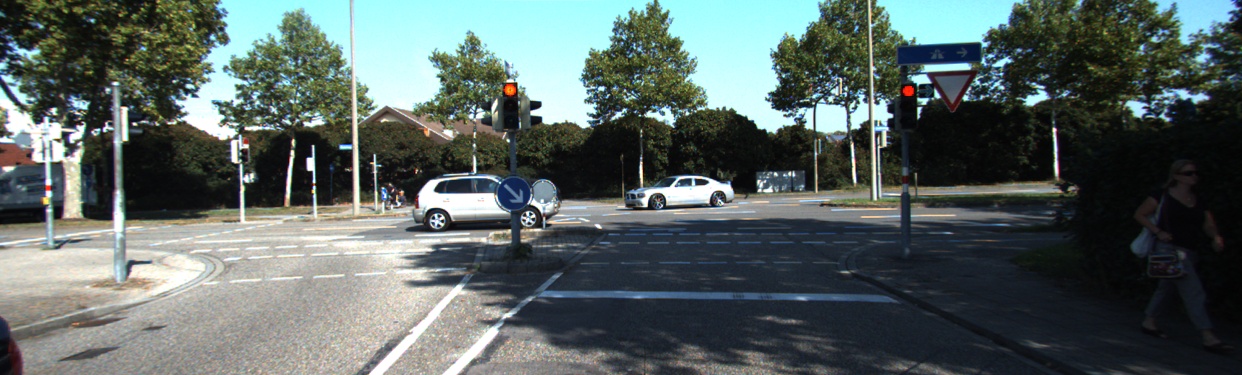}
    \end{subfigure}\hspace*{\fill}
    \begin{subfigure}{0.495\linewidth}
        \includegraphics[width=\linewidth]{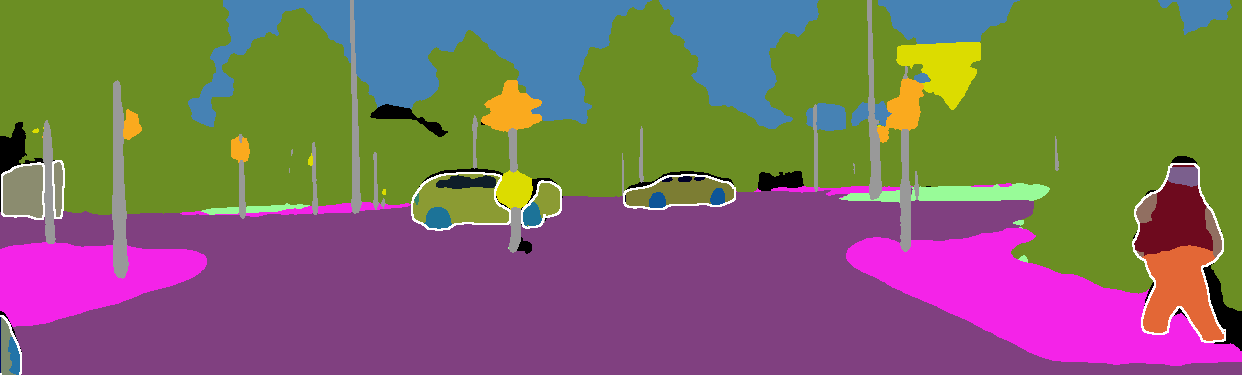}
    \end{subfigure}
    
    \begin{subfigure}{0.495\linewidth}
        \includegraphics[width=\linewidth]{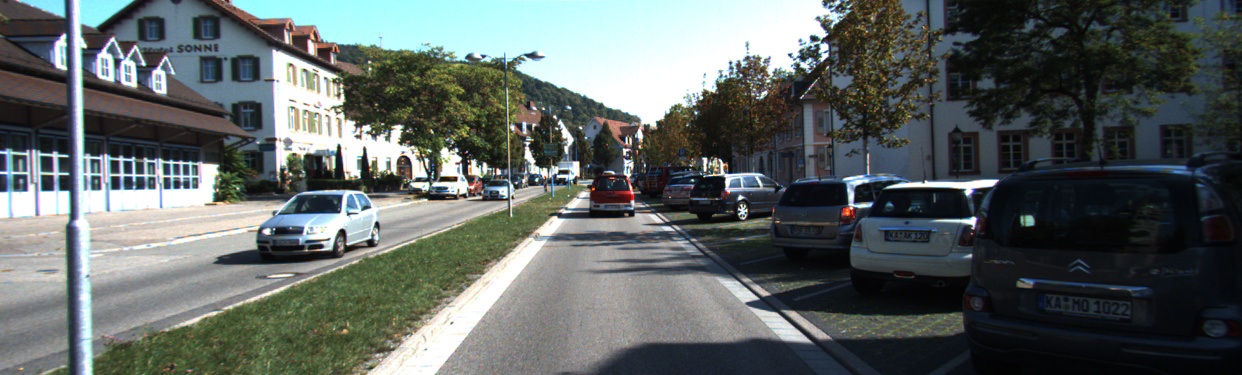}
        \subcaption*{Input Images}
    \end{subfigure}\hspace*{\fill}
    \begin{subfigure}{0.495\linewidth}
        \includegraphics[width=\linewidth]{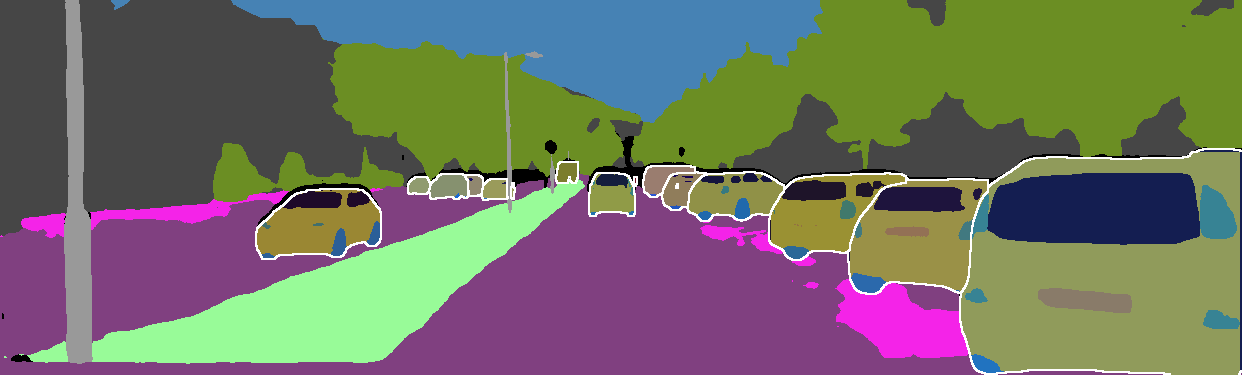}
        \subcaption*{Results of JPPF}
    \end{subfigure}
    
    \caption{Visual results of our JPPF on the KITTI dataset \citep{geiger2013vision} without fine-tuning}
    \label{fig:appendix:kitti}
\end{figure*}

\begin{figure*}
    \centering

    \begin{subfigure}{0.495\linewidth}
        \includegraphics[width=\linewidth]{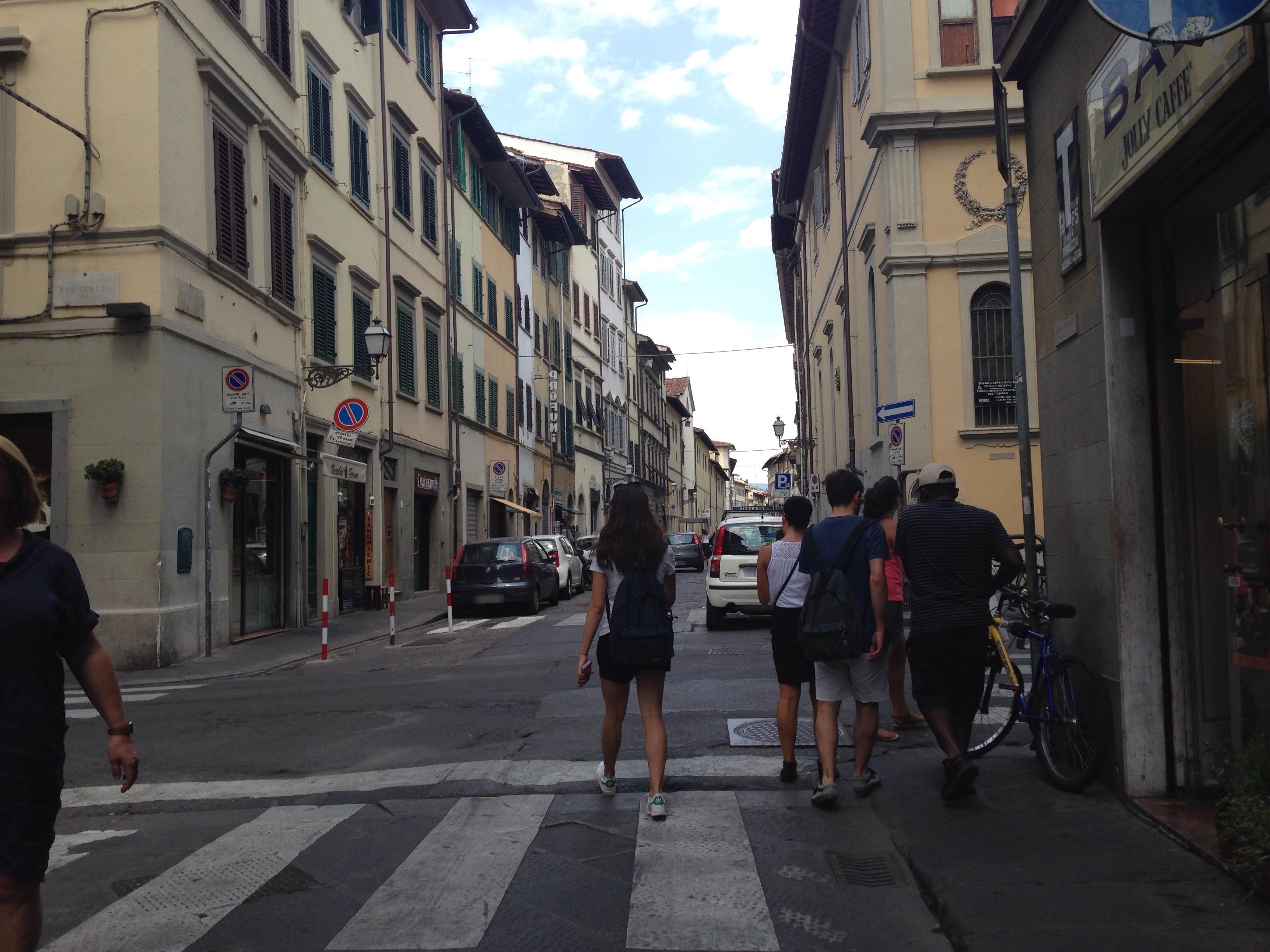}
    \end{subfigure}\hspace*{\fill}
    \begin{subfigure}{0.495\linewidth}
        \includegraphics[width=\linewidth]{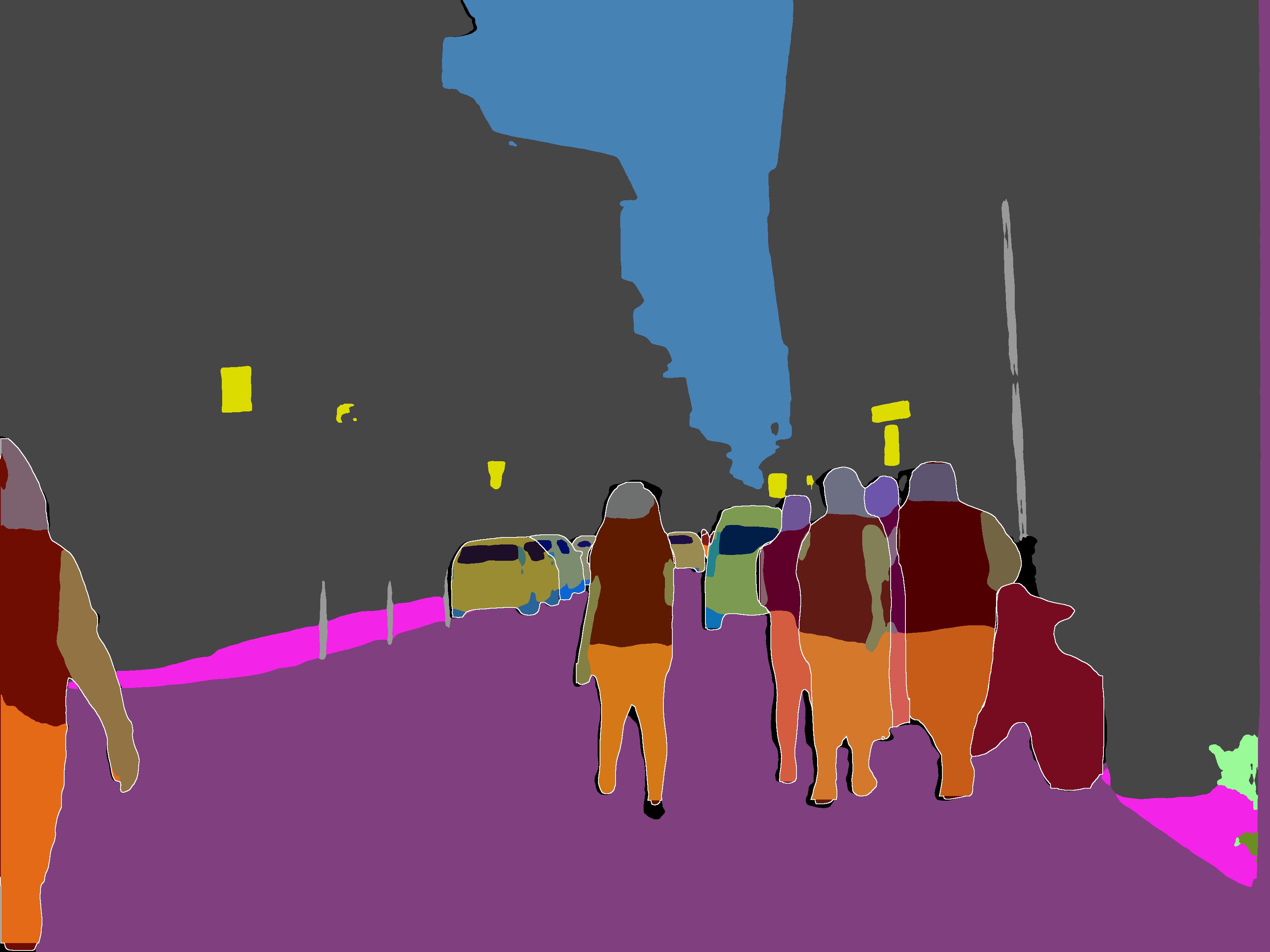}
    \end{subfigure}
    
    \begin{subfigure}{0.495\linewidth}
        \includegraphics[width=\linewidth]{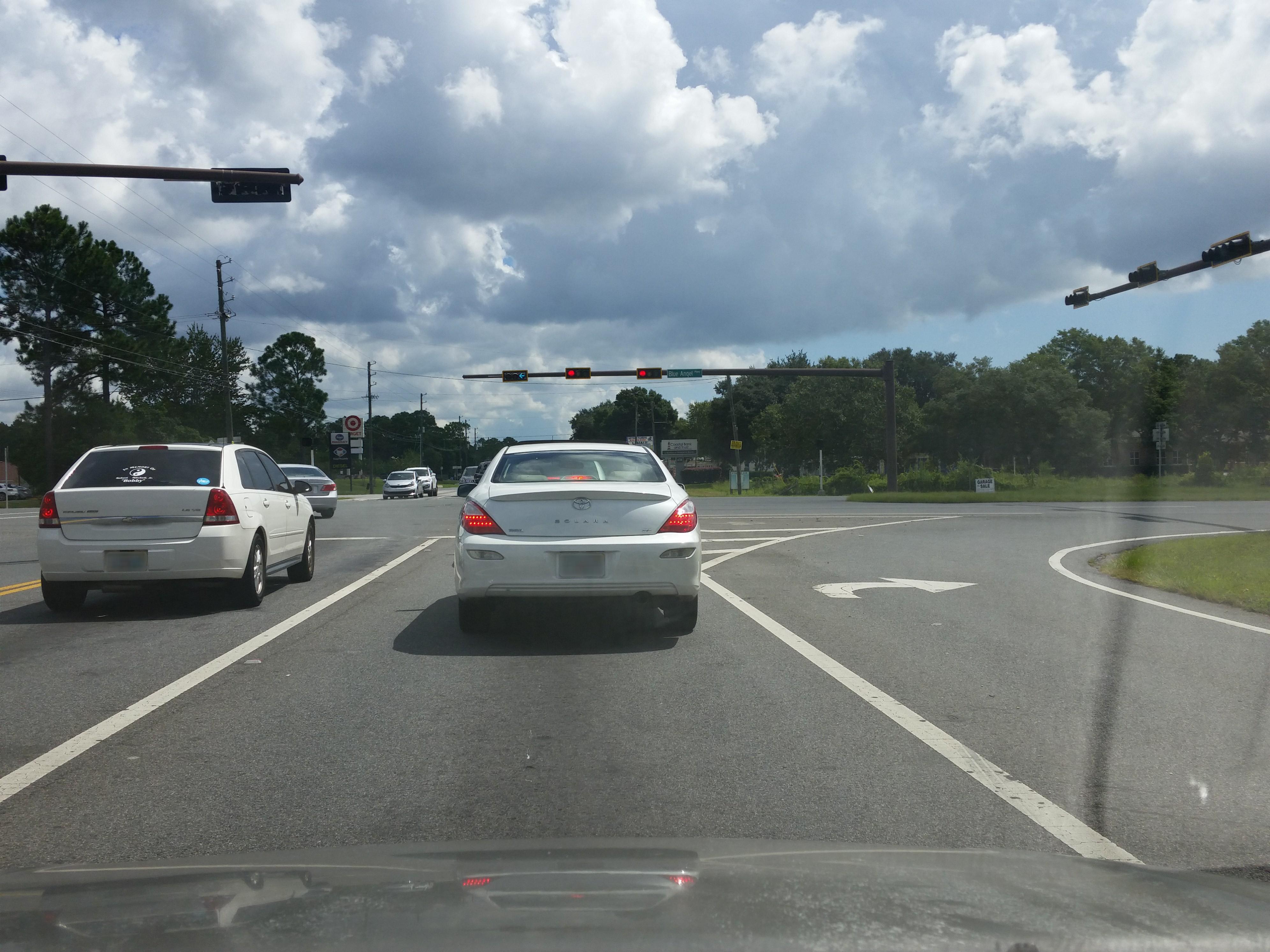}
        \subcaption*{Input Images}
    \end{subfigure}\hspace*{\fill}
    \begin{subfigure}{0.495\linewidth}
        \includegraphics[width=\linewidth]{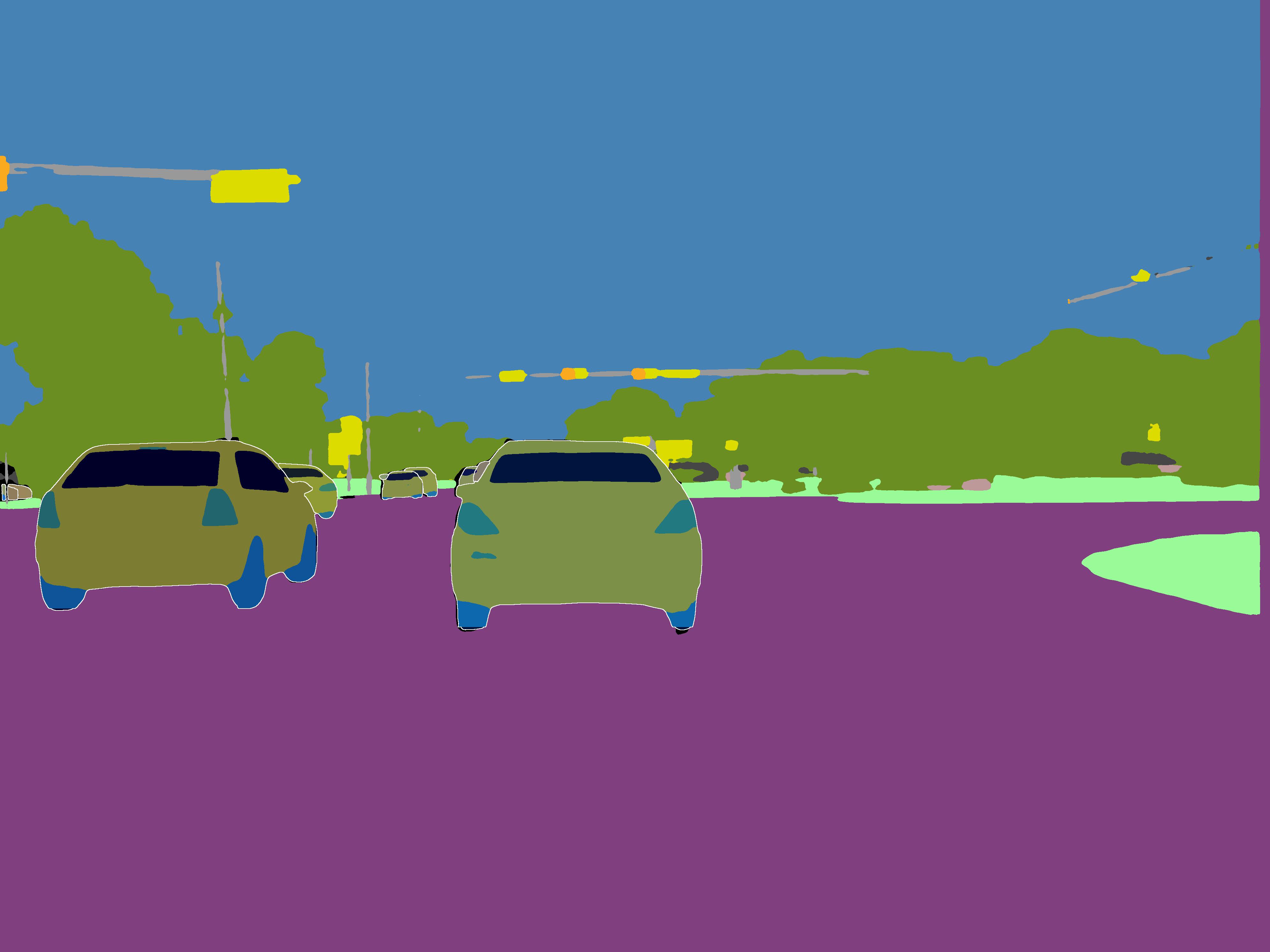}
        \subcaption*{Results of JPPF}
    \end{subfigure}
    
    \caption{Visual results of our JPPF on the Mapillary Vistas dataset \citep{neuhold2017mapillary} without fine-tuning}
    \label{fig:appendix:vistas}
\end{figure*}

\begin{figure*}
    \centering

    \begin{subfigure}{0.32\linewidth}
        \includegraphics[width=\linewidth]{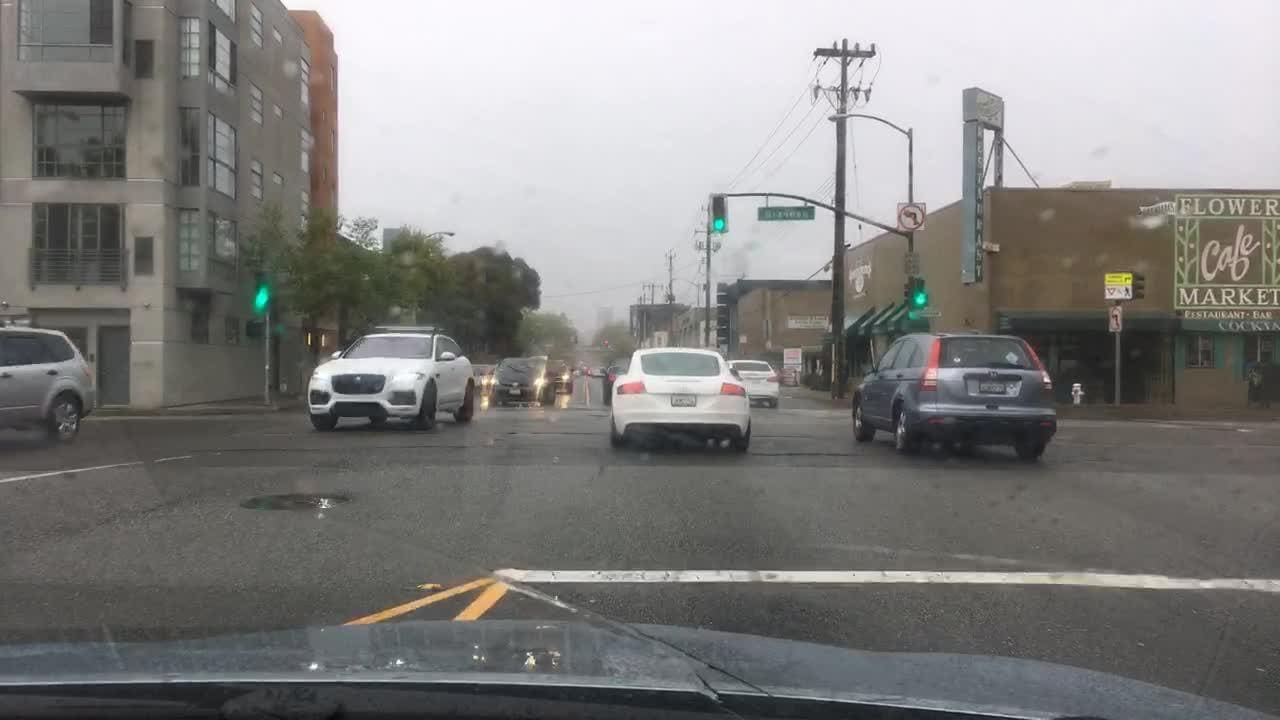}
    \end{subfigure}\hspace*{\fill}
    \begin{subfigure}{0.32\linewidth}
        \includegraphics[width=\linewidth]{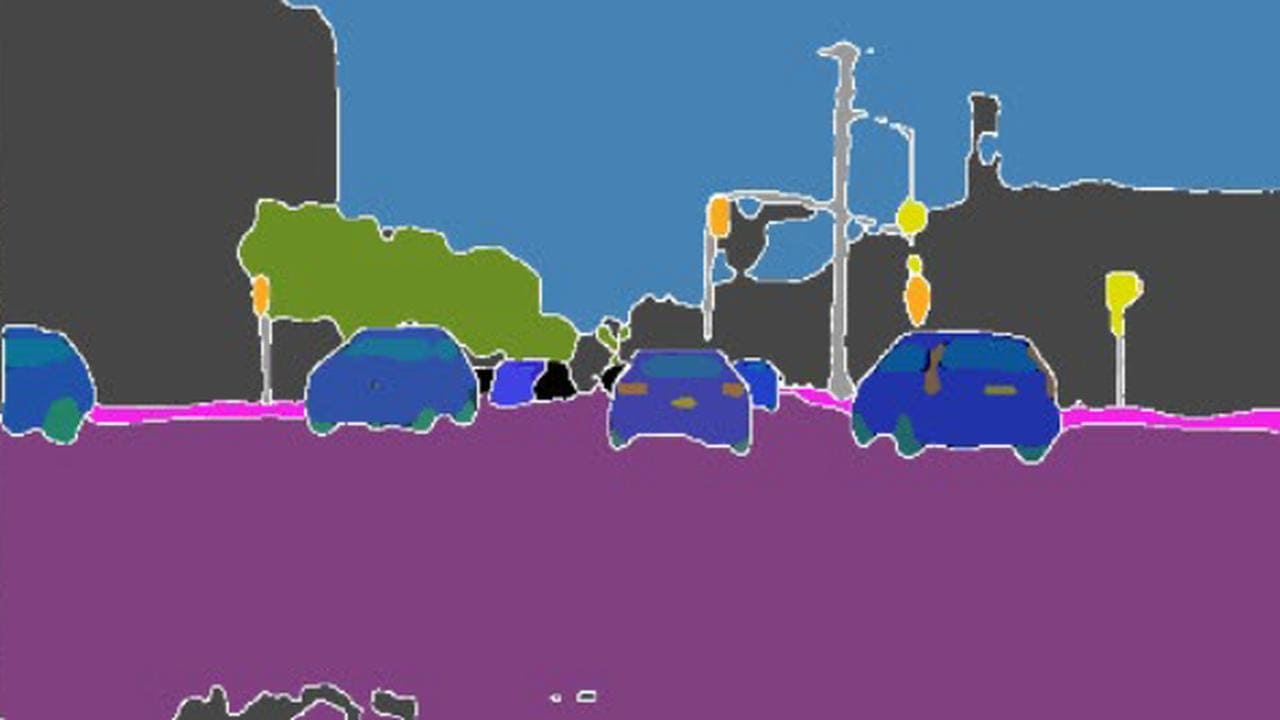}
    \end{subfigure}\hspace*{\fill}
    \begin{subfigure}{0.32\linewidth}
        \includegraphics[width=\linewidth]{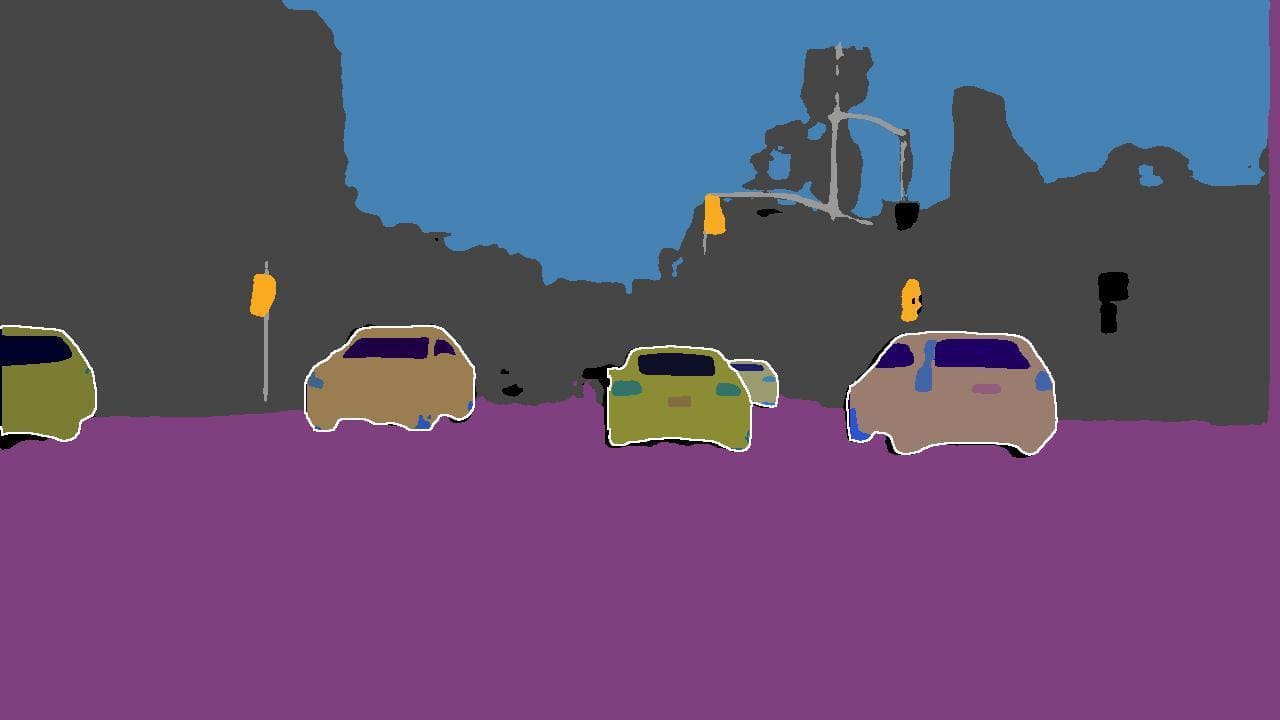}
    \end{subfigure}

    \begin{subfigure}{0.32\linewidth}
        \includegraphics[width=\linewidth]{b9c71ec0-1a77e45e_rgb}
    \end{subfigure}\hspace*{\fill}
    \begin{subfigure}{0.32\linewidth}
        \includegraphics[width=\linewidth]{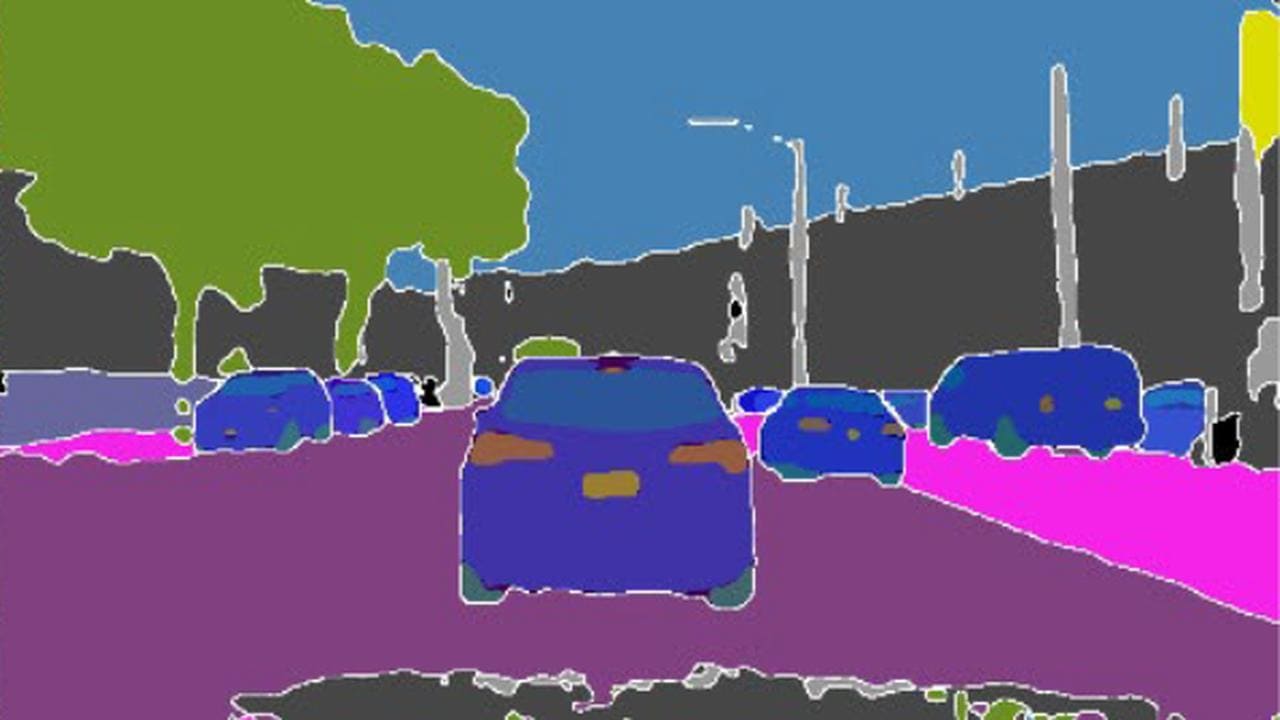}
    \end{subfigure}\hspace*{\fill}
    \begin{subfigure}{0.32\linewidth}
        \includegraphics[width=\linewidth]{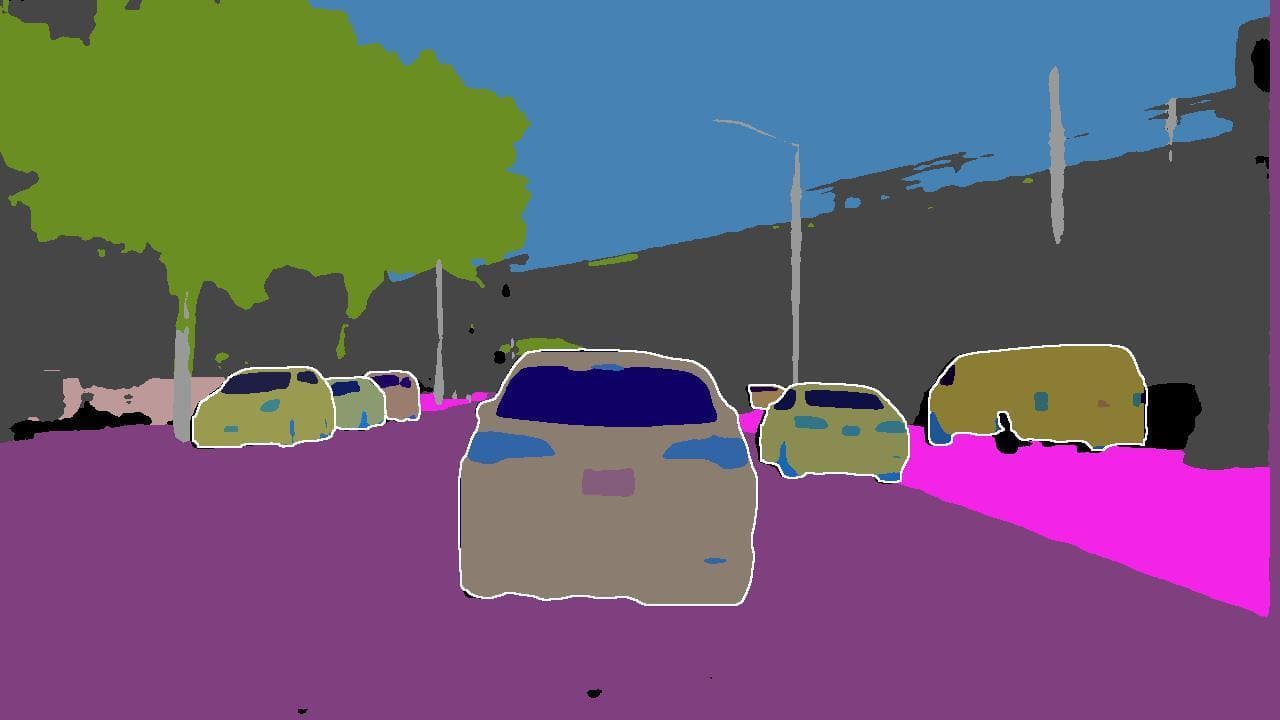}
    \end{subfigure}
    
    \begin{subfigure}{0.32\linewidth}
        \includegraphics[width=\linewidth]{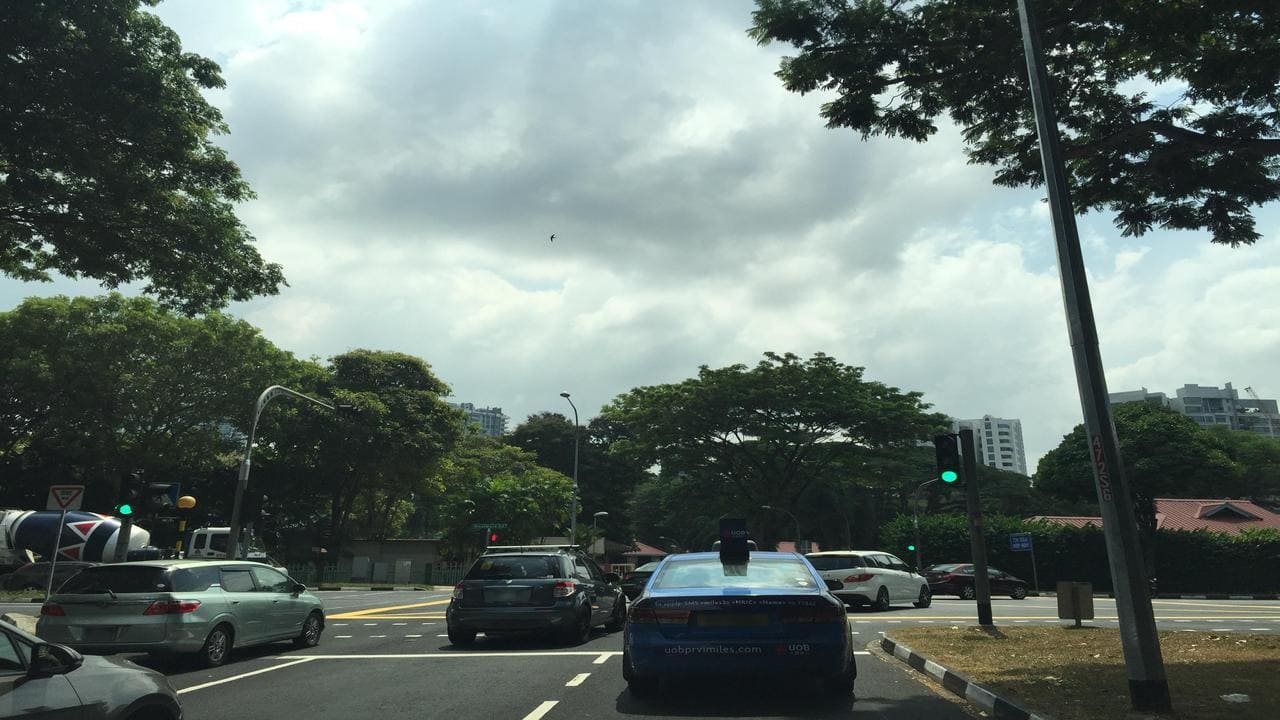}
    \end{subfigure}\hspace*{\fill}
    \begin{subfigure}{0.32\linewidth}
        \includegraphics[width=\linewidth]{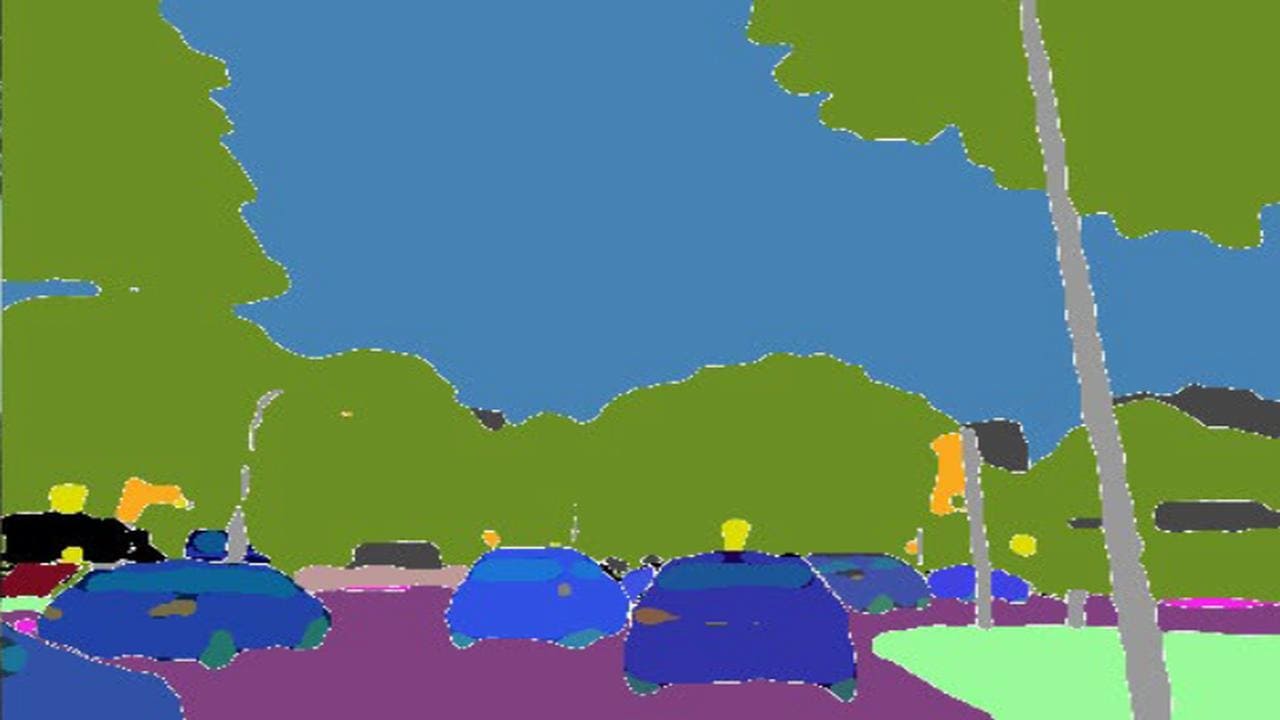}
    \end{subfigure}\hspace*{\fill}
    \begin{subfigure}{0.32\linewidth}
        \includegraphics[width=\linewidth]{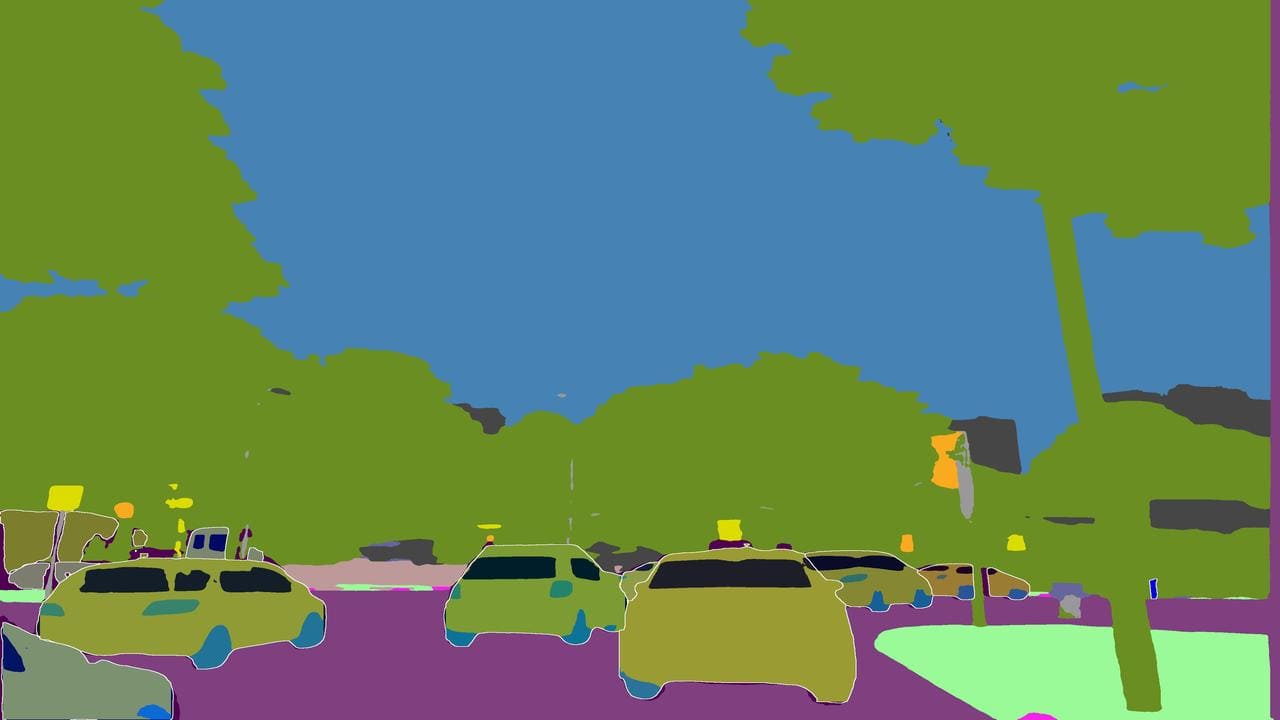}
    \end{subfigure}
    
    \begin{subfigure}{0.32\linewidth}
        \includegraphics[width=\linewidth]{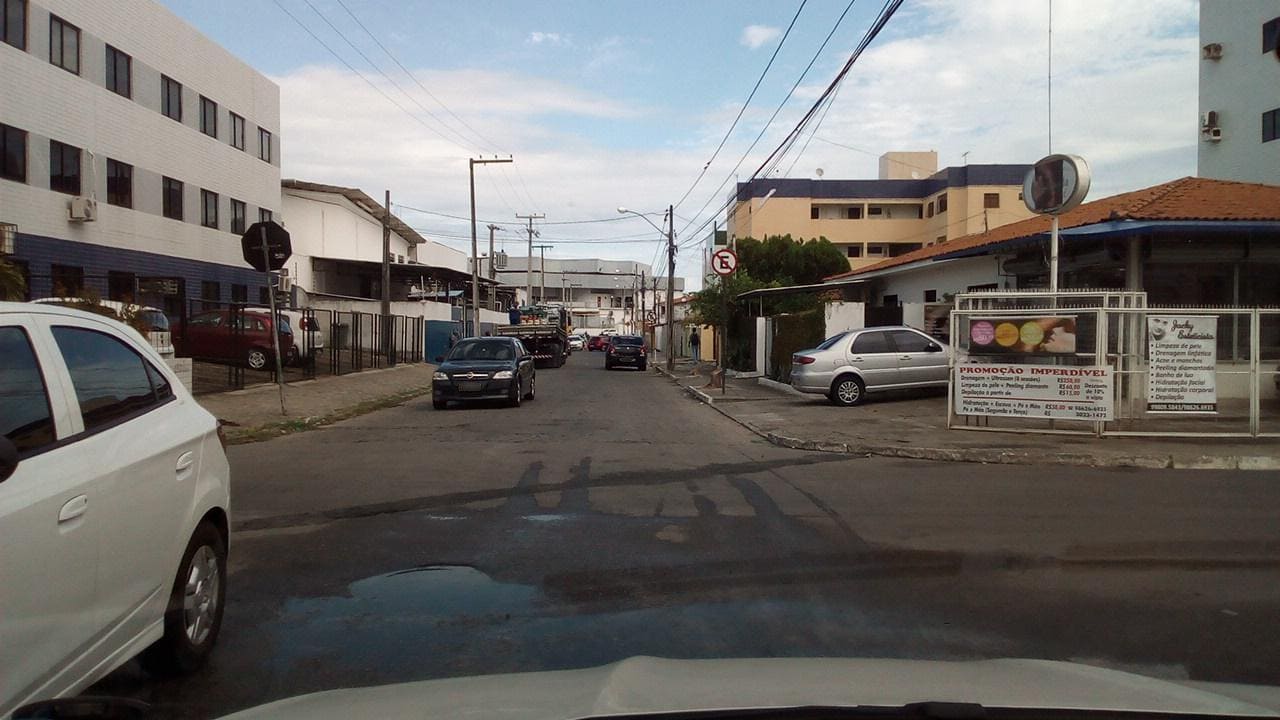}
        \subcaption*{Input Images}
    \end{subfigure}\hspace*{\fill}
    \begin{subfigure}{0.32\linewidth}
        \includegraphics[width=\linewidth]{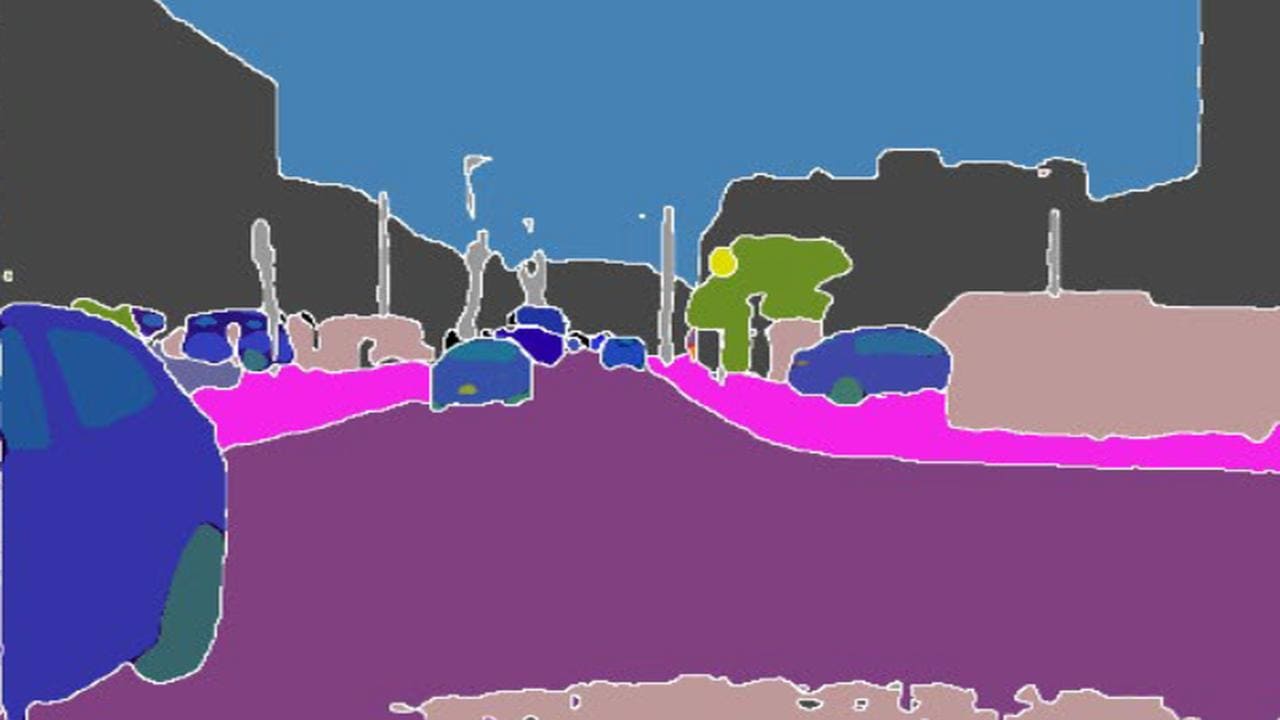}
        \subcaption*{PPF++ \cite{li2023panopticpartformer}}
    \end{subfigure}\hspace*{\fill}
    \begin{subfigure}{0.32\linewidth}
        \includegraphics[width=\linewidth]{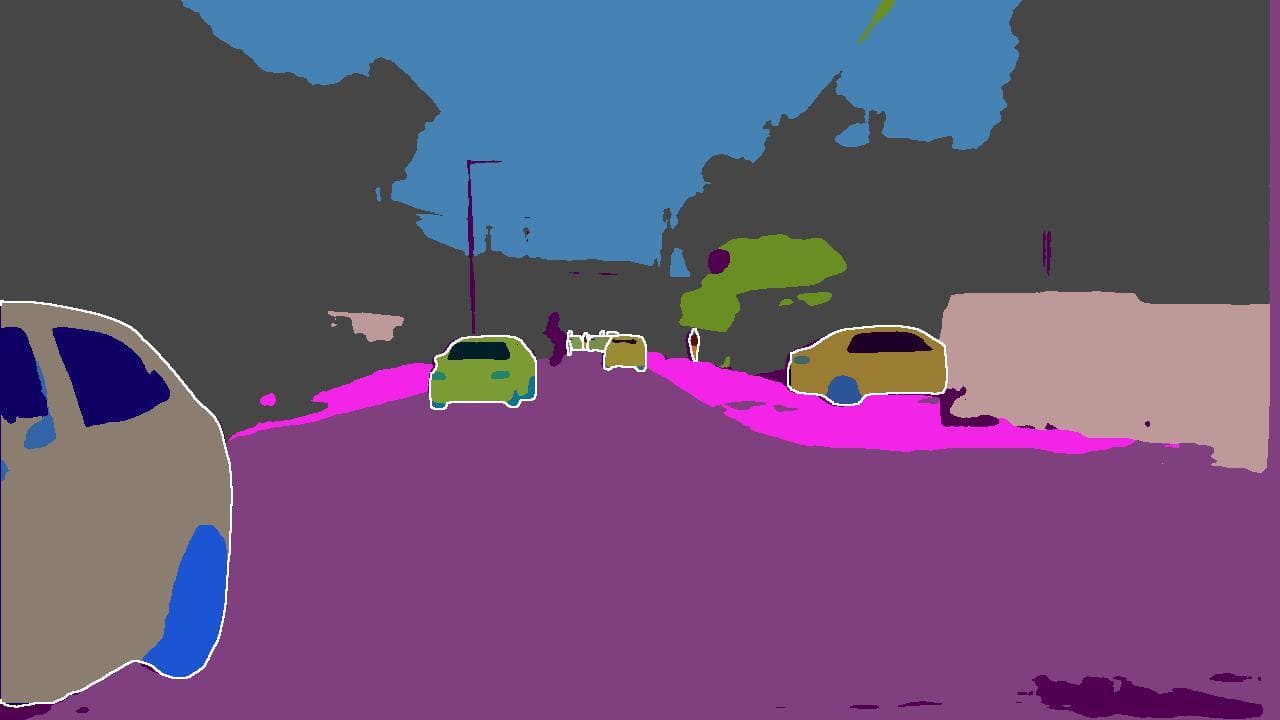}
        \subcaption*{Results of JPPF}
    \end{subfigure}
    
    \caption{We visually compare the generalization capabilities of PPF++ \cite{li2023panopticpartformer} and our JPPF on BDD100K \cite{yu2020bdd100k} (first two rows) and Mapillary Vistas \cite{neuhold2017mapillary} (last two rows) without fine-tuning}
    \label{fig:appendix:generalization}
\end{figure*}




\end{appendices}


\end{document}